\documentclass[runningheads]{llncs}
 
\usepackage{eccv}

\usepackage{graphicx}
\usepackage{amsmath,amssymb} 
\usepackage{color}
\usepackage[width=122mm,left=12mm,paperwidth=146mm,height=193mm,top=12mm,paperheight=217mm]{geometry}
\usepackage{float} 

\usepackage{cite}
\usepackage{xcolor, colortbl}
\usepackage[normalem]{ulem}
\usepackage[accsupp]{axessibility}  
\newcommand{\terrain}{\includegraphics[scale=0.030]{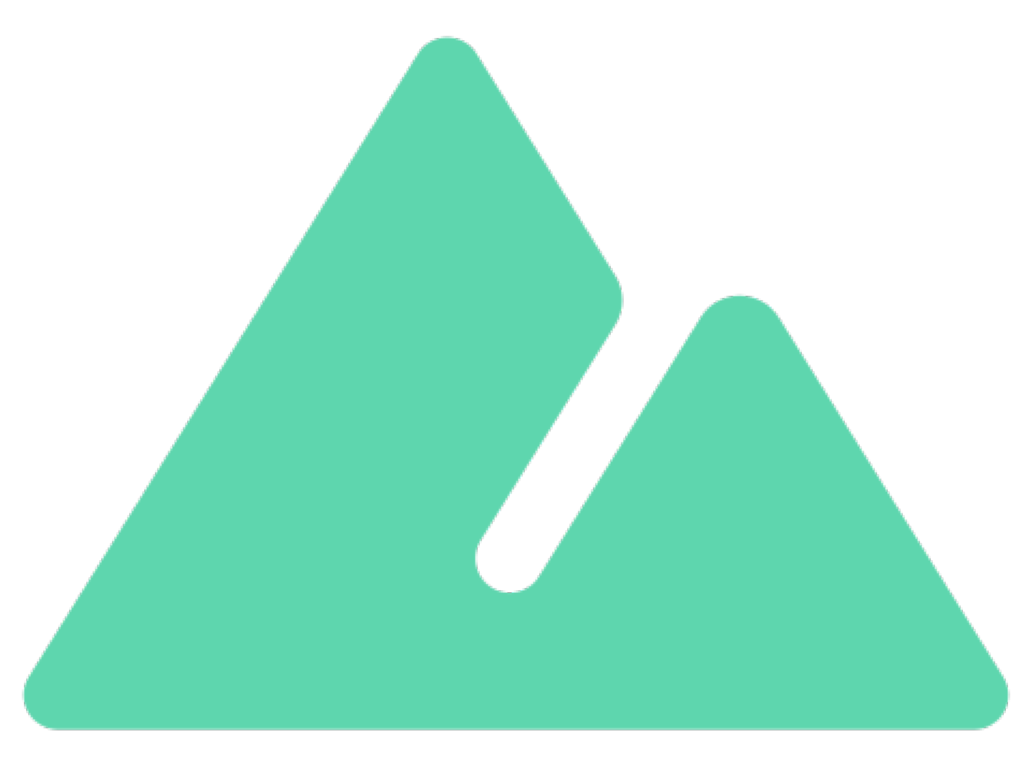}}
\newcommand{\sidewalk}{\includegraphics[scale=0.030]{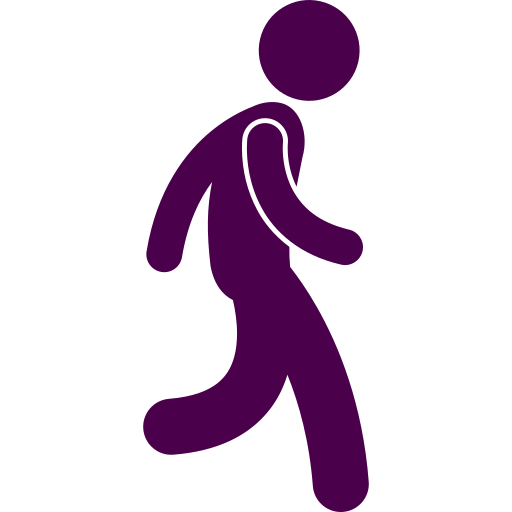}}
\newcommand{\pole}{\includegraphics[scale=0.025]{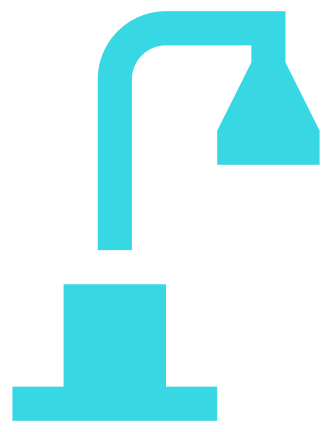}}

\usepackage{longtable, array, booktabs}
\usepackage{makecell}
\usepackage{arydshln}
\usepackage{multirow}
\usepackage{pifont}
\newcommand{\cmark}{\ding{51}}%
\newcommand{\xmark}{\ding{55}}%
\usepackage{array}
\newcommand{\boldline}{\specialrule{1.0pt}{0pt}{0pt}}
\usepackage{subcaption}
\definecolor{darkred}{rgb}{0.6148, 0., 0.}
\newcolumntype{?}{!{\vrule width 1pt}}
\newcolumntype{|}{!{\vrule width .5pt}}
\usepackage{color, colortbl, xcolor}
\usepackage{multicol}
\usepackage{tikz}
\usepackage{tikz-cd}
\usetikzlibrary{arrows,automata}
\usepackage{gensymb}
\usepackage{soul}
\usepackage{hhline}
\usepackage{graphicx}
\usepackage{wrapfig}
\newcommand*{\ov}[1]{
  $\m@th\overline{\mbox{#1\rule{0pt}{3mm}}}$
}
\definecolor{lightyellow}{rgb}{1., 1., 0.9}
\definecolor{lightred}{rgb}{1.,.8,.79}
\definecolor{lightblue}{rgb}{0.678,0.847,0.9}

\usepackage{caption}
\usepackage{tablefootnote}

\newcommand{\mycircle}[1]{\tikz{\node[draw=#1, color=white, fill=#1, circle,minimum
width=0.25cm,minimum height=0.25cm,inner sep=0pt, circular drop shadow, very thick] at (0,0) {};}}

\newcommand{\mystar}[1]{\tikz{\node[draw=#1, color=white, fill=#1, star,minimum
width=0.25cm,minimum height=0.25cm,inner sep=0pt, circular drop shadow, very thick] at (0,0) {};}}

\definecolor{demo_color}{rgb}{0.85, 0.85, 0.85}
\newcommand{\voxelembeds}{\protect\mycircle{demo_color}}
\newcommand{\rangembeds}{\protect\mystar{demo_color}}
\newcommand{\twoembeds}[2]{
\definecolor{color_name}{rgb}{#1}
\protect\mycircle{color_name}\mystar{color_name} \textbf{#2} \hspace{-2pt}}
\usetikzlibrary{shadows}
\usetikzlibrary{plotmarks}
\usetikzlibrary{shapes}

\definecolor{rangered}{rgb}{1.,  0.26, 0.26}
\definecolor{voxelblue}{rgb}{0.25,  0.41, 0.9}
\usepackage{eccvabbrv}

\usepackage{graphicx}
\usepackage{booktabs}

\usepackage[accsupp]{axessibility}  
\usepackage{fontawesome5}

\usepackage[pagebackref,breaklinks,colorlinks,citecolor=eccvblue]{hyperref}

\usepackage{orcidlink}

\makeatletter
\def\blfootnote{\gdef\@thefnmark{}\@footnotetext}
\makeatother
\usepackage[ruled,vlined,linesnumbered,resetcount,algochapter]{algorithm2e}
\begin{document}

\title{ItTakesTwo: Leveraging Peer Representations for Semi-supervised LiDAR Semantic Segmentation} 

\titlerunning{ItTakesTwo}

\author{Yuyuan Liu*\inst{1}$^{\text{(\faEnvelope[regular])}}$ \and Yuanhong Chen*\inst{1} \and Hu Wang\inst{2,1} \and Vasileios Belagiannis\inst{3} \and \\Ian Reid\inst{2, 1} \and Gustavo Carneiro\inst{4,1}}

\authorrunning{Y.~Liu et al.}

\institute{Australian Institute for Machine Learning (AIML), University of Adelaide, AU \and
Mohamed bin Zayed University of Artificial Intelligence (MBZUAI), UAE \and
University of Erlangen–Nuremberg, Germany \and
Centre for Vision, Speech and Signal Processing (CVSSP), University of Surrey, UK\\}

\maketitle

\begin{abstract}
The costly and time-consuming annotation process to produce large training sets for modelling semantic LiDAR segmentation methods has motivated the development of semi-supervised learning (SSL) methods. However, such SSL approaches often concentrate on employing consistency learning only for individual LiDAR representations. This narrow focus results in limited perturbations that generally fail to enable effective consistency learning. 
Additionally, these SSL approaches employ contrastive learning based on the sampling from a limited set of positive and negative embedding samples.
This paper introduces a novel semi-supervised LiDAR semantic segmentation framework called ItTakesTwo (IT2). IT2 is designed to ensure consistent predictions from peer LiDAR representations, thereby improving the perturbation effectiveness in consistency learning. Furthermore, our contrastive learning employs informative samples drawn from a distribution of positive and negative embeddings learned from the entire training set. 
Results on public benchmarks show that our approach achieves remarkable improvements over the previous state-of-the-art (SOTA) methods in the field. \textit{The code is available at:} \url{https://github.com/yyliu01/IT2}.
\keywords{Semi-supervised Learning \and LiDAR Semantic Segmentation}
\end{abstract}    
\section{Introduction}
\label{sec:introduction}
\blfootnote{* denotes equal contribution. \; $^{\text{(\faEnvelope[regular])}}$ Email: \email{yuyuan.liu@adelaide.edu.au}.}Outdoor LiDAR semantic segmentation is emerging as an important area of research given its application to self-driving vehicles~\cite{grigorescu2020survey,guo2020deep}. Although LIDAR data has advantages relative to photometric camera data, it also presents unique challenges, such as data sparsity~\cite{zhu2021cylindrical} and varying point density~\cite{kong2023lasermix,milioto2019rangenet++}. Most existing works alleviate these issues by transforming the LiDAR point-clouds to other representations, including range-based image~\cite{wu2018squeezeseg,kong2023rethinking,fan2021rangedet,milioto2019rangenet++}, voxel grids~\cite{graham20183d,zhu2021cylindrical,hou2022point}, or fusing them with other modalities~\cite{liu2023bevfusion, liu2023segment} and views~\cite{xu2021rpvnet}. 
Despite their success, those methods rely on supervised learning that requires large labelled training sets, which can be problematic given the costly and time-consuming labelling process~\cite{liu2022perturbed,ouali2020semi}.

\begin{figure}[t!]
    \centering
    \includegraphics[width=\textwidth]{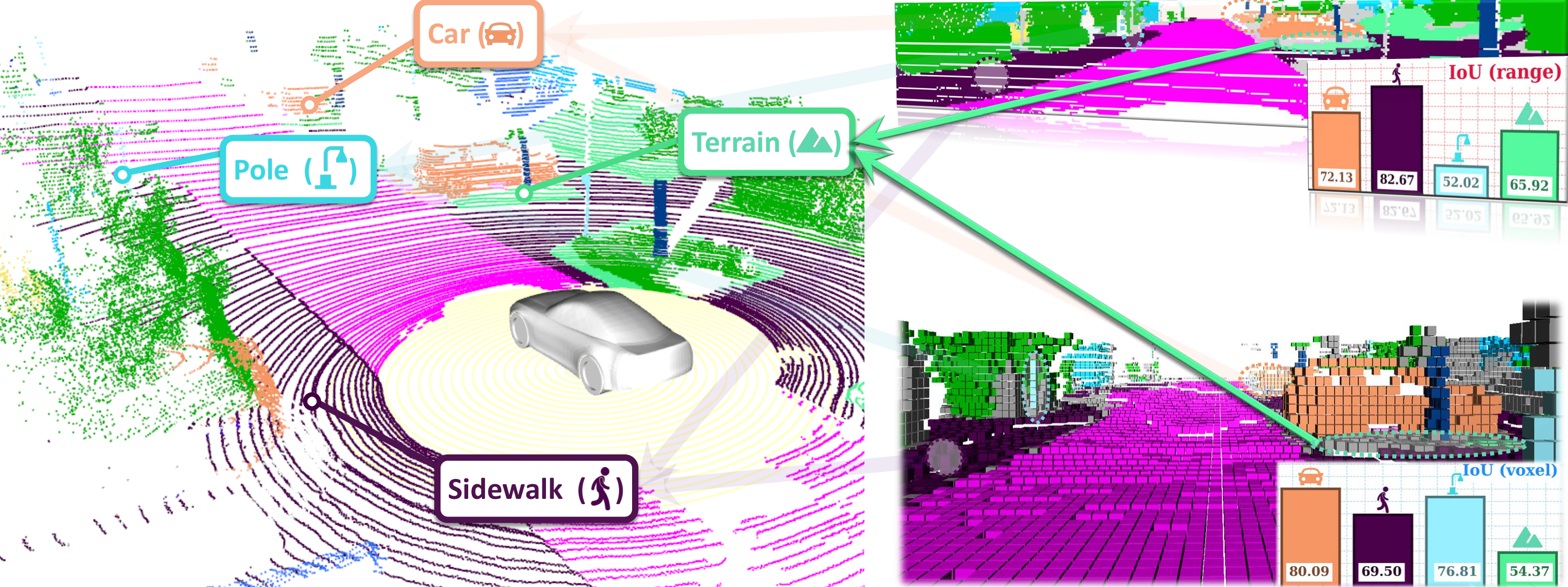}
    \caption{
    The \textbf{left} figure represents the 3D point cloud scan (ground truth of four classes -- Car, Pole, Terrain, and Sidewalk) while the \textbf{right} figures denote the prediction results (into the same four classes), provided by our model, from \textcolor{rangered}{range (top-right)} and \textcolor{voxelblue}{voxel (bottom-right)} representations, where the incorrect predictions are marked in \textcolor{gray}{gray}. Please notice that: 1) the mapping of the four classes from the 3D point cloud scan to each representation demonstrates the consistent distribution of semantic classes, and 2) the histograms of the predictions for each representation exhibit distinct differences.    
    }
    \label{fig:demo}
    \vspace{-7.5pt}
\end{figure}
These issues have motivated semi-supervised learning (SSL) innovations for LiDAR semantic segmentation~\cite{kong2023lasermix, jiang2021guided, li2023less}, which  require a small labelled dataset used in conjunction with a large unlabelled dataset. 
Consistency Learning, one of the main strategies explored in SSL for image semantic segmentation, penalises inconsistent predictions from perturbed inputs, where these perturbations take various forms, 
including input augmentation~\cite{yang2021st++,zou2020pseudoseg}, adversarial noise~\cite{liu2022perturbed,hung2018adversarial}, network perturbations~\cite{chen2021semi} and dropout~\cite{yang2023revisiting}. The incorporation of strong and varied perturbations is crucial for enhancing generalisation~\cite{french2019semi, liu2022perturbed}. 

Current research~\cite{kong2023lasermix, li2023less} applies such consistency learning in SSL for LiDAR points using many strategies, such as laser-beam augmentation~\cite{kong2023lasermix} or Mean Teacher (MT)~\cite{li2023less} framework. 
Despite their success, these approaches solely rely on a single LiDAR representation (e.g., voxel grid) for training, which could potentially limit the model's ability to generalize effectively due to the restricted perturbations.
Such an issue can be mitigated with the use of multiple LiDAR representations for consistency learning that explores the fundamental Clustering assumption\footnote{\protect Cluster hypothesis (or assumption)~\cite{manning2009introduction}: the data of various forms that behave similarly with respect to information relevance should be clustered together.}, where the results from different representations naturally indicate the same underlying semantic information from the 3D LiDAR points.
For example, as illustrated in Fig.~\ref{fig:demo}, the class Terrain (\terrain) in the \textcolor{rangered}{top-right pixels} and \textcolor{voxelblue}{bottom-right voxels} should be "clustered" according to the (left) original 3D point data. 
Thus, the peer LiDAR representations can be regarded as a novel form similar to perturbation, but yet under-explored in the field.
Other disadvantages of focusing on only one of the representations when using small labelled datasets are: the range view suffers from the information loss during the projection process, particularly for the tail categories, such as Pole (\pole\hspace{1pt}), which has a low mIoU of 52.02; and the voxel grid tends to show lower accuracy in dense areas that are far away from the LiDAR source, such as Sidewalk (\sidewalk\hspace{1pt}), which has a low mIoU of 69.50.
Such limited pseudo label results will inevitably introduce confirmation bias (i.e., the problem of relying on and reinforcing wrong predictions during training), reducing even further the efficacy of consistency learning based on single representation~\cite{kong2023lasermix,li2023less,jiang2021guided}.

Another effective technique for semi-supervised LiDAR segmentation is contrastive learning~\cite{jiang2021guided,li2023less,cheng2021sspc,hou2021exploring}, which consists of attracting positive and repelling negative embedding pairs at instance~\cite{nunes2022segcontrast} or category~\cite{jiang2021guided,li2023less} levels, thereby enhancing the latent space point distribution.
However, to generate these pairs, methods typically rely on a limited number of embeddings extracted from randomly selected training samples due to the high computational cost associated with dense contrastive learning tasks~\cite{jiang2021guided,li2023less}. Unfortunately, this limited sample size hinders a comprehensive understanding of the distribution across the entire embedding space~\cite{caron2020unsupervised}, resulting in suboptimal convergence of LiDAR segmentation in SSL. This limitation becomes particularly pronounced when dealing with embeddings from different models in multiple views, where potential noisy data~\cite{miech2020end,tian2020contrastive} or model bias~\cite{xu2022multi} from one network can adversely affect others, thereby impeding the optimization process.

In this work, we propose a novel semi-supervised LiDAR semantic segmentation framework, named ItTakesTwo (IT2).
IT2 employs consistency learning based on peer-representations perturbation, which enforces the consistent predictions across two different representations. 
Such perturbation contributes with a more effective SSL generalisation and alleviates issues related to pseudo label qualities from a single representation.
Furthermore, we propose cross-distribution contrastive learning, where instead of randomly selecting training samples in the embedding space, we learn informative  embedding distributions (also known as prototype) via Gaussian Mixture Models from both representations. Then the embedding samples are drawn from such prototypical distributions to produce representative features that enhance the understanding of the latent space in peer representation scenarios. 
In addition, instead of applying a single type of data augmentation to both LiDAR representations~\cite{kong2023lasermix}, we empirically  show that applying different augmentations to distinct representations provides better generalisation.
To summarise, our work makes the following key contributions:
\begin{enumerate}
    \item A novel SSL LiDAR segmentation framework (IT2) that leverages peer-representation perturbation in consistency learning, encouraging the same predictions for the points with the same semantic meaning in different LiDAR representations;
    \item A new contrastive learning where informative positive and negative pairs of embeddings are sampled from prototype distributions (modeled via Gaussian Mixture Models) of both representations; and   
    \item An innovative representation-specific data augmentation that enables IT2 achieve better generalisation.
\end{enumerate}
Our IT2 has state-of-the-art (SOTA) performance on nuScenes~\cite{caesar2020nuscenes}, SemanticKitti~\cite{behley2019semantickitti} and ScribbleKitti~\cite{unal2022scribble}, and we also achieve best performance with different sampling strategies to form the labelled and unlabelled subsets. 
Notably, in nuScenes~\cite{caesar2020nuscenes} dataset, we outperform the previous SOTA~\cite{kong2023lasermix} by around $4.6\%$ in the limited labelled data situations.

\section{Related Work}
\label{sec:related_work}
\noindent \textbf{LiDAR Semantic Segmentation} is an essential task for autonomous driving~\cite{guo2020deep}. To more effectively extract the feature information from the unordered LiDAR point data, current approaches~\cite{milioto2019rangenet++, zhu2021cylindrical, zhuang2021perception} transform such point data into different representations. Range representation methods~\cite{milioto2019rangenet++, zhao2021fidnet, kong2023rethinking} transfer the LiDAR points to the spherical coordinates~\cite{milioto2019rangenet++} and project them to the image space~\cite{zhao2021fidnet, kong2023rethinking}.
Voxel representation methods~\cite{graham20183d, zhu2021cylindrical, lai2023spherical} voxelise the LiDAR points and project them into different sub-types of geometry views to alleviate the sparse data challenges~\cite{graham20183d,cciccek20163d}, encompassing cartesian~\cite{vora2021nesf}, spherical~\cite{lai2023spherical} and cylindrical~\cite{zhu2021cylindrical} coordinate systems.
Furthermore, some methods~\cite{liu2023segment,liang2022bevfusion} take inputs from a combination of LiDAR representations~\cite{reichardt2023360deg,zhuang2021perception} or incorporate inputs from camera modalities~\cite{liu2022bevfusion, liang2022bevfusion}, fusing them in the intermediate layer to enhance feature understanding.
Although these supervised methods show promising results~\cite{zhu2021cylindrical,zhao2021fidnet,xu2021rpvnet}, there has been a longstanding interest in minimising the extensive labeling efforts, with approaches such as scribble annotation~\cite{unal2022scribble}, box supervision~\cite{liu2022box2seg}, active learning~\cite{liu2022less,xu2023hierarchical}, and SSL~\cite{kong2023lasermix,jiang2021guided}.
In this paper, we focus on SSL to alleviate such annotation costs. 

\noindent \textbf{Semi-supervised LiDAR Semantic Segmentation} leverages  unlabelled data to enhance the model's generalisation beyond the small labelled dataset.
The current literature on LiDAR SSL primarily focuses on indoor~\cite{cheng2021sspc,deng2022superpoint,xu2023hierarchical,li2022hybridcr} or object~\cite{kohli2020semantic,sun2023semi} point cloud scenarios, which have uniformly distributed and high density points in the scan. 
Such scenario is markedly different from the representation of outdoor LiDAR points that pose unique challenges, such as complex structure~\cite{kong2023lasermix} and low density points~\cite{zhu2021cylindrical}. 
Previous semi-supervised LiDAR segmentation methods~\cite{li2023less,jiang2021guided,nunes2022segcontrast} mainly rely on contrastive learning to improve the latent space understanding. For instance, GPC~\cite{jiang2021guided} utilises confidence-threshold to constrain potentially noisy pseudo labels, and Lim3D~\cite{li2023less} uses a memory bank to save negative samples. 
To increase the variation of the LiDAR data, LaserMix~\cite{kong2023lasermix} proposes a novel augmentation that mixes laser beams in various spatial positions and holds the consistency learning under the Mean Teacher (MT) architecture~\cite{tarvainen2017mean}.  However, these methods use only single LiDAR representations which, as explained in~\cref{sec:introduction}, imposes many limitations, such as limited perturbation effectiveness and poor performance for specific categories. Our framework introduces a novel peer-representation perturbation that employs various LiDAR representations during training, resulting in improved generalisation.

\noindent \textbf{Contrastive Learning}~\cite{khosla2020supervised,wang2021exploring,caron2020unsupervised,xie2020pointcontrast} enforces latent embeddings to be closer to instances with the same semantic meaning (i.e. positive) and to be far away from instances with different semantic meanings (i.e. negative). 
This approach has been explored in LiDAR segmentation~\cite{li2023less,jiang2021guided,nunes2022segcontrast,cheng2021sspc,deng2022superpoint}, but the reliance on actual training samples leads to sub-optimal training as only a subset of embedding pairs~\cite{jiang2021guided,li2023less} can be selected for computing the contrastive loss.
This becomes more challenging when working with embeddings obtained from different representations~\cite{xu2022multi,tian2020contrastive,liu2022trusted}, as potentially noisy predictions~\cite{xu2022multi,miech2020end} from each representation can result in confirmation bias during training~\cite{arazo2020pseudo}. 
Prototypical contrastive learning~\cite{li2020prototypical,caron2020unsupervised,le2020contrastive,liang2022gmmseg,he2020momentum}  explores the use of a set of representative embeddings for the contrastive learning via memory bank~\cite{wang2021exploring,he2020momentum}, optimal transport~\cite{caron2020unsupervised}, or deep clustering~\cite{li2020prototypical,le2020contrastive}.  Nevertheless, such prototypes represent the modes of a more complex underlying distribution of prototypes that has not been explored before for SSL LiDAR segmentation.
In our framework, we employ Gaussian Mixture Models (GMMs) to represent class-specific embedding distributions from both LiDAR representations and to generate informative embeddings for all classes, enhancing the effectiveness of the contrastive learning.

\vspace{-5pt}

\section{Methodology}
\newcommand{\xdasharrow}[2][->]{
\tikz[baseline=-\the\dimexpr\fontdimen22\textfont2\relax]{
\node[anchor=south,font=\large, inner ysep=1.5pt](x){#2};
\draw[line width = .35mm, dotted,#1](x.south west)--(x.south east);
}
}

\vspace{-2.5pt}
\begin{figure*}[t!]
\twoembeds{0.4862745098039216,0.0,0.4862745098039216}{sidewalk} 
\twoembeds{1,0.0,1}{road}
\twoembeds{0.3,0.9117647058823529,0.5490196078431373}{terrain}
\twoembeds{0.0,0.6862745098039216,0.0}{vegetation}
\twoembeds{1.0,0.6196078431372549,0.0}{car}
\twoembeds{0.0,0.35,0.7019607843137255}{trunk}
    \centering
    \includegraphics[width=\textwidth]{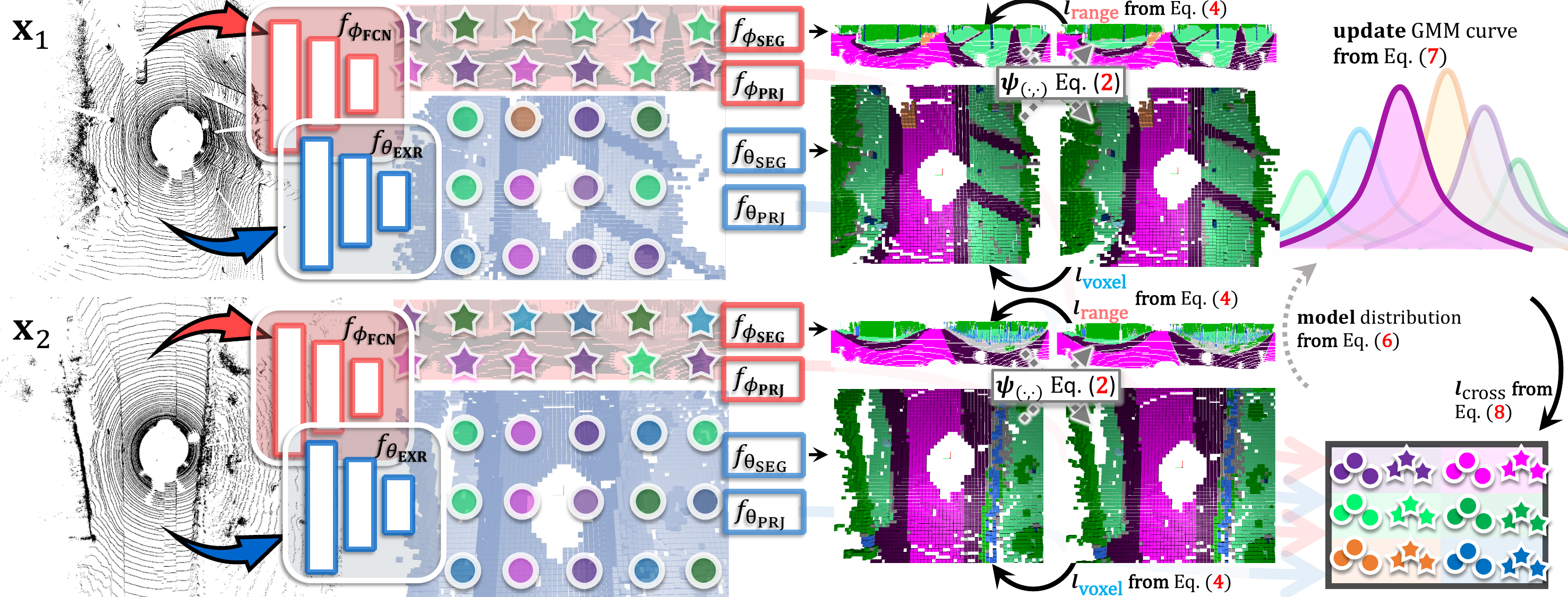}
    
    \caption{Illustration of our approach with unlabeled data. The point scans in the batch ($\mathbf{x}_1$ and $\mathbf{x}_2$ on the left) are transformed into \textcolor{rangered}{range} (and processed by $f_{\phi_{\textbf{FCN}}}(.)$) and \textcolor{voxelblue}{voxel} (and processed by $f_{\theta_{\textbf{EXR}}}(.)$) representations. Subsequently, the results from $f_{\phi_\textbf{SEG}}(f_{\phi_{\textbf{FCN}}}(.))$ and $f_{\theta_\textbf{SEG}}(f_{\theta_{\textbf{EXR}}}(.))$ are transformed to the other representation with Eq.~\eqref{eq:pseudo_label} and cross-supervised with Eq.~\eqref{eq:range_voxel_loss}. Also, the embedding \{\rangembeds,\hspace{1pt}\voxelembeds\} selected by the pseudo labels are utilised to model the GMMs with Eq.~\eqref{eq:gmm} and 
    penalised by the informative samples (generated from GMMs) based on $\ell_{cross}(.)$ in Eq.~\eqref{eq:contras}. Please note the arrow \textcolor{gray!70}{\textbf{\xdasharrow{$\;$}}} indicates that the operation does not require gradient computation.
    }
    \label{fig:main_struct}
    \vspace{-10pt}
\end{figure*}
Our approach is based on a labelled point dataset $\mathcal{P}_L=\{(\mathbf{x}_i, \mathbf{y}_i)\}_{i=1}^{|\mathcal{P}_L|}$ consisting of $|\mathcal{P}_L|$ scans and an unlabelled set $\mathcal{P}_U=\{(\mathbf{x}_i\}_{i=1}^{|\mathcal{P}_U|}$ with $|\mathcal{P}_U|$ scans.  
Each scan $\mathbf{x}_i \in \mathcal{P} \subset \mathbb{R}^{3\times C \times N}$ has \textbf{3} spatial coordinates, with $C$ denoting the number of features (intensity, elongation, etc.), $N$ representing the number of points, and $\mathbf{y}_i \in \mathcal{Y} \subset \{0, 1\}^{Y\times N}$ represents the  \textit{one-hot} label with $Y$ being the number of classes in the labelled set. 
This point cloud dataset is transformed into the range and voxel representations to build the two peer representation datasets that is required by our method.
These transformations are defined by~\cite{xu2021rpvnet} as:
\begin{itemize}
    \item $\psi_{p \rightarrow v}: \mathcal{P}\rightarrow \mathcal{V}$ is a point to voxel projection function that forms $\mathcal{V}_{L}=\{(\mathbf{x}^v_i,\mathbf{y}^v_i) | (\mathbf{x}_i,\mathbf{y}_i) \in \mathcal{P}_{L} \}_{i=1}^{|\mathcal{P}_{L}|}$
    and $\mathcal{V}_{U}=\{\mathbf{x}^v_i | \mathbf{x}_i \in \mathcal{P}_{U} \}_{i=1}^{|\mathcal{P}_{U}|}$, with each volume $\mathbf{x}^v_i \in \mathcal{V} \subset \mathbb{R}^{C_v \times H \times W \times L}$ having size $H \times W \times L$ and $C_v$ features, and
    the projected label being $\mathbf{y}^v_i\in\{0, 1\}^{Y \times H \times W \times L}$ in \textit{one-hot} format.
    \item $\psi_{p \rightarrow r}: \mathcal{P}\rightarrow \mathcal{R}$ is a function to project point to the range image\footnote{\noindent \protect Range-to-point projection can lead to information loss~\cite{milioto2019rangenet++}, which is typically mitigated by post-processing with K-Nearest Neighbors (KNN). In pursuit of efficiency, our IT2 does not use any post-processing during training.}. With this, we build $\mathcal{R}_{L}=\{(\mathbf{x}^r_i,\mathbf{y}^r_i) | (\mathbf{x}_i,\mathbf{y}_i) \in \mathcal{P}_{L} \}_{i=1}^{|\mathcal{P}_{L}|}$ and $\mathcal{R}_{U}=\{\mathbf{x}^r_i | \mathbf{x}_i \in \mathcal{P}_{U} \}_{i=1}^{|\mathcal{P}_{U}|}$, with each range image $\mathbf{x}^r_i \in \mathcal{R} \subset \mathbb{R}^{C_r \times U \times V}$ having size $U \times V$ with $C_r$ features, and the projected label being $\mathbf{y}^r_i\in \{0, 1\}^{Y \times U \times V}$.
\end{itemize}
The inverse of the transformations above are respectively denoted by $\psi_{v\rightarrow p}$ and $\psi_{r\rightarrow p}$. 
Based on the transformed range and voxel representations from $\mathcal{P}_L$ and $\mathcal{P}_U$, we define the labelled set $\mathcal{D}_{L} = \{(\mathbf{x}^v_i,\mathbf{y}^v_i), (\mathbf{x}^r_i,\mathbf{y}^r_i) | (\mathbf{x}^v_i,\mathbf{y}^v_i) \in \mathcal{V}_{L}, (\mathbf{x}^r_i,\mathbf{y}^r_i) \in \mathcal{R}_{L},
\}_{i=1}^{|\mathcal{D}_L|}$ and the larger unlabelled set $\mathcal{D}_U = \{ (\mathbf{x}^v_i, \mathbf{x}^r_i) | \mathbf{x}^v_i \in \mathcal{V}_{U}, \mathbf{x}^r_i \in \mathcal{R}_{U} \}_{i=1}^{|\mathcal{D}_U|}$ 
where $|\mathcal{D}_L| \leq |\mathcal{D}_U|$. These datasets are used to train our IT2 methodology in~\cref{method:it2}, while~\cref{method:cross} describes our Gaussian Mixture Models (GMMs) based prototypical contrastive learning. Lastly, we introduce the details of our augmentation method in~\cref{method:aug}.
\vspace{-5pt}

\subsection{ItTakesTwo (IT2) Framework}
\label{method:it2}
As illustrated in~\cref{fig:main_struct}, our approach enforces prediction consistency between two peer representations to explore a potentially more effective perturbation. 
Our method can use various LiDAR representations, but 
we follow~\cite{kong2023lasermix} and employ range and voxel networks for a fair comparison in the benchmark.
\textbf{The range image network} consists of an FCN Network (e.g., ResNet~\cite{he2016deep}) that maps the input to a latent dense feature space with $f_{\phi_{\textbf{FCN}}} : \mathcal{R} \rightarrow \mathcal{I}$, where $\mathcal{I} \subset \mathbb{R}^{U\times V \times D_r}$ with $D_r$ denoting feature dimensionality. 
Then a segmentation head will map $\mathcal{I}$ to the categorical segmentation map with $f_{\phi_{\textbf{SEG}}}: \mathcal{I} \rightarrow \mathbb{R}^{U\times V \times Y}$. Similarly, \textbf{the voxel network} extracts features with $f_{\theta_{\textbf{EXR}}} : \mathcal{V} \rightarrow \mathcal{K}$, where $\mathcal{K} \in \mathbb{R}^{H\times W \times L \times D_v}$ with $D_v$ denoting the feature dimensionality, followed by a segmentation head $f_{\theta_{\textbf{SEG}}}: \mathcal{K} \rightarrow \mathbb{R}^{H\times W \times L \times Y}$. The \textbf{peer representation consistency training} depends on the following categorical distributions for both views:
\begin{equation}
\begin{aligned}
    & \mathbf{\hat{y}}^r(\omega^r) = \texttt{\small softmax}^{(\omega^r)}(f_{\phi_{\textbf{SEG}}}(f_{\phi_{\textbf{FCN}}}(\mathbf{x}^r))), \\
    & \mathbf{\hat{y}}^v(\omega^v) = \texttt{\small softmax}^{(\omega^v)}(f_{\theta_{\textbf{SEG}}}(f_{\theta_{\textbf{EXR}}}(\mathbf{x}^v))),
\end{aligned}
\end{equation}
where $\mathbf{\hat{y}}^r$ and $\mathbf{\hat{y}}^v$ are the categorical distributions for the range image (in $\omega^r$ in the $U \times V$ lattice $\Omega^r$) and voxel (in $\omega^v$ in the $H \times W \times L$ lattice $\Omega^v$) representations, and $\texttt{\small softmax}^{(\omega)}(.)$ computes the softmax activation at $\omega$. 
Based on these outputs, we perform the following cross-view transformations:
\begin{equation}
\label{eq:pseudo_label}
\resizebox{.65\hsize}{!}{$
\begin{aligned}
    & \mathbf{\Tilde{y}}^r(\omega^r) = \texttt{\small argmax} \: \Psi^{(\omega^r)}_{v \rightarrow r}(\mathbf{\hat{y}}^v), 
    \quad \mathbf{\mathbf{c}}^r(\omega^r) = \texttt{\small max}  \: \Psi^{(\omega^r)}_{v \rightarrow r}(\mathbf{\hat{y}}^v), \\
    & \mathbf{\Tilde{y}}^v(\omega^v) = \texttt{\small argmax} \: \Psi^{(\omega^v)}_{r \rightarrow v}(\mathbf{\hat{y}}^r), 
    \quad \mathbf{\mathbf{c}}^v(\omega^v) = \texttt{\small max}  \: \Psi^{(\omega^v)}_{r \rightarrow v}(\mathbf{\hat{y}}^r),
\end{aligned}
$}
\end{equation}
where $\mathbf{\Tilde{y}}^r$ and $\mathbf{\Tilde{y}}^v$ represent the transformed hard pseudo label maps, $\mathbf{c}^r$ and $\mathbf{c}^v$ denote the transformed confidence maps for range and voxel representations, respectively, $\Psi_{v \rightarrow r} = \psi_{p \rightarrow r} \circ \psi_{v \rightarrow p}$, $\Psi_{r \rightarrow v} = \psi_{p \rightarrow v} \circ \psi_{r \rightarrow p}$, and the superscript ${\omega}$ indicates that we only consider the output at $\omega$. 
We use the cross-view pseudo labels from Eq.~\eqref{eq:pseudo_label} to formulate the loss function to train IT2, as follows:
\begin{equation}
\resizebox{.99\hsize}{!}{$
\begin{aligned}
     \ell_{\text{IT2}}(\mathcal{D}_L,  \mathcal{D}_U, f_{\phi_{\textbf{FCN}}}, f_{\phi_{\textbf{EXR}}}, f_{\phi_{\textbf{SEG}}}) =  
      \ell_{\textcolor{rangered}{\text{range}}} (\mathcal{D}_L, \mathcal{D}_U, f_{\phi_{\textbf{FCN}}}, f_{\phi_{\textbf{SEG}}}) + \ell_{\textcolor{voxelblue}{\text{voxel}}} (\mathcal{D}_L, \mathcal{D}_U, f_{\theta_{{\textbf{EXR}}}},f_{\theta_{\textbf{SEG}}}),
\end{aligned}$}
\end{equation}
where we formulate the loss for each view as:
\begin{equation}
\resizebox{.9\hsize}{!}{$
\begin{aligned}
\ell_{\textcolor{rangered}{\text{range}}}(\mathcal{D}_L, \mathcal{D}_U, f_{\phi_{\textbf{FCN}}}, f_{\phi_{\textbf{SEG}}}) = 
 \frac{1}{|\mathcal{D}_L| \times |\Omega^r| } \sum_{i=1}^{|\mathcal{D}_L|} 
\sum_{\omega^r \in \Omega^r} \ell(\mathbf{y}_i^r(\omega^r),f^{(\omega^r)}_{\phi_{\textbf{SEG}}}(f_{\phi_{\textbf{FCN}}}(\mathbf{x}_i^r))\\
 +\frac{1}{|\mathcal{D}_U| \times |\Omega^r| } \sum_{i=1}^{|\mathcal{D}_U|} 
\sum_{\omega^r \in \Omega^r} \ell(\mathbf{\Tilde{y}}_i^r(\omega^r),f^{(\omega^r)}_{\phi_{\textbf{SEG}}}(f_{\phi_{\textbf{FCN}}}(\mathbf{x}_i^r)), \\
\ell_{\textcolor{voxelblue}{\text{voxel}}}(\mathcal{D}_L, \mathcal{D}_U, f_{\theta_{{\textbf{EXR}}}},f_{\theta_{\textbf{SEG}}}) = \frac{1}{|\mathcal{D}_L| \times |\Omega^v|} \sum_{i=1}^{|\mathcal{D}_L|} 
\sum_{\omega^v \in \Omega^v} \ell(\mathbf{y}_i^v{(\omega^v)},f^{(\omega^v)}_{\theta_{\textbf{SEG}}}(f_{\theta_{\textbf{EXR}}}(\mathbf{x}_i^v))\\
+\frac{1}{|\mathcal{D}_U| \times |\Omega^v|} \sum_{i=1}^{|\mathcal{D}_U|} \sum_{\omega^v \in \Omega^v} \ell(\mathbf{\Tilde{y}}_i^v{(\omega^v)},f^{(\omega^v)}_{\theta_{\textbf{SEG}}}(f_{\theta_{\textbf{EXR}}}(\mathbf{x}_i^v)),
\end{aligned}$}
\label{eq:range_voxel_loss}
\end{equation}
with the loss function $\ell(.)$ denoting the combination of cross-entropy loss and Lovasz-softmax~\cite{berman2018lovasz} loss, following~\cite{kong2023lasermix, li2023less, zhu2021cylindrical}.

\subsection{Cross-distribution Contrastive Learning}
\label{method:cross}

We model the cross distribution of the embedding space using a multi-layer perceptron to generate latent embeddings in parallel with the segmentation head.
Specifically, we obtain the range image embedding space via $f^{r}_{\phi_{\textbf{PRJ}}}: \mathcal{I} \rightarrow \mathbb{R}^{U\times V\times Z}$ with $Z$ denoting the pixel-wise number of dimensions in the projected space.
Similarly, the voxel embedding space is obtained via $f^{v}_{\theta_{\textbf{PRJ}}}: \mathcal{K} \rightarrow \mathbb{R}^{H\times W \times L \times Z}$, where $\mathcal{I}$ and $\mathcal{K}$ are intermediate feature spaces introduced in~\cref{method:it2}. We formulate the observed embedding results with:
\begin{equation}
\resizebox{.75\hsize}{!}{$
    \begin{aligned}
    & \mathbf{z}^r(\omega^r) = f^{(\omega^r)}_{\phi_{{\textbf{PRJ}}}}(f_{\phi_{\textbf{FCN}}}(\mathbf{x}^r)), 
    & \mathbf{z}^v(\omega^v) = f^{(\omega^v)}_{\theta_{{\textbf{PRJ}}}}(f_{\theta_{{\textbf{EXR}}}}(\mathbf{x}^v)),
    \end{aligned}
    \label{eq:range_voxel_representations}
    $}
\end{equation}
where $\mathbf{z}^r$ and $\mathbf{z}^v$ represent the latent point-wise embeddings of the range and voxel representations, respectively.

As mentioned in Sec.~\ref{sec:related_work}, contrastive learning methods~\cite{li2023less,jiang2021guided,nunes2022segcontrast,cheng2021sspc,deng2022superpoint} only use a limited subset of actual training samples present in a mini batch, 
potentially leading to suboptimal convergence with confirmation bias.
We propose a novel approach to represent the range and voxel embeddings with class-specific Gaussian Mixture Models (GMMs). Such approach enables the sampling of the most informative embeddings, fostering a comprehensive understanding of the distribution across the feature space of different representations.
In practice, we concatenate the embeddings from both range and voxel based on their categories, where such embedding set is defined by $\mathcal{G}^{y} = \{ (\mathbf{z},\mathbf{c},y) | \mathbf{z} = \mathbf{z}_i^r(\omega^r), \mathbf{c} = \mathbf{c}_i^r(\omega^r), y = \mathbf{\Tilde{y}}_i^r(\omega^r) \} \bigcup \{(\mathbf{z},\mathbf{c},y) | \mathbf{z} = \mathbf{z}_i^v(\omega^v), \mathbf{c} = \mathbf{c}_i^v(\omega^v), y = \mathbf{\Tilde{y}}_i^v(\omega^v)\}$ for the unlabelled data indexed by $i \in \{1,...,|\mathcal{D}_{U}|\}$, with the related confidence (i.e., $\mathbf{c}^r$ and  $\mathbf{c}^v$) and pseudo labels (i.e., $\mathbf{\Tilde{y}}^r$ and  $\mathbf{\Tilde{y}}^v$) from Eq.~\eqref{eq:pseudo_label} at positions $\omega^r \in \Omega^r$ and $\omega^v \in \Omega^v$.
The class-specific GMMs are defined for both representations by 
\begin{equation}
\resizebox{.75\hsize}{!}{$
\begin{aligned}
    \mathbf{P}_{\Gamma^{y}}(\mathbf{z}|y) = 
      \sum_{m=1}^M \mathbf{P}_{\Gamma^{y}}(m|y) \mathbf{P}_{\Gamma^{y}}(\mathbf{z}|m, y) 
     = \sum_{m=1}^M \ {\boldsymbol{\pi}_{m}^{y}} \mathcal{N}(\textbf{z}; {\boldsymbol{\mu}_{m}^{y}}, {\boldsymbol{\Sigma}_{m}^{y}}),
\end{aligned}
$}
\label{eq:gmm}
\end{equation}
where $\boldsymbol{\pi}^{y}_m=\mathbf{P}_{\Gamma^{y}}(m|y)$ denotes the prior probability for the $m$-th GMM component of class $y \in \{0,1\}^{Y}$, 
$\boldsymbol{\mu}^{y}_m \in \mathbb{R}^Z$ and $\mathbf{\Sigma}^{y}_m \in \mathbb{R}^{Z \times Z}$ represent the mean and covariance of the corresponding GMM component and we fix $M=5$ for all experiments.
The GMM parameter optimisation in~\cref{eq:gmm} relies on the  EM~\cite{dempster1977maximum} algorithm using the training set $\mathcal{G}^{y}$ defined above with the following steps:
\begin{equation}
\label{eq:EM}
\begin{alignedat}{2}
    & \textbf{E step:}  \quad  && \mathbf{q}_{n, m}^{y} = \frac{{\boldsymbol{\pi}}_m^{y} \ \mathcal{N}(\mathbf{z}_n \ | \ {\boldsymbol{\mu}_m^{y}}, {\boldsymbol{\Sigma}_m^{y}})}{\sum_m^M{{\boldsymbol{\pi}_m^{y}} \ \mathcal{N}(\mathbf{z}_n \ | \ {\boldsymbol{\mu}_m^{y}}, {\boldsymbol{\Sigma}_m^{y}})}} \text{, for } (\mathbf{z}_n,\mathbf{c}_n,y_n) \in \mathcal{G}^{y} \\   
     & \textbf{M step:} \quad && {\boldsymbol{\mu}_m^{y}} =  \frac{1}{|\mathcal{G}^{y}|} \sum_{(\mathbf{z}_n,\mathbf{c}_n,y_n) \in \mathcal{G}^{y}} {\mathbf{c}_n\mathbf{q}^{y_n}_{n,m}}\mathbf{z}_n \\
     & \quad && {\boldsymbol{\Sigma}_m^{y}} =  \frac{1}{|\mathcal{G}^{y}|} \sum_{(\mathbf{z}_n,\mathbf{c}_n,y_n) \in \mathcal{G}^{y}} \mathbf{c}_n{\mathbf{q}_{n,m}^{y_n}} (\mathbf{z}_n-{\boldsymbol{\mu}_m^{y_n}}) (\mathbf{z}_n-{\boldsymbol{\mu}_m^{y_n}})^\intercal,
\end{alignedat}
\end{equation}
where $\mathcal{G}^{y}$ represents the set of embedding samples that share the same label $y$.
The posterior $\mathbf{q}_{n,m}^{y}$ estimates the responsibility of the $n^{th}$ embedding $\mathbf{z}$ from class $y$ in $\mathcal{G}^{y}$ being assigned to the $m$-th component.
The confidence $\mathbf{c}_n$ represents the potential correctness of the pseudo label from~\cref{eq:pseudo_label}, which is used as a weighting factor to reduce the effect of low-confidence segmentation results~\cite{cheung2004rival,cheung2005maximum} in the estimation of the GMM parameters.  
Note that we maintain a uniform weight of $\boldsymbol{\pi}_m^{y}=1/M$ for each component~\cite{xu2018benefits, mena2020sinkhorn} to prevent one component from dominating others. \\

\noindent\textbf{GMM-based Prototypical Contrastive Learning.} The class-specific GMMs from Eq.~\eqref{eq:EM} are used to generate a set of highly representative embeddings for the contrastive learning process explained next.
These representative class-specific embeddings is denoted by $\mathcal{S}^{y} = \{ \mathbf{s} | \mathbf{s} \sim \mathbf{P}_{\Gamma^{y}}(\mathbf{z}|y) \}$, with $ \mathbf{P}_{\Gamma^{y}}(\mathbf{z}|y)$ being the class-specific GMM defined in~\cref{eq:gmm}. 
Subsequently, these generated virtual prototypes are utilised to guide the contrastive learning for the incoming embedding features from both range and voxel representations~\cite{wang2021exploring}, without joining the backpropagation computation graph.
We define the set of incoming features with the anchor set
$\mathcal{A} =\{ (\mathbf{z},y) \}$ that is formed similarly to how we form the set $\mathcal{G}^y$ without including the confidence value.
The samples in $\mathcal{A}$ are selectively chosen for backpropagation~\cite{wang2021exploring}, but the informative samples in $\mathcal{S}^{y}$ contribute to a comprehensive understanding of the cross-distribution for both representations.
The contrastive learning relies on the minimisation of the following loss:
\begin{equation}
\label{eq:contras}
\resizebox{\hsize}{!}{$
    \begin{aligned}
         \ell_{\text{cross}}(\mathcal{D}_L, \mathcal{D}_U, f_{\phi_{\textbf{PRJ}}}, f_{\phi_{\textbf{FCN}}}, f_{\theta_{\textbf{PRJ}}}, f_{\theta_{\textbf{EXR}}}  ) = 
        \sum_{(\mathbf{z},y) \in \mathcal{A}}
        \sum_{\mathbf{s} \in \mathcal{S}^y} -\log\frac{\texttt{exp}(\mathbf{z} \cdot \mathbf{s}/\tau)}
        {\texttt{exp}(\mathbf{z} \cdot \mathbf{s}/\tau) + \sum_{\mathbf{s}^{-} \in \bar{\mathcal{S}}^{y} }\texttt{exp}(\mathbf{z} \cdot \mathbf{s}^{-}/\tau)},
    \end{aligned}$}
\end{equation}
where $\bar{\mathcal{S}}^{y}$ is the set of virtual negative embeddings for class $y$.

\subsection{Representation Specific Data Augmentation}
\label{method:aug}
Following~\cite{liu2022perturbed, yang2023revisiting}, we apply spatial data augmentations to the pseudo-labeling strategy in~\cref{eq:pseudo_label} to increase the robustness of our approach to complex scenes. 
After experimenting with classic data augmentation processes (e.g., CutMix~\cite{yun2019cutmix} and LaserMix~\cite{kong2023lasermix}), we notice that the application of distinct spatial augmentations to different viewpoints in our IT2 framework results in more effective generalization. We have empirically found the multi-boxes CutMix~\cite{yun2019cutmix} augmentation for the range image and single-inclination LaserMix~\cite{kong2023lasermix} augmentation for the voxel representation yield the most robust results. 
Although this representation-specific data augmentation is not our main contribution, it is shown empirically to produce relevant improvements. Please see details in supplementary Sec.~\textcolor{red}{2}.

\subsection{Overall Training}

The overall training of our IT2 minimises the following loss function:
\begin{equation}
\begin{aligned}
    & \mathcal{L} (\mathcal{D}_L,  \mathcal{D}_U, \mathcal{D}_L,  \mathcal{D}_U, f_{\phi_{\textbf{FCN}}}, f_{\phi_{\textbf{EXR}}}, f_{\phi_{\textbf{SEG}}}f_{\phi_{\textbf{PRJ}}},f_{\phi_{\textbf{PRJ}}})  = \\
    & \ell_{\text{IT2}}(\mathcal{D}_L,  \mathcal{D}_U, f_{\phi_{\textbf{FCN}}}, f_{\phi_{\textbf{EXR}}}, f_{\phi_{\textbf{SEG}}}) +  \ell_{\text{cross}}(\mathcal{D}_L, \mathcal{D}_U, f_{\phi_{\textbf{PRJ}}}, f_{\phi_{\textbf{FCN}}}, f_{\theta_{\textbf{PRJ}}}, f_{\theta_{\textbf{EXR}}}  ),
\end{aligned}
\end{equation}
where the entire training pipeline is performed in an end-to-end manner, where in each iteration, we minimise the loss from $\ell_{\text{IT2}}(.)$ and $ \ell_{\text{cross}}(.)$ simultaneously.

\vspace{-5pt}

\section{Experiment}
\subsection{Experimental Setup}
\label{sec:exp_setup}

We evaluate our approach on three benchmarks, including  nuScenes~\cite{caesar2020nuscenes}, SemanticKITTI~\cite{behley2019semantickitti} and ScribbleKITTI~\cite{unal2022scribble}. The 
\textbf{nuScenes}~\cite{caesar2020nuscenes} dataset has $29,130$ training scans and $6,019$ validation scans. Each LiDAR scan contains $32$ beams with $[-30\degree, 10\degree]$ vertical field of views (FOV). The dataset has $23$ categories, but following previous works~\cite{kong2023lasermix, zhu2021cylindrical}, we merge them into $16$ for training and validation. The \textbf{SemanticKITTI}~\cite{behley2019semantickitti} and its extension \textbf{ScribbleKITTI}~\cite{unal2022scribble} are popular laser-based segmentation benchmarks, where the latter one is based on line-scribbles labelling to reduce the annotation effort. The SemanticKITTI~\cite{behley2019semantickitti} dataset has $19$ categories for semantic segmentation, where every LiDAR scan contains $64$ beams with $[3\degree, -25\degree]$ vertical FOV. There are $22$ sequences of LiDAR scans in the dataset, where sequences 0 to 9 are used for training and validation. Following~\cite{kong2023lasermix,jiang2021guided}, we use $19,130$ scans (sequences 0 to 7 and sequence 9) for training and $4,071$ scans from sequence 8 for validation.

\noindent \textbf{Sampling strategies.} Previous works~\cite{kong2023lasermix,jiang2021guided,li2023less} propose different sampling strategies to select the labelled data for the semi-supervised learning setting, including: \textit{1)} \ul{uniform}~\cite{kong2023lasermix} sampling of the labelled data across all the training sequences, 
\textit{2)} \ul{partial}~\cite{jiang2021guided} sampling of the labelled data only from certain sequences, and \textit{3)} sampling of \ul{significant}~\cite{li2023less} labelled frames from the sequences. In the experiment, we examine our approach under all these sampling protocols.\\
\textbf{Metrics.} We measure performance with mean intersection over union (mIoU). In each benchmark, we use the sampling strategies above to label $n\%$ of the training set, with the rest $(100-n)\%$ being unlabelled.\\
\textbf{Implementation details.}  In the range view, we use the FIDNet architecture~\cite{zhao2021fidnet} with ResNet34~\cite{he2016deep} backbone. The input range image has size $32\times 1920$ for nuScenes~\cite{caesar2020nuscenes} and $64\times2048$ for SemanticKITTI~\cite{behley2019semantickitti} and ScirbbleKITTI~\cite{unal2022scribble}, following~\cite{kong2023lasermix}. In the voxel representation, we use the Cylinder3D architecture~\cite{zhu2021cylindrical}, where we fix the intermediate layer at $16$ dimensions with smaller resolution as specified in~\cite{kong2023lasermix} for the uniform sampling experiments to enable a fair comparison.
For contrastive learning, we build a $3$-layer projector~\cite{wang2021exploring,khosla2020supervised}. The projector outputs an embedding map with $64$ channels, where we apply the temperature factor $\tau=.1$ in Eq.~\eqref{eq:contras}. See supplementary Sec. \textcolor{red}{3} and Sec. \textcolor{red}{7} for more details.
\begin{table*}[t!]
\centering
\renewcommand{\arraystretch}{1.1}
\caption{\textbf{Uniform sampling results} across various benchmarks, where we follow~\cite{kong2023lasermix} to utilise half feature dimension and lower resolution in Cylinder3D~\cite{zhu2021cylindrical} for all the experiments. Best results are highlighted in \textcolor{darkred}{red}.}
\vspace{-5pt}
\label{tab:uniform}
\resizebox{\textwidth}{!}{
\begin{tabular}{c|c?c|c|c|c?c|c|c|c?c|c|c|c}
\boldline
\multirow{2}{*}{Repr.} & \multicolumn{1}{c?}{\multirow{2}{*}{Method}} & \multicolumn{4}{c?}{nuScenes~\cite{caesar2020nuscenes}}   & \multicolumn{4}{c?}{SemanticKITTI~\cite{behley2019semantickitti}} & \multicolumn{4}{c}{ScribbleKITTI~\cite{unal2022scribble}} \\
\cline{3-14}
                       &                         & 1\%  & 10\% & 20\% & 50\% & 1\%    & 10\%  & 20\% & 50\% & 1\%   & 10\%  & 20\% & 50\% \\
\boldline
                       \cellcolor{lightred}& sup.                &  38.3   &  57.5   &  62.7    &  67.6     &  36.2    &  52.2    &  55.9    &  57.2      &  33.1    &  47.7    &  49.9    &   52.5           \\
                       
\cline{2-14}
                       \cellcolor{lightred}& MT~\cite{tarvainen2017mean}                  &  42.1    &  60.4    &  65.4    &   69.4     & 37.5       & 53.1      &   56.1     &  57.4    & 34.2    &  49.8      & 51.6     &   53.3      \\
                       \cellcolor{lightred}& CBST~\cite{zou2018unsupervised}                &  40.9    &  60.5    &  64.3    &   69.3     & 39.9       & 53.4      &   56.1     &  56.9    & 35.7    &  50.7      & 52.7     &   54.6      \\
                       \cellcolor{lightred}& CPS~\cite{chen2021semi}                 &  40.7    &  60.8    &  64.9    &   68.0     & 36.5       & 52.3      &   56.3     &  57.4    & 33.7    &  50.0      & 52.8     &   54.6       \\
                       \cellcolor{lightred}& LaserMix~\cite{kong2023lasermix}             &  49.5    &  68.2    &  70.6    &   73.0     & 43.4       & 58.8      &   59.4     &  61.4    & 38.3    &  54.4      & 55.6     &   \textcolor{darkred}{58.7}        \\
                       
\cline{2-14}
                       \cellcolor{lightred}{\multirow{-5}{*}[2.4ex]{\rotatebox[origin=c]{90}{\centering \textit{Range}}}}
                       &  \multicolumn{1}{c?}{{IT2} {\scriptsize \textcolor{gray}{(Ours)}}}                          &  \textcolor{darkred}{56.5}    &  \textcolor{darkred}{71.3}    &  \textcolor{darkred}{73.4}    &  \textcolor{darkred}{74.0}    &  \textcolor{darkred}{51.9}     &   \textcolor{darkred}{60.3}   &  \textcolor{darkred}{61.7}    &  \textcolor{darkred}{62.1}    &  \textcolor{darkred}{46.6}    &  \textcolor{darkred}{57.1}     & \textcolor{darkred}{57.3}     &  
                       58.6
                       \\

\boldline

                       \cellcolor{lightblue}& sup.                   &   50.9    & 65.9     & 66.6     & 71.2     &  45.4     &  56.1    &  57.8    & 58.7     &  39.2    &  48.0     &  52.1    &  53.8    \\
                       
\cline{2-14}
                       \cellcolor{lightblue}& MT~\cite{tarvainen2017mean}                  &  51.6    &  66.0    &  67.1   &  71.7  &  45.4  &  57.1  &  59.2  &  60.0  &  41.0  &  50.1  &  52.8  &   53.9                                  \\
                       \cellcolor{lightblue}& CBST~\cite{zou2018unsupervised}               &  53.0    &  66.5    &  69.6   &  71.6  &  48.8  &  58.3  &  59.4  &  59.7  &  41.5  &  50.6  &  53.3  &   54.5                                   \\
                       \cellcolor{lightblue}& CPS~\cite{chen2021semi}               &  52.9    &  66.3    &  70.0   &  72.5  &  46.7  &  58.7  &  59.6  &  60.5  &  41.4  &  51.8  &  53.9  &   54.8                                   \\
                       \cellcolor{lightblue}& LaserMix~\cite{kong2023lasermix}          &  55.3    &  69.9    &  71.8   &  73.2  &  50.6  &  60.0  &  61.9  &  62.3  &  44.2  &  53.7  &  55.1  &   56.8                                   \\
\cline{2-14}
                       \cellcolor{lightblue}{\multirow{-6}{*}{\rotatebox[origin=c]{90}{\textit{Voxel}}}}
                       & \multicolumn{1}{c?}{{IT2} {\scriptsize \textcolor{gray}{(Ours)}}}                           & \textcolor{darkred}{57.5}     &  \textcolor{darkred}{72.1}    &  \textcolor{darkred}{73.6}    &  \textcolor{darkred}{74.1}    &  \textcolor{darkred}{52.0}     &  \textcolor{darkred}{61.4}    &  \textcolor{darkred}{62.1}    &  \textcolor{darkred}{62.5}   &   \textcolor{darkred}{47.9}    &  \textcolor{darkred}{56.7}     &   \textcolor{darkred}{57.5}   &  \textcolor{darkred}{58.3}   \\

\boldline
\end{tabular}}
\end{table*}
\vspace{-10pt}
\subsection{Results on Different Partition Protocols}
\label{sec:exp_sota}

\begin{table}[t!]
\renewcommand{\arraystretch}{1.1}
\caption{\textbf{Partial \& significant sampling results} in the SemanticKITTI~\cite{behley2019semantickitti} benchmark.  We compare with previous SOTA based on Cylinder3D~\cite{zhu2021cylindrical} architecture, where \ul{TTA} denotes test time augmentation. $\dagger$ indicates that we re-implement results based on the supported checkpoints from their GitHub -- please see Supplementary Sec. \textcolor{red}{1} for more details. Best results in each evaluation protocol are highlighted in \textcolor{darkred}{red}.}
\vspace{-5pt}
\label{tab:partial}
\centering
\resizebox{.9\linewidth}{!}{
\begin{tabular}{c|c?c|c|c|c|c?c|c|c|c}
\boldline
\multirow{2}{*}{TTA} & \multirow{2}{*}{Method} & \multicolumn{4}{c|}{SemanticKITTI{\scriptsize \textcolor{red}{(par.)}}} & \multirow{2}{*}{Method} & \multicolumn{4}{c}{SemanticKITTI{\scriptsize \textcolor{red}{(sig.)}}}\\
\cline{3-6}
\cline{8-11}
                      &                         & 5\%    & 10\%    & 20\%     & 40\%  &  & 5\%    & 10\%    & 20\%    & 40\%  \\
\boldline
\multirow{2}{*}{\xmark}   & GPC$^\dagger$~\cite{jiang2021guided}                       &    39.9    &   45.9      &  55.7           &   57.1   &lim3D~\cite{li2023less} & 56.8 & 59.6 & 60.5 & 61.4 \\
\cline{2-11}
               &   IT2 {\scriptsize \textcolor{gray}{(Ours)}}                   &  \textcolor{darkred}{43.4}      &   \textcolor{darkred}{48.1}      &   \textcolor{darkred}{56.5}            &  \textcolor{darkred}{60.9}  & IT2 {\scriptsize \textcolor{gray}{(Ours)}} &\textcolor{darkred}{60.3} &\textcolor{darkred}{63.3} & \textcolor{darkred}{64.0} & \textcolor{darkred}{64.8} \\
\boldline
\multirow{2}{*}{\textcolor{red}{\cmark}}   & GPC$^\dagger$~\cite{jiang2021guided}                      &    40.2    &   46.3      &  56.5          &   57.6  & lim3D~\cite{li2023less} & 59.5 & 62.2 & 63.1 & 63.3 \\
\cline{2-11}
               &   IT2 {\scriptsize \textcolor{gray}{(Ours)}}                   &    \textcolor{darkred}{43.8}    &  \textcolor{darkred}{49.2}   &    \textcolor{darkred}{58.8}    &  \textcolor{darkred}{62.4}      &     IT2 {\scriptsize \textcolor{gray}{(Ours)}}  & \textcolor{darkred}{61.7} & \textcolor{darkred}{64.4} & \textcolor{darkred}{64.7} & \textcolor{darkred}{65.8} \\
\boldline
\end{tabular}}
\vspace{-10pt}
\end{table}

\textbf{Uniform sampling results.} We first compare our method using the uniform sampling benchmark~\cite{kong2023lasermix} in Tab.~\ref{tab:uniform}, with the experimental setup from~\cite{kong2023lasermix} (i.e., architecture, split ID and the resolution of input representations). Our framework outperforms the previous SOTA Lasermix~\cite{kong2023lasermix} by a large margin for almost all partition protocols. For example, we observe $\approx 3.4\%$ and $\approx 2.0\%$ improvements on average for range and voxel representations, respectively. Notably, we highlight that our approach yields significant improvements for the $1\%$ partition protocol in the line-scribble dataset~\cite{unal2022scribble}, providing $8.28\%$ improvement in the range image and $3.66\%$ in the voxel representation.\\
\textbf{Partial and significant sampling results.} In Tab.~\ref{tab:partial}, we compare our results with GPC~\cite{jiang2021guided} and lim3D~\cite{li2023less} in partial sampling (par.) and significant sampling (sig.), based on both original performance and test time augmentation (TTA). 
In the \textbf{partial sampling} protocol
\footnote{\protect The original GPC paper~\cite{jiang2021guided} calculates the mIoU differently compared to common practices~\cite{kong2023lasermix, zhu2021cylindrical, li2023less}. The reported performance is based on the re-implementation from the official supported checkpoints of their GitHub, where more details are in Supplementary Sec. \textcolor{red}{1}.}, 
our approach yields superior results across all the labelled data situations for both with and without TTA. Notably, our results surpass GPC~\cite{jiang2021guided} by 4.5\% and 3.6\% in the limited 5\% labelled data condition, demonstrating its good generalisation capability. In the \textbf{significant sampling} results, our method also achieves better performance than lim3D~\cite{li2023less} for all the partition protocols. For example, our results are better than the previous SOTA by about 2.2\% and 1.6\% in 10\% and 20\% labelled data with TTA, respectively. 
\vspace{-8pt}

\subsection{Ablation Study}
\label{sec:exp_ablation}
\textbf{Improvements from our contributions.} Table~\ref{tab:ablation} presents our ablation study on the nuScenes~\cite{caesar2020nuscenes} and ScribbleKITTI~\cite{unal2022scribble} datasets across various partition protocols. We investigate the roles of the IT2 architecture, cross-distribution contrastive learning (Ctrs.), and representation-related data augmentation (Aug.). 
The baseline utilises network perturbation through single representations~\cite{chen2021semi} from~\cite{kong2023lasermix}. Our IT2 architecture results in improvements of approximately $4.1\%$ and $2.5\%$ for the voxel representation in 10\% labelled data for nuScenes~\cite{caesar2020nuscenes} and ScribbleKITTI~\cite{unal2022scribble}, respectively.
The Ctrs. further improves by $\approx 2.4\%$ and $\approx 1.1\%$ for range and voxel representations in both datasets in 10\% labelled data. Such enhancements show the effectiveness of our cross-distribution prototypes and contrastive learning strategy. Furthermore, Aug. results in improvements of $2.2\%$ and $1.4\%$ for range and voxel representations in the ScribbleKITTI~\cite{unal2022scribble} dataset for 10\% labelled data, demonstrating the better generalisation brought by the representation-specific data augmentation.

\begin{table}[t!]

\renewcommand{\arraystretch}{1.1}
\caption{\textcolor{darkred}{Ablation study} in the nuScenes~\cite{caesar2020nuscenes} and ScribbleKITTI~\cite{unal2022scribble} datasets via uniform sampling. IT2 represents our architecture, Ctrs. denotes our cross-distribution contrastive learning and Aug. is our representation-specific data augmentation. The \colorbox{lightyellow}{yellow rows} denote our single representation baseline with CPS~\cite{chen2021semi} from~\cite{kong2023lasermix}.}
\vspace{-5pt}
\label{tab:ablation}
\centering
\resizebox{\linewidth}{!}{
\begin{tabular}{c|ccc?c|c|c?c|c|c}
\boldline
\multirow{2}{*}{Repr.} & \multirow{2}{*}{IT2} & \multirow{2}{*}{Ctrs.} & \multirow{2}{*}{Aug.} & \multicolumn{3}{c?}{nuScenes~\cite{caesar2020nuscenes}} & \multicolumn{3}{c}{ScribbleKITTI~\cite{unal2022scribble}}\\
\cline{5-10}
                     &                      &                        &                          & 10\%    & 20\%    & 50\%  & 10\%    & 20\%    & 50\%   \\
\boldline
                     \cellcolor{lightred}& \xmark &    \xmark &       \xmark                         &    \cellcolor{lightyellow}  60.8  &   \cellcolor{lightyellow} 64.9    &  \cellcolor{lightyellow} 68.0  & \cellcolor{lightyellow} 50.0  & \cellcolor{lightyellow} 52.8 & \cellcolor{lightyellow} 54.6   \\
\cdashline{2-10}
                     \cellcolor{lightred}& \cmark &    \xmark &       \xmark                         &  67.3 {\tiny \textcolor{gray}{(6.5$\uparrow$)}} &   70.8  {\tiny \textcolor{gray}{(5.9$\uparrow$)}}         &   72.6 {\tiny \textcolor{gray}{(4.6$\uparrow$)}} & 53.2 {\tiny \textcolor{gray}{(3.2$\uparrow$)}} & 55.0 {\tiny \textcolor{gray}{(2.2$\uparrow$)}} & 56.2 {\tiny \textcolor{gray}{(1.6$\uparrow$)}} \\
                     \cellcolor{lightred}& \cmark &    \cmark &       \xmark                            & 70.3 {\tiny \textcolor{gray}{(9.5$\uparrow$)}}  &  72.6 {\tiny \textcolor{gray}{(7.7$\uparrow$)}}          &  73.5 {\tiny \textcolor{gray}{(5.5$\uparrow$)}} & 54.9 {\tiny \textcolor{gray}{(4.9$\uparrow$)}} & 56.1 {\tiny \textcolor{gray}{(3.3$\uparrow$)}} & 57.2 {\tiny \textcolor{gray}{(2.6$\uparrow$)}}\\
                     \cellcolor{lightred}{\multirow{-2.5}{*}[2.4ex]{\rotatebox[origin=c]{90}{\centering \textit{Range}}}} & \cmark &    \cmark                  &        \cmark                          &  \textcolor{darkred}{71.3} {\tiny \textcolor{gray}{(10.5$\uparrow$)}}     &  \textcolor{darkred}{73.4}  {\tiny \textcolor{gray}{(8.5$\uparrow$)}}   &  \textcolor{darkred}{74.0} {\tiny \textcolor{gray}{(6.0$\uparrow$)}} & \textcolor{darkred}{57.1} {\tiny \textcolor{gray}{(7.1$\uparrow$)}} & \textcolor{darkred}{57.3} {\tiny \textcolor{gray}{(4.5$\uparrow$)}} & \textcolor{darkred}{58.6} {\tiny \textcolor{gray}{(4.0$\uparrow$)}} \\
\boldline
                     \cellcolor{lightblue}& \xmark &    \xmark &       \xmark                          & \cellcolor{lightyellow} 66.3       &  \cellcolor{lightyellow} 70.0      & \cellcolor{lightyellow} 72.5  & \cellcolor{lightyellow} 51.8   & \cellcolor{lightyellow} 53.9 & \cellcolor{lightyellow} 54.8 \\
\cdashline{2-10}
                     \cellcolor{lightblue}& \cmark &    \xmark &       \xmark                           &    70.4 {\tiny \textcolor{gray}{(4.1$\uparrow$)}}  &  71.9  {\tiny \textcolor{gray}{(1.9$\uparrow$)}}   &    73.1 {\tiny \textcolor{gray}{(0.6$\uparrow$)}} & 54.3 {\tiny \textcolor{gray}{(2.5$\uparrow$)}} & 55.6 {\tiny \textcolor{gray}{(2.7$\uparrow$)}} & 56.6 {\tiny \textcolor{gray}{(1.8$\uparrow$)}} \\
                     \cellcolor{lightblue}& \cmark &    \cmark &       \xmark                           &    71.6 {\tiny \textcolor{gray}{(5.3$\uparrow$)}} &  72.9  {\tiny \textcolor{gray}{(2.9$\uparrow$)}}   &     73.9 {\tiny \textcolor{gray}{(1.4$\uparrow$)}} & 55.3 {\tiny \textcolor{gray}{(3.5$\uparrow$)}} & 56.5 {\tiny \textcolor{gray}{(3.6$\uparrow$)}} & 57.6 {\tiny \textcolor{gray}{(2.8$\uparrow$)}} \\
                     \cellcolor{lightblue}{\multirow{-2.5}{*}[2.4ex]{\rotatebox[origin=c]{90}{\centering \textit{Voxel}}}} & \cmark &    \cmark                  &        \cmark                           &    \textcolor{darkred}{72.1} {\tiny \textcolor{gray}{(5.8$\uparrow$)}}   &   \textcolor{darkred}{73.6} {\tiny \textcolor{gray}{(3.6$\uparrow$)}}   &  \textcolor{darkred}{74.1} {\tiny \textcolor{gray}{(1.6$\uparrow$)}} & \textcolor{darkred}{56.7} {\tiny \textcolor{gray}{(4.9$\uparrow$)}} & \textcolor{darkred}{57.5} {\tiny \textcolor{gray}{(4.6$\uparrow$)}} & \textcolor{darkred}{58.3} {\tiny \textcolor{gray}{(3.5$\uparrow$)}} \\
\boldline
\end{tabular}
}
\vspace{-5pt}
\end{table}

\begin{figure}[t!]
 \begin{minipage}{\textwidth}
    \begin{minipage}[t]{.49\linewidth}
    \begin{flushleft}
  \vspace{-2pt}
    \includegraphics[width=\textwidth]{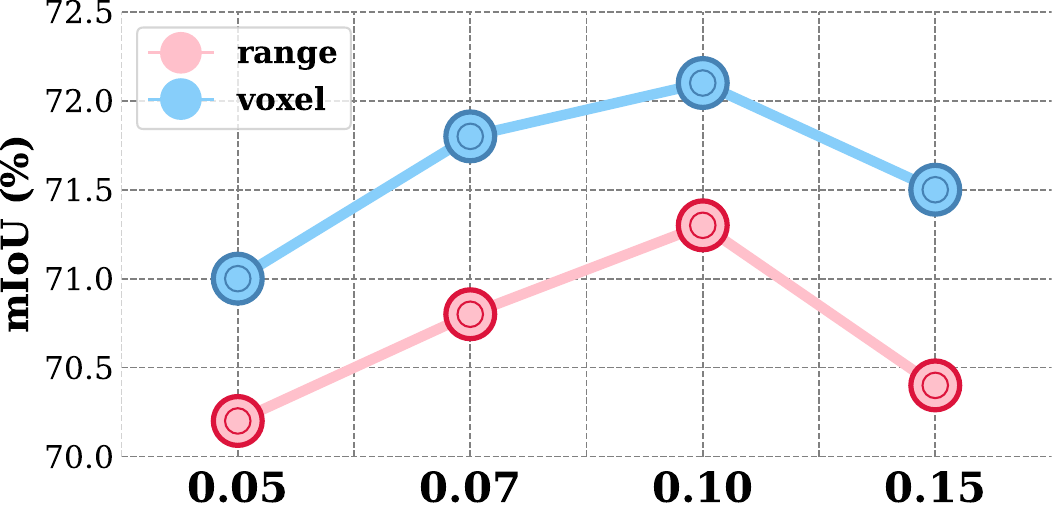}
    \vspace{-15pt}
    \captionof{figure}{\textcolor{darkred}{Ablation study of temperature value $\tau$} from Eq.~\eqref{eq:contras} in  nuScenes~\cite{caesar2020nuscenes} dataset under 10\% labelled partition protocol.}
    \label{fig:ablation_contrastive}
    \end{flushleft}
    \end{minipage}
  \begin{minipage}[t]{0.49\linewidth}
  \vspace{-12pt}
    \captionof{table}{\textcolor{darkred}{Contrastive Learning study} in  nuScenes~\cite{caesar2020nuscenes}, where ContrasSeg~\cite{wang2021exploring} is re-implemented from the SOTA image (pixel-level) semantic segmentation.}
    \renewcommand{\arraystretch}{1.1}
    \label{tab:ablation_contrastive}
    \resizebox{\textwidth}{!}{
    \begin{tabular}{c|c?c|c}
    \boldline
    \multirow{2}{*}{Method}                      & \multirow{2}{*}{Repr.} & \multicolumn{2}{c}{nuScenes~\cite{caesar2020nuscenes}} \\
    \cline{3-4}
                                                 &                        & 10\%          & 20\%                               \\
    \boldline
    \multirow{2}{*}{ContrasSeg~\cite{wang2021exploring}}        & range                &   70.3              &     72.3                              \\
                                                 & voxel                  &    71.2           & 72.8                                   \\
    \hline
    \multirow{2}{*}{Ours} & range                  &  71.3 {\tiny \textcolor{gray}{(1.0$\uparrow$)}}           &     73.4 {\tiny \textcolor{gray}{(1.1$\uparrow$)}}                                \\
                                                 & voxel                  &      72.1 {\tiny \textcolor{gray}{(0.9$\uparrow$)}}        &     73.6 {\tiny \textcolor{gray}{(0.8$\uparrow$)}}            \\
    \boldline
    \end{tabular}}
    \end{minipage}
    \vspace{-15pt}
\end{minipage}
\end{figure}
 
\begin{figure}[t!]
\begin{minipage}[t]{.5\linewidth}
\vspace{-1pt}
\includegraphics[width=\textwidth]{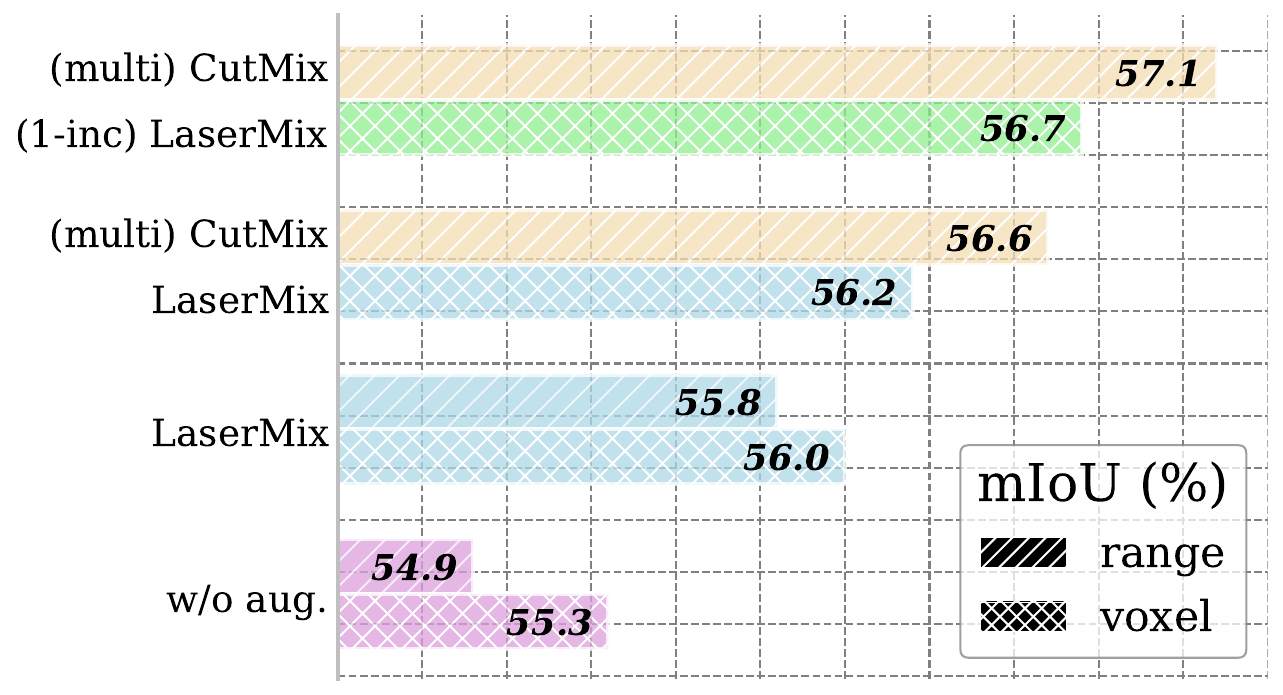}
\captionof{figure}{\textcolor{darkred}{Ablation study} for the repr.-specific augmentations in ScribbleKITTI~\cite{unal2022scribble} with 10\% labelled data, where 'multi' is multi-boxes and '1-inc' is single inclination.}
\label{fig:aug}
\end{minipage}
\begin{minipage}[t]{.5\linewidth}
\renewcommand{\arraystretch}{1.25}
\vspace{-10pt}
\captionof{table}{\textcolor{darkred}{Ablation study} for the \textit{bird's eye view} (BEV) experiment in the nuScenes~\cite{caesar2020nuscenes}, where \textbf{results} are from PolarNet~\cite{zhang2020polarnet}. The \colorbox{lightyellow}{baseline} rely on single view~\cite{chen2021semi}, followed by BEV with \textcolor{rangered}{range image} or \textcolor{voxelblue}{voxel grid}.}
\label{tab:polar}
\resizebox{\linewidth}{!}{
\begin{tabular}{l?c|c|c}
\boldline
\multicolumn{1}{c?}{\multirow{2}{*}{Repr.}} & \multicolumn{3}{c}{BEV (PolarNet~\cite{zhang2020polarnet})}  \\
\cline{2-4}
                     & 10\%         & 20\%        & 50\%                \\
\hline
\rowcolor{lightgray!20}
\multicolumn{1}{c?}{{sup.}}              &  58.5         &  

 63.9      &     68.4           \\
 
\rowcolor{lightyellow} \hspace{.9pt} w/ bev            &     62.4   {\tiny \textcolor{gray}{(3.9$\uparrow$)}}  &   66.7  {\tiny \textcolor{gray}{(2.8$\uparrow$)}}  &  70.0 {\tiny \textcolor{gray}{(1.6$\uparrow$)}}          \\
 \hdashline
\cellcolor{lightred} w/ range         &       64.8  {\tiny \textcolor{gray}{(6.3$\uparrow$)}}   &    67.9   {\tiny \textcolor{gray}{(4.0$\uparrow$)}}    &    70.6  {\tiny \textcolor{gray}{(2.2$\uparrow$)}}    \\
\cellcolor{lightblue} w/ voxel     &        66.3 {\tiny \textcolor{gray}{(7.8$\uparrow$)}}   &    69.1  {\tiny \textcolor{gray}{(5.2$\uparrow$)}}   &      71.6  {\tiny \textcolor{gray}{(3.2$\uparrow$)}}           \\
\boldline          
\end{tabular}
}
\end{minipage}
\vspace{-5pt}
\end{figure}

\noindent\textbf{Cross-distribution contrastive learning.} We investigate the impact of the hyperparameter $\tau$ from Eq.~\eqref{eq:contras} in Fig.~\ref{fig:ablation_contrastive} using the 10\% labelled protocol on the nuScenes~\cite{caesar2020nuscenes} dataset. Our observations indicate that $\tau=0.10$ yields the best results, so we fix this value for all experiments. In Tab.~\ref{tab:ablation_contrastive}, we compare our approach with our re-implementation of the label-guided pixel-wise contrastive learning~\cite{wang2021exploring} based on our IT2 architecture with different ratio of labelled data, where our method shows  $\approx 1\%$ improvement for both representations, demonstrating the effectiveness of our proposed cross-distribution contrastive learning.\\
\noindent\textbf{Representation-specific augmentation.} In Fig.~\ref{fig:aug}, we have demonstrated the augmentation results for each representation in the ScribbleKITTI\cite{unal2022scribble} dataset, where 'multi' denotes multi-box~\cite{yun2019cutmix}, and '1-inc' represents single inclination setups~\cite{kong2023lasermix}. In our framework, one representation's results rely on the other's generalisation. For instance, '(multi) CutMix' reaches 57.1\% mIoU with '(1-inc) LaserMix,' improving by 0.5\% compared to the results with 'LaserMix.' Meanwhile, '(multi) CutMix' combined with '(1-inc) LaserMix' improves by 1.3\% in the range and 0.7\% in voxel representations compared to LaserMix~\cite{kong2023lasermix}, highlighting the improved generalisation benefits from our augmentation.\\
\noindent\textbf{Ablation study for other representations.} 
In Tab.~\ref{tab:polar}, we conduct ablation studies with an additional representation, the Bird's Eye View (BEV), on the nuScenes~\cite{caesar2020nuscenes} dataset, where \ul{all the results} in the table are based on PolarNet~\cite{zhang2020polarnet}.
We re-implement the supervised results and single representation supervision~\cite{chen2021semi} in the \colorbox{lightgray!20}{first} and \colorbox{lightyellow}{second} rows, respectively. We emphasize that employing peer representations can significantly enhance performance compared to the single-representation baseline~\cite{chen2021semi}. For instance, under the 10\% labelled partition protocol, employing BEV (based on  PolarNet~\cite{zhang2020polarnet}) with the \textcolor{rangered}{range image} (based on FidNet~\cite{zhao2021fidnet}) and with the \textcolor{voxelblue}{voxel grid} (based on Cylinder3D~\cite{zhu2021cylindrical}) lead to improvements of 2.4\% and 3.9\% in mIoU, respectively.
\vspace{-8pt}
\begin{figure*}[t!]
    \centering
\begin{subfigure}[b]{.24\linewidth}
    \includegraphics[width=\textwidth]{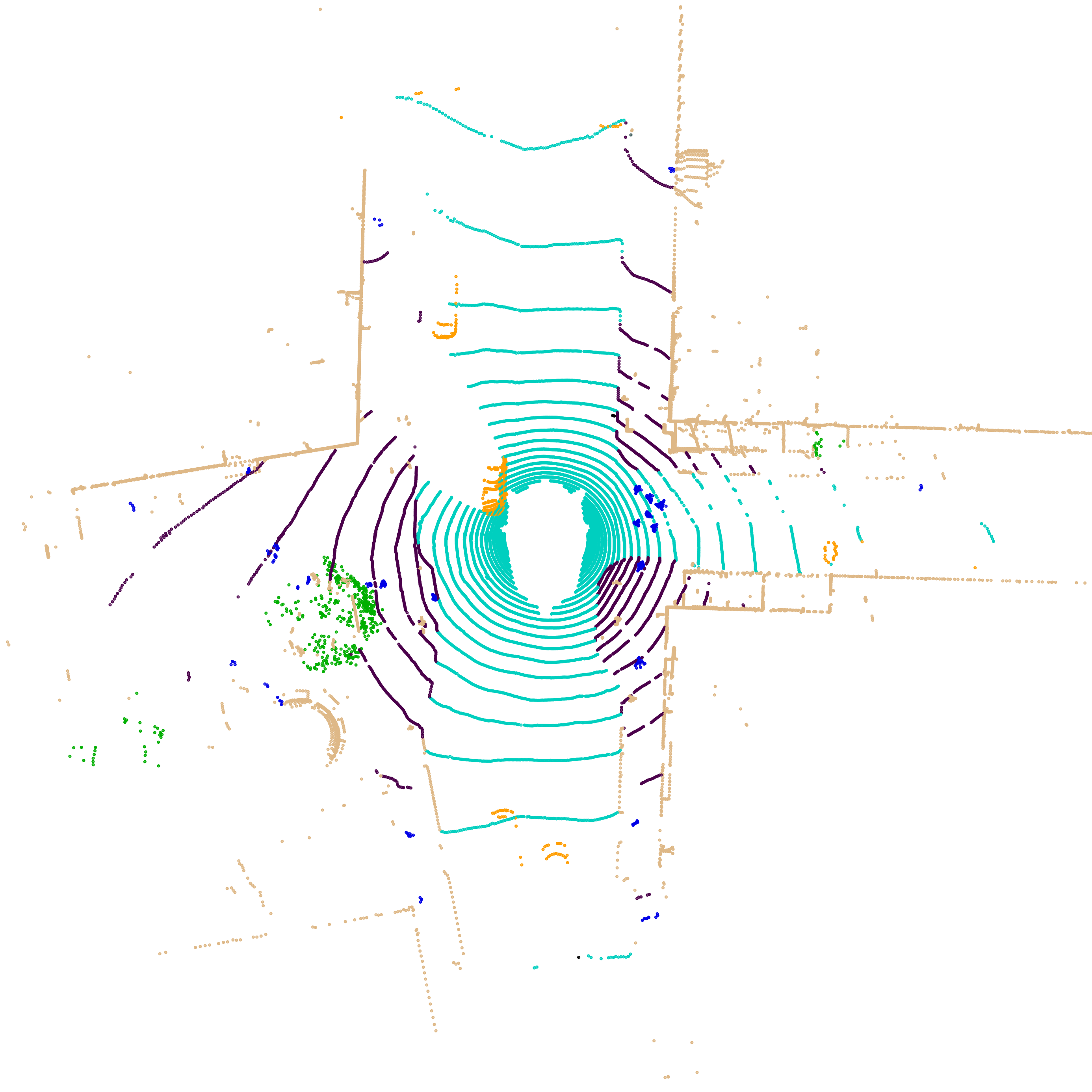}
    \includegraphics[width=\textwidth]{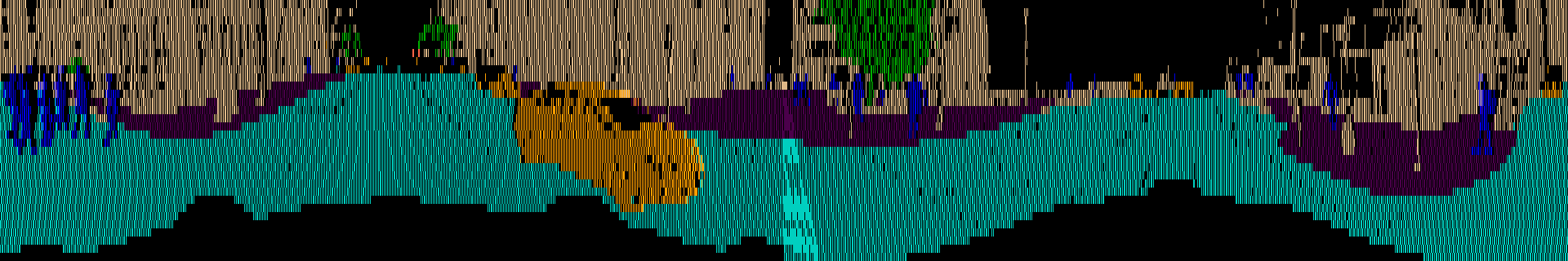}
    \includegraphics[width=\textwidth]{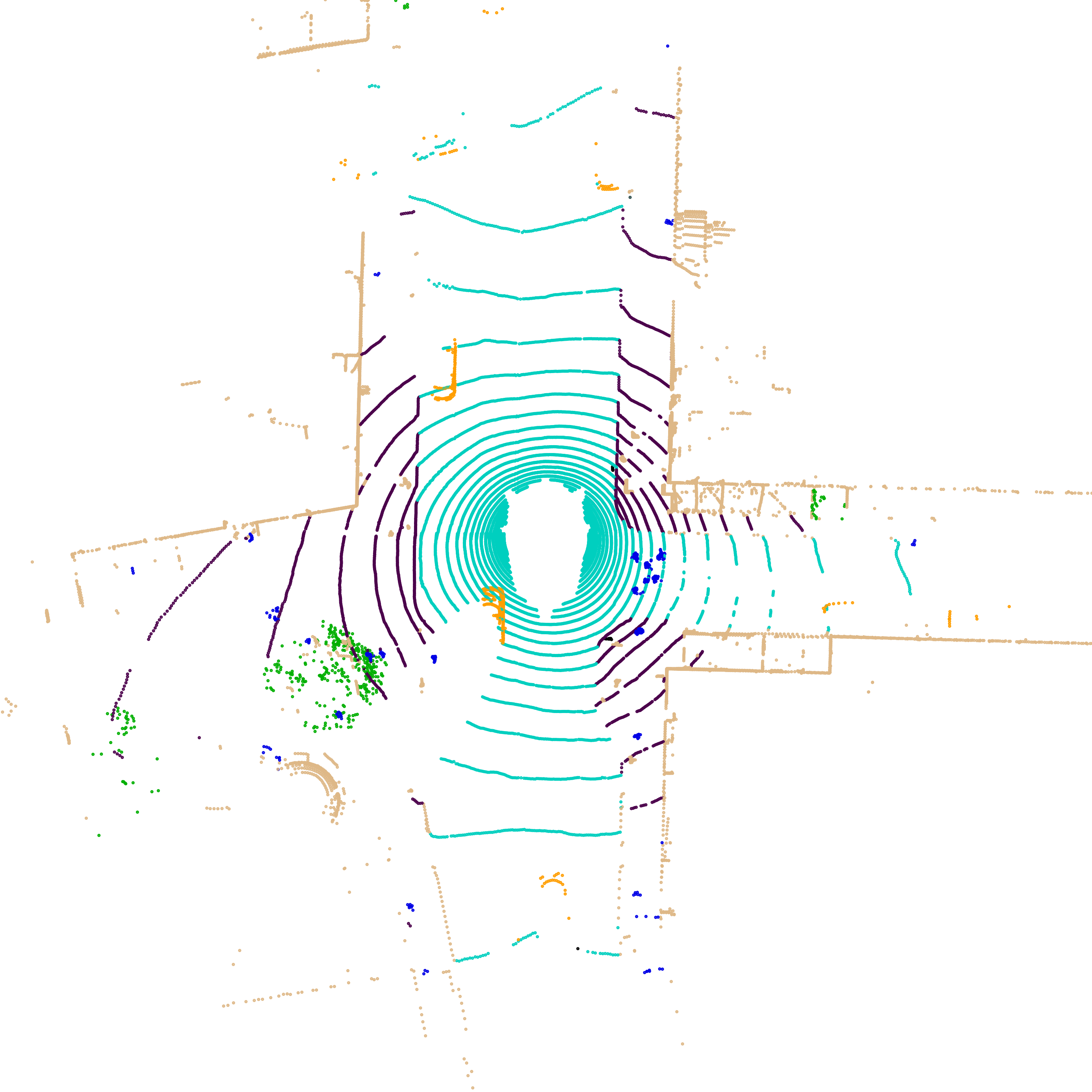}
    \includegraphics[width=\textwidth]{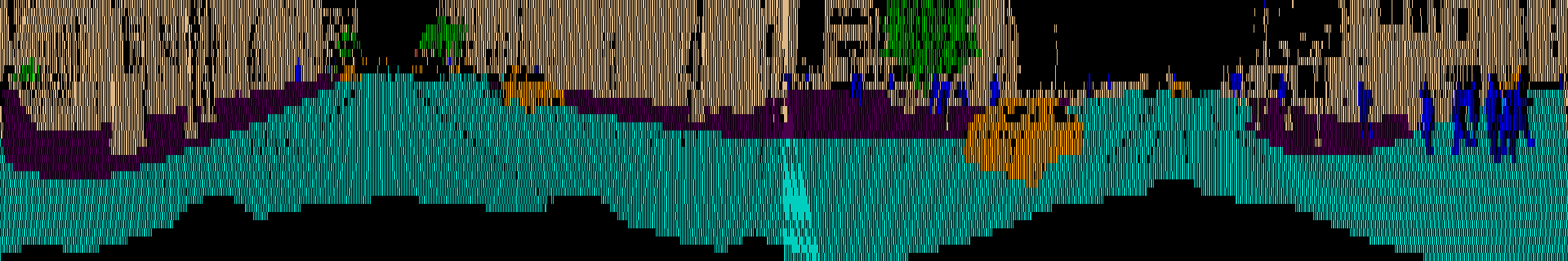}
    \caption{Ground truth}
\end{subfigure}
\begin{subfigure}[b]{.24\linewidth}
    \includegraphics[width=\textwidth]{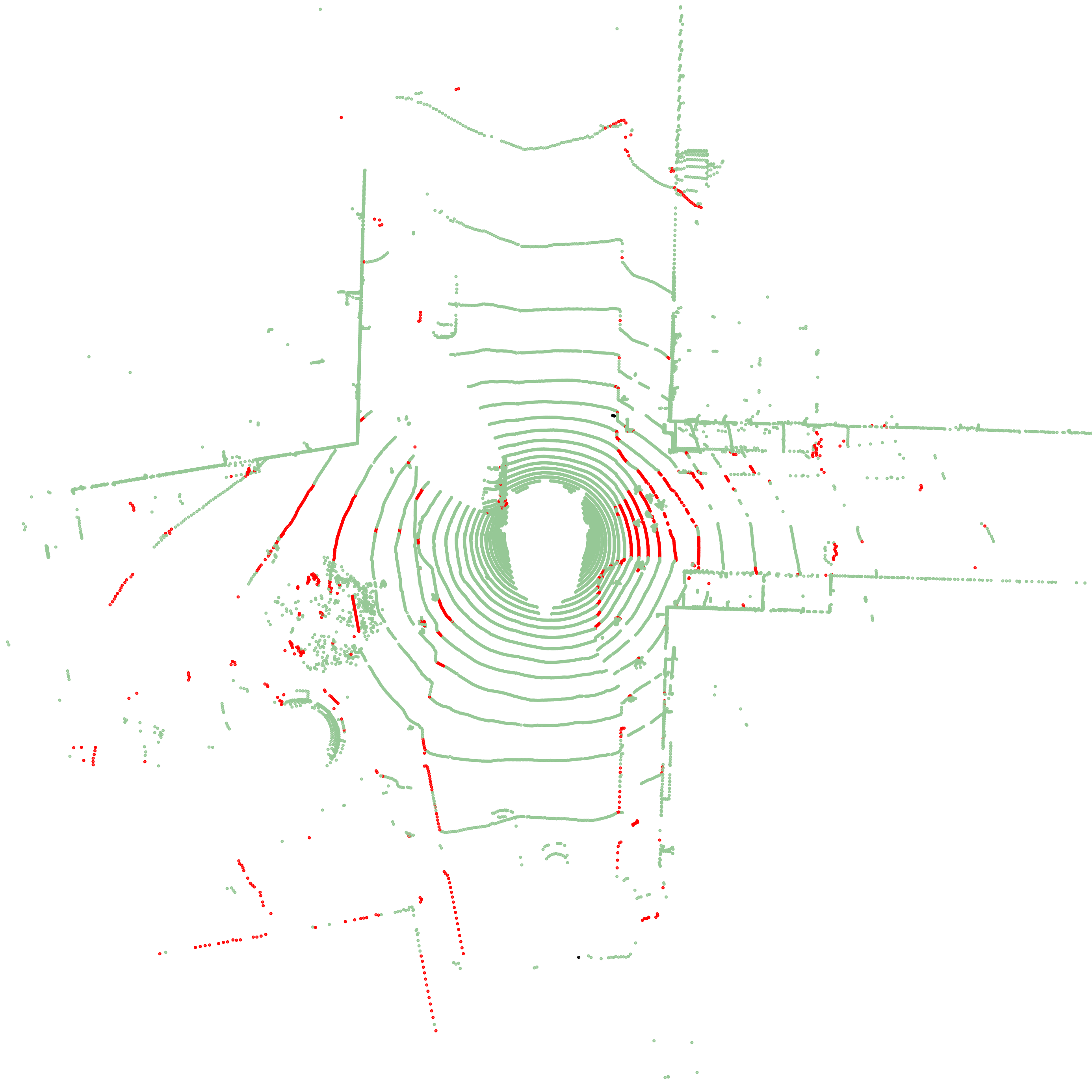}
    \includegraphics[width=\textwidth]{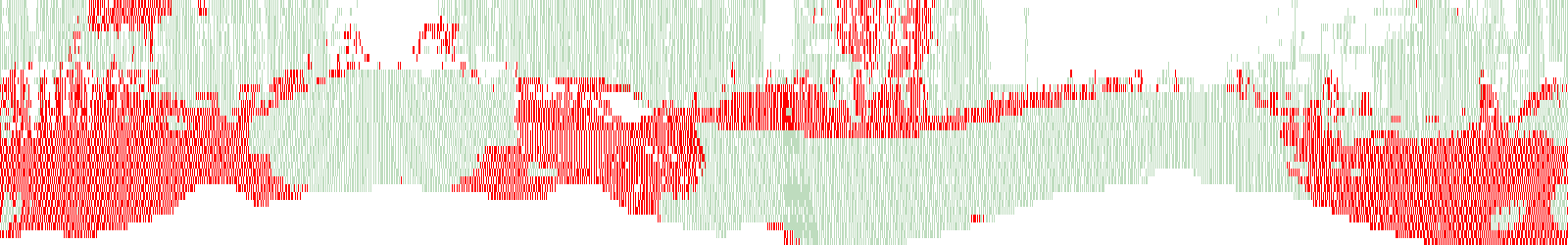}
    \includegraphics[width=\textwidth]{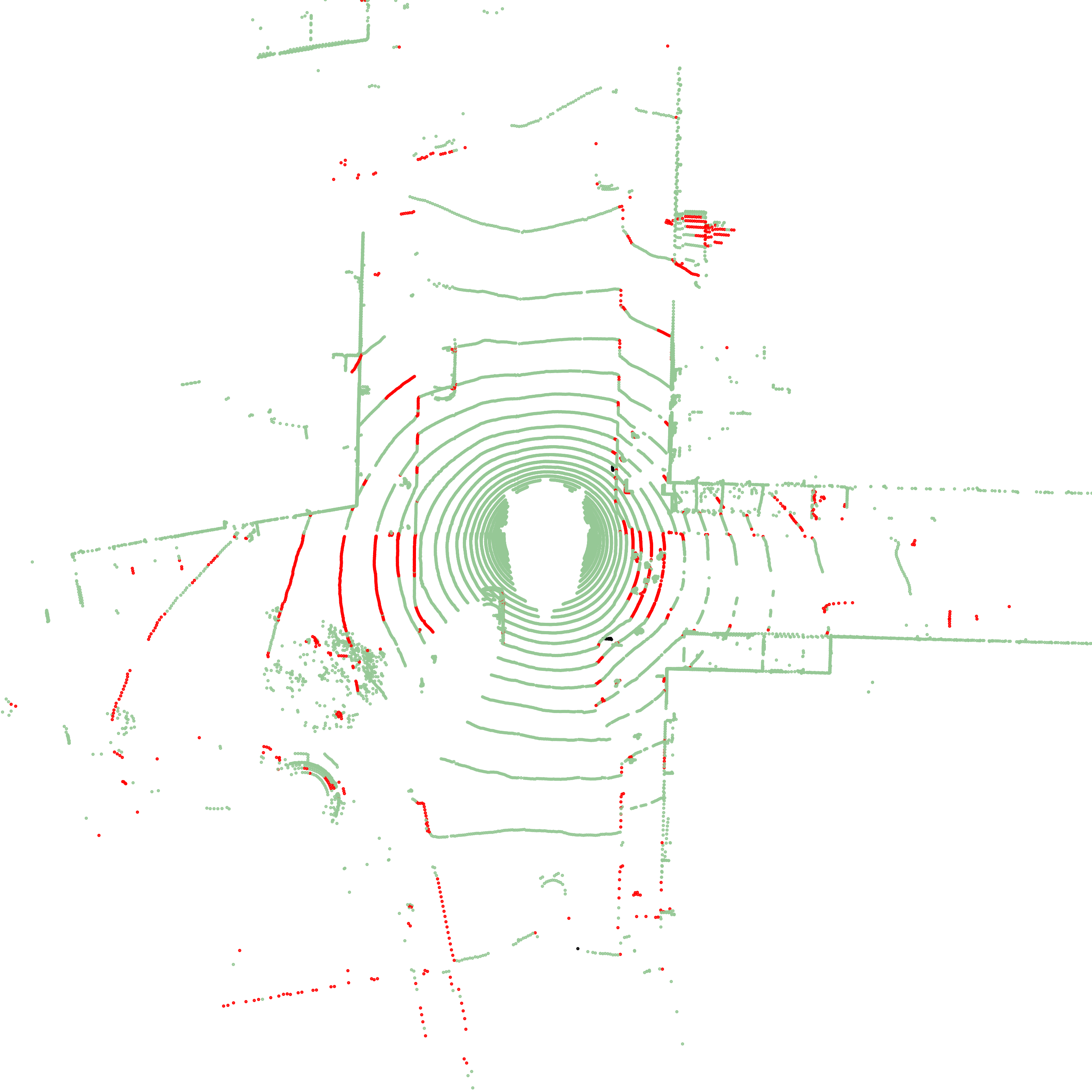}
    \includegraphics[width=\textwidth]{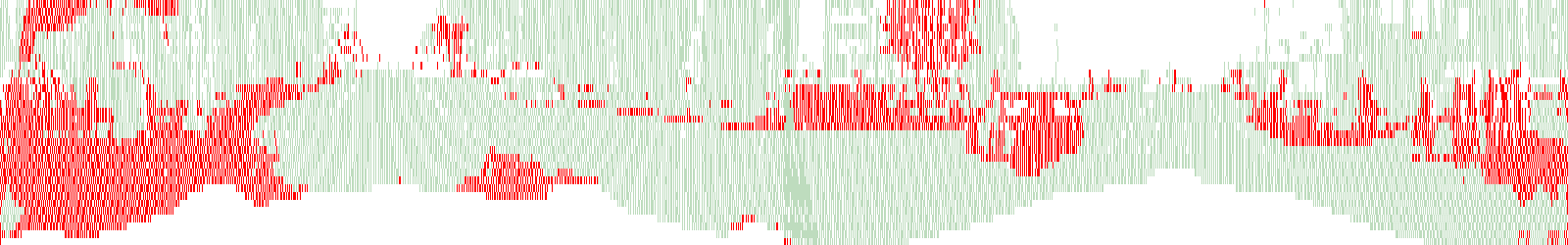}
    \caption{MT~\cite{tarvainen2017mean}}
\end{subfigure}
\begin{subfigure}[b]{.24\linewidth}
    \includegraphics[width=\textwidth]{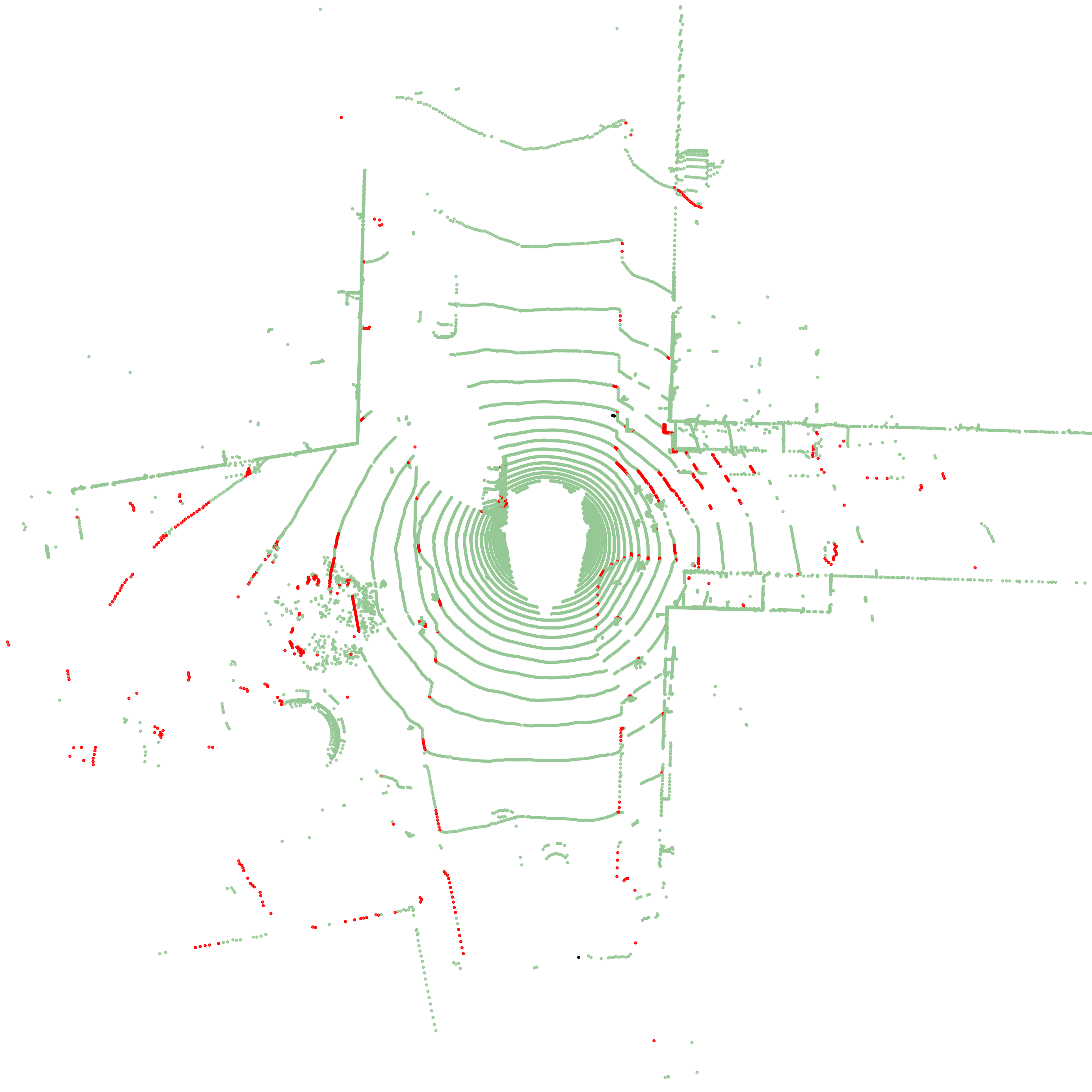}
    \includegraphics[width=\textwidth]{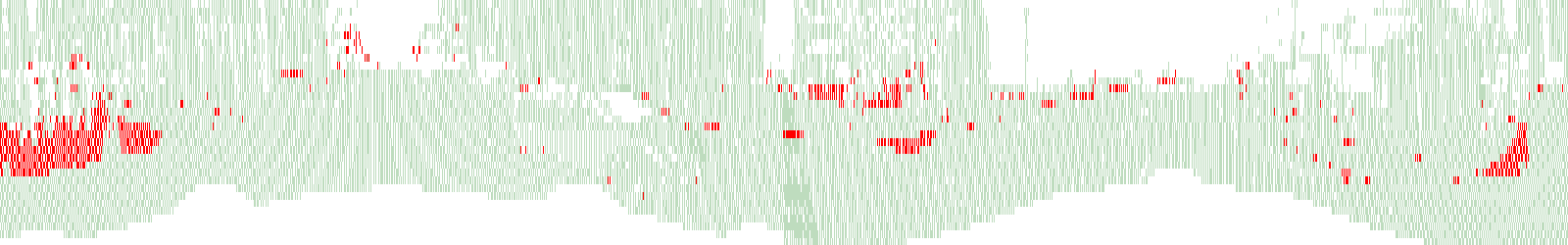}
    \includegraphics[width=\textwidth]{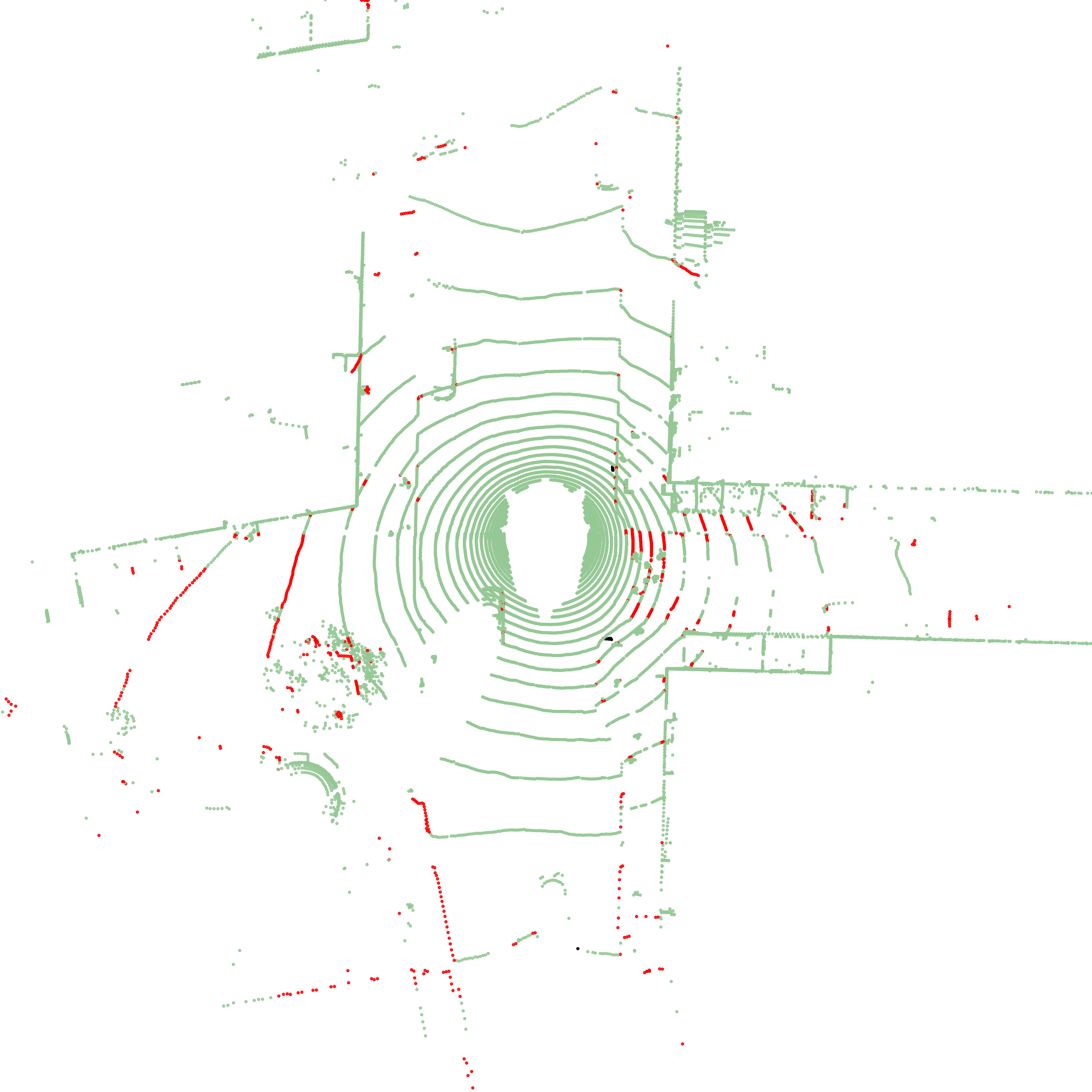}
    \includegraphics[width=\textwidth]{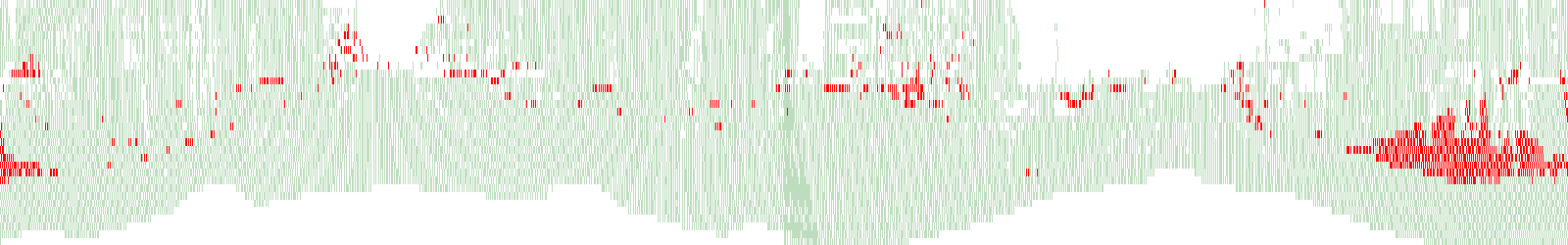}
    \caption{LaserMix~\cite{kong2023lasermix}}
\end{subfigure}
\begin{subfigure}[b]{.24\linewidth}
    \includegraphics[width=\textwidth]{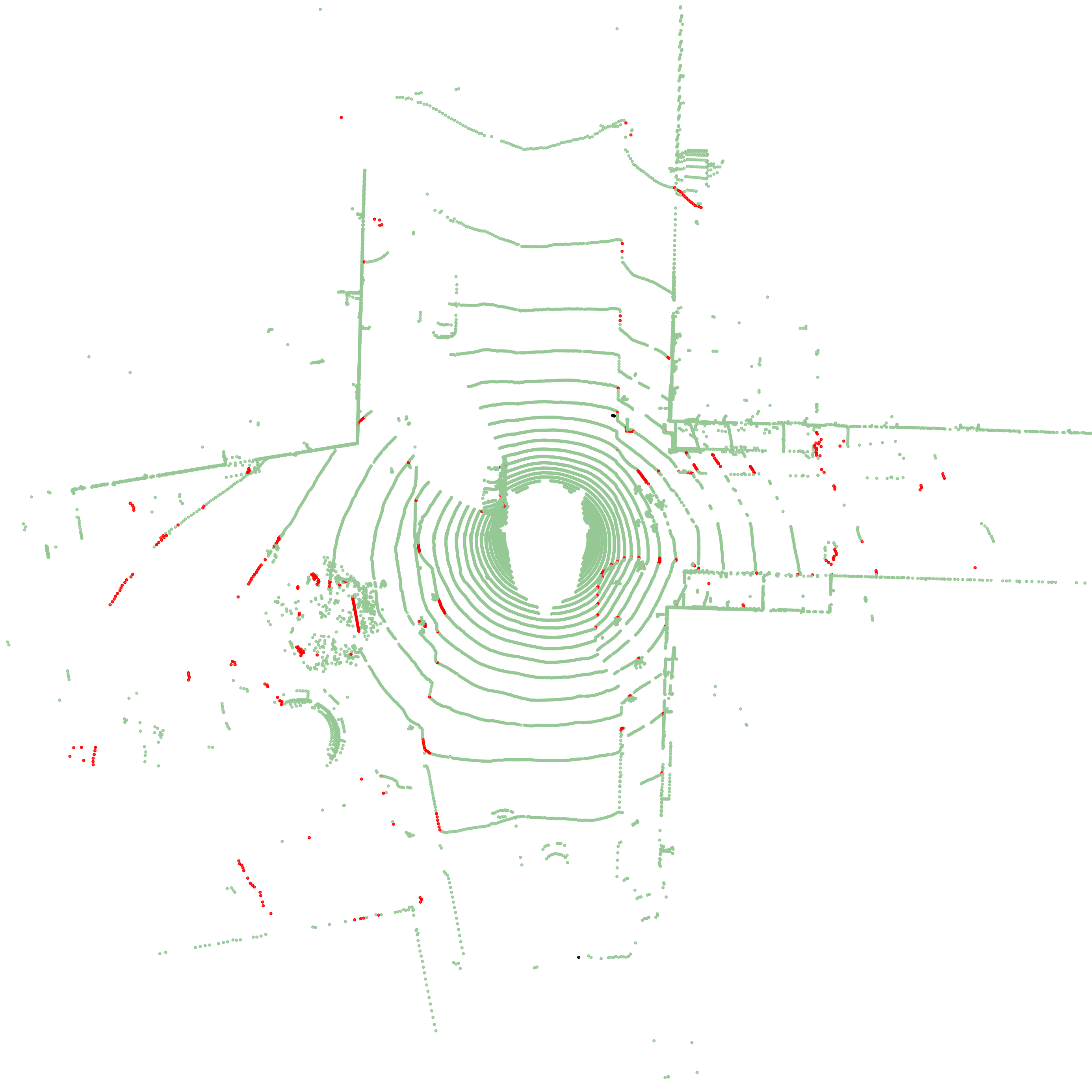}
    \includegraphics[width=\textwidth]{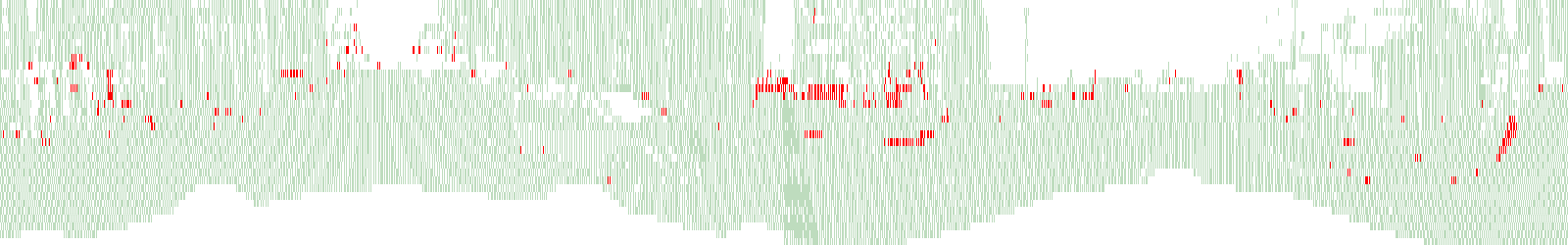}
    \includegraphics[width=\textwidth]{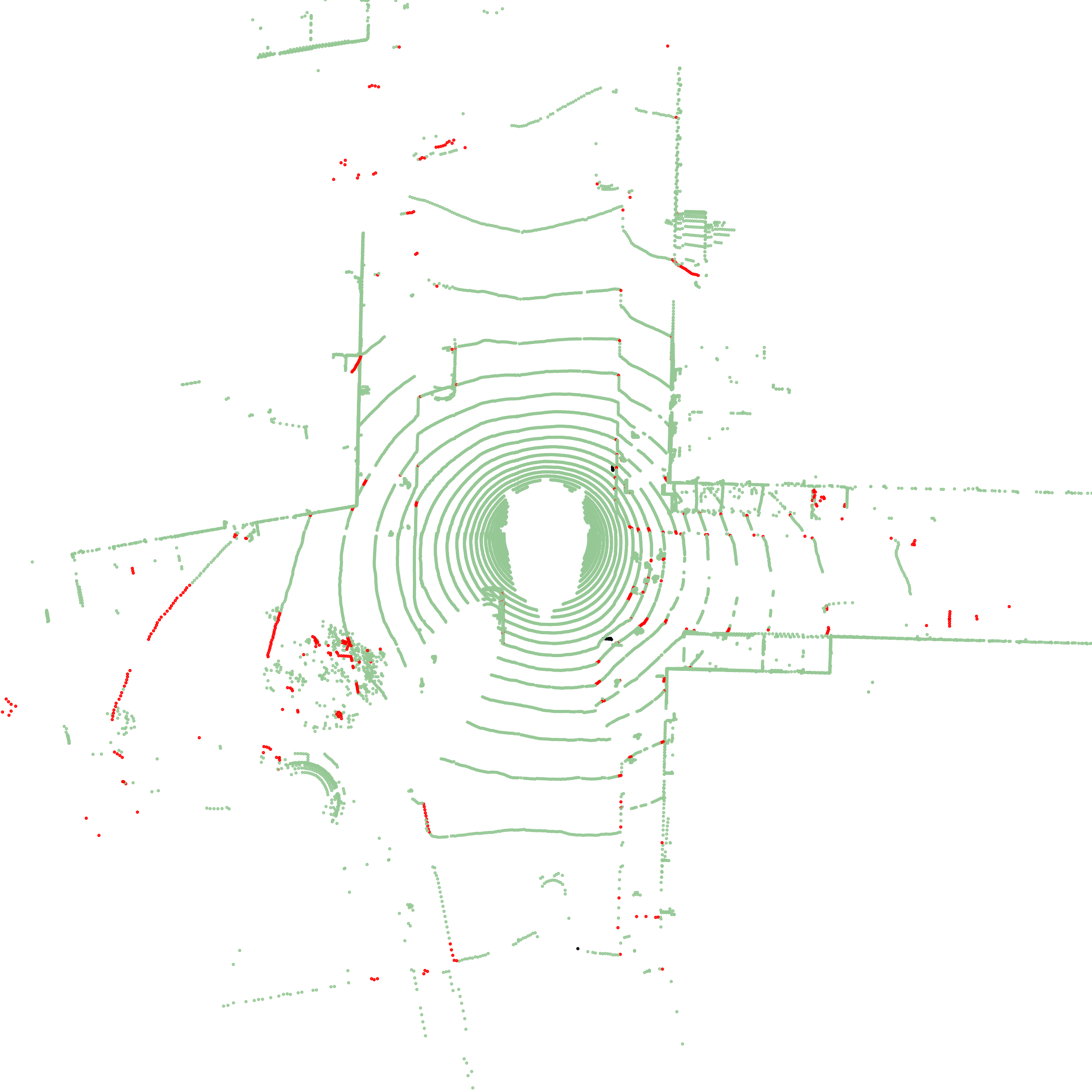}
    \includegraphics[width=\textwidth]{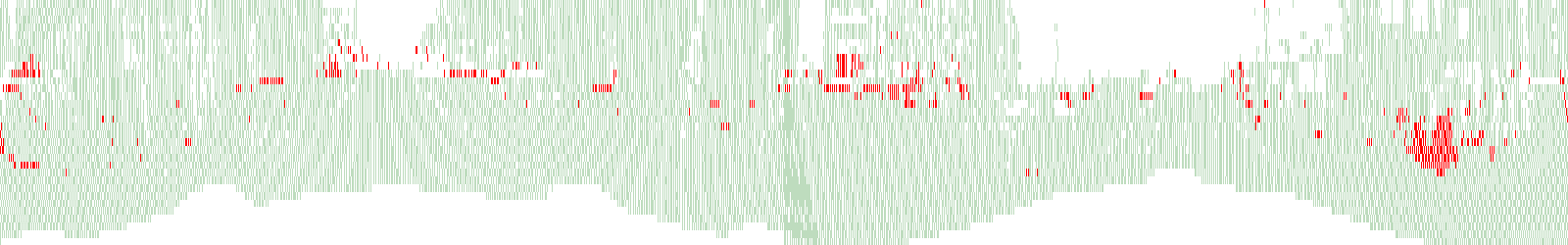}
    \caption{Ours}
\end{subfigure}
    \caption{\textbf{Error maps} visualised from LiDAR points (top) and \textit{range view} (bottom) in 2 examples of the  nuScenes~\cite{caesar2020nuscenes} dataset. The correct predictions are in \textcolor{green!60}{green} and the mistakes are highlighted in \textcolor{red}{red}.}
    \label{fig:error_map}
    \vspace{-10pt}
\end{figure*}
\subsection{Visualisation}
\begin{figure*}[t!]
\vspace{2pt}

    \centering
    \begin{subfigure}[t]{0.325\textwidth}
    \includegraphics[width=\textwidth]{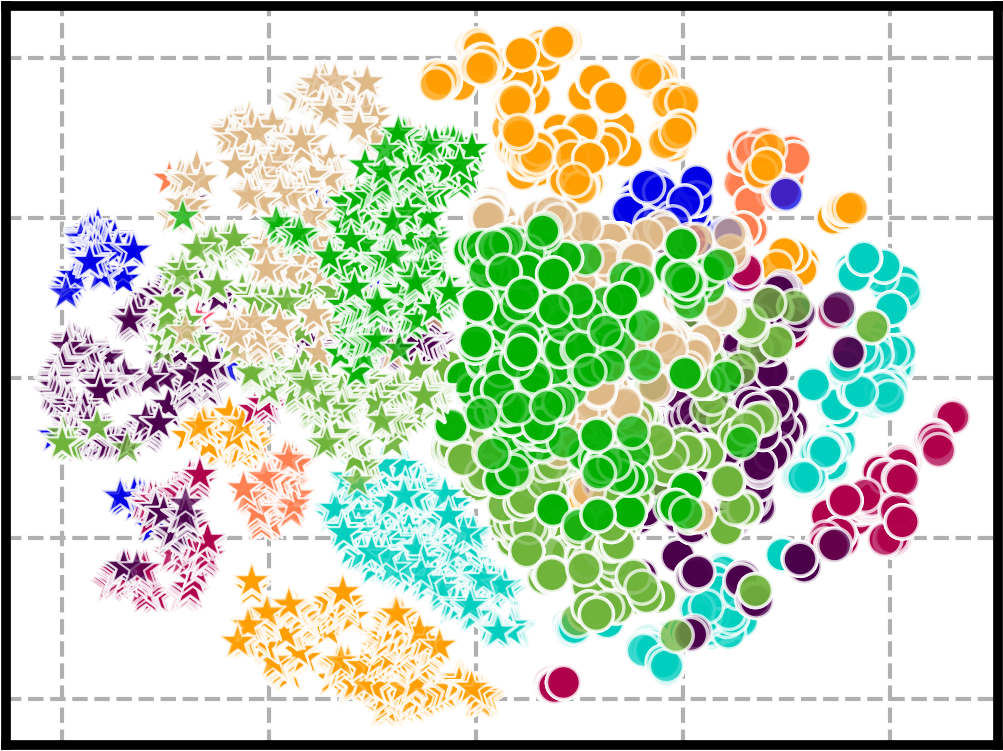}
    \caption{w/o Contras.}
    \label{fig:tsne_a}
    \end{subfigure}
    \begin{subfigure}[t]{0.325\textwidth}
    \includegraphics[width=\textwidth]{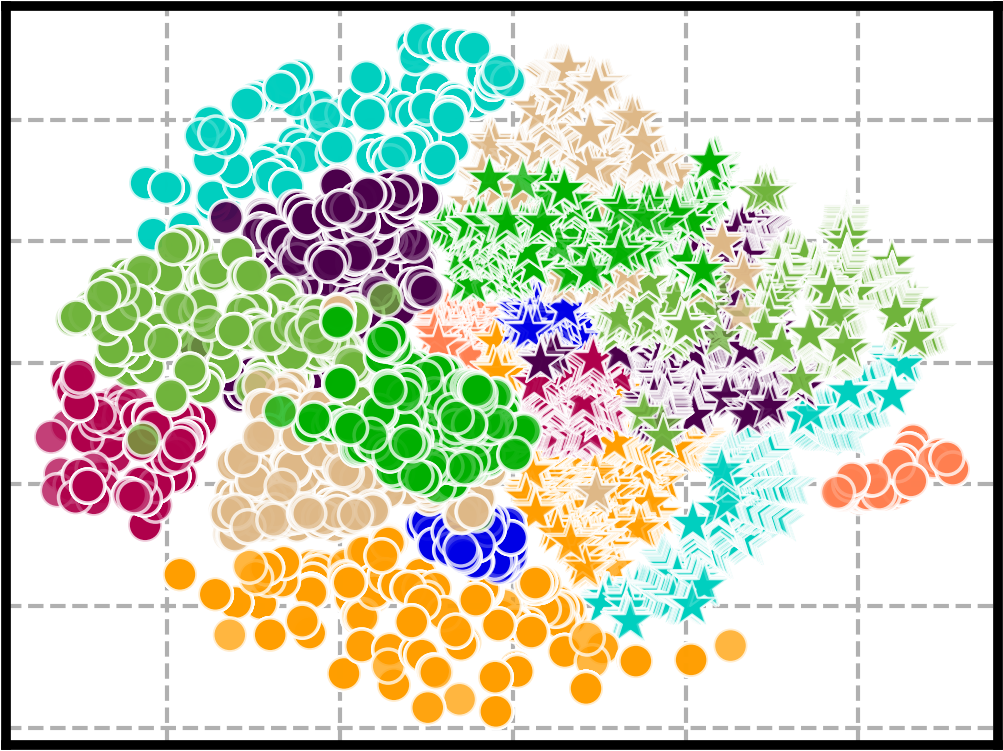}
    \caption{ContrasSeg~\cite{wang2021exploring}}
    \label{fig:tsne_b}
    \end{subfigure}
    \begin{subfigure}[t]{0.325\textwidth}
    \includegraphics[width=\textwidth]{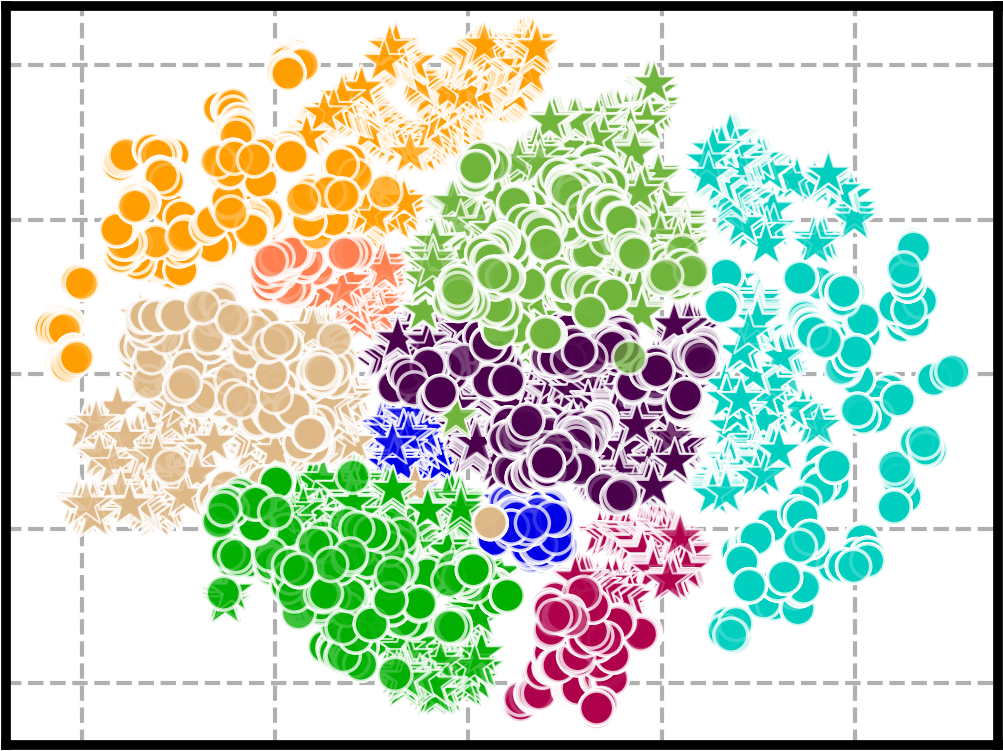}
    \caption{Ours}
    \label{fig:tsne_c}
    \end{subfigure}
    \caption{\textbf{T-SNE} visualisation for the pixel- (and voxel-) wise latent embedding features in the nuScenes~\cite{caesar2020nuscenes} dataset. The features \rangembeds~ and \voxelembeds~ indicate range and voxel representations and they are colored according to the category label. }
    \vspace{-5pt}
    \label{fig:tsne_visual}
\end{figure*}

\textbf{Visualisation of prediction error maps.} We compare the error maps from predictions in Fig.~\ref{fig:error_map}, including Mean Teacher~\cite{tarvainen2017mean} (2$^\text{nd}$ column), LaserMix~\cite{kong2023lasermix} (3$^\text{rd}$ column) and ours (last column). The visualisations show the bird's eye view's output in the first row, followed by the range view results in the second row. Notably, our method's prediction demonstrates a superior qualitative result, exhibiting fewer mistaken predictions (highlighted in \textcolor{red}{red}) and more correct predictions (marked in \textcolor{green}{green}) for both views. \\
\textbf{T-SNE visualisation of the latent features}. Figure~\ref{fig:tsne_visual} displays the latent embedding distribution from the nuScenes~\cite{caesar2020nuscenes} dataset.  Fig.~\ref{fig:tsne_a} shows the results without contrastive learning, Fig.~\ref{fig:tsne_b} displays the results based on the label-guided contrastive learning~\cite{wang2021exploring}, and our prototypical contrastive learning is in Fig.~\ref{fig:tsne_c}. The samples denoted by \rangembeds~ and \voxelembeds~ indicate the latent embeddings from range and voxel models, respectively, where different colours represent the classes. Fig.~\ref{fig:tsne_a} displays an unstructured latent representation due to the absence of contrastive learning, but Fig.~\ref{fig:tsne_b} successfully clusters class embeddings from to the same modalities (e.g., \{\rangembeds~, \rangembeds~\} and \{\voxelembeds~, \voxelembeds~\}), yet it fails to capture the semantic relationships between different modalities within the same category. Our method, depicted in Fig.~\ref{fig:tsne_c},  successfully clusters embeddings from the same class and different modalities \{\rangembeds~, \voxelembeds~\}, enabling a more effective distribution of embeddings for subsequent segmentation. 
\label{sec:exp_visual}

\vspace{-8pt}

\section{Conclusion and Future Work}
In this study, we introduced the novel semi-supervised LiDAR semantic segmentation method ItTakesTwo (IT2).
IT2 improves generalisation with consistent predictions from peer LiDAR representations, and it also introduces a more effective contrastive learning using informative embeddings sampled from a learned distribution of positives and negatives.
Finally, through experimental studies, we identify an efficient way to apply data augmentation to each representation. Our approach achieves best results across various benchmarks under all partition protocols, demonstrating its effectiveness.

The \textbf{limitations} of the current work are twofold: 1) it only explores two representations for consistency learning, potentially leaving out other valuable input modalities; and 2) it utilises single models for pseudo-label generation, overlooking the potential benefits of fusing results from different representations. Addressing these limitations might further enhance the generalisation in the limited labelled data situation, so we plan to work on this point in the future.

\section{Acknowledgements.}
The project is supported by the Australian Research Council (ARC) through grant FT190100525.

%
\bibliographystyle{splncs04}
\bibliography{ref}

\clearpage
\begin{center}
    
\textbf{\Large ItTakesTwo: Leveraging Peer Representations for Semi-supervised LiDAR Semantic Segmentation\\ \textcolor{red}{\textbf{(Supplementary Material)}}} 
\vspace{20pt}
\end{center}

\begin{table}[ht!]
\vspace{-30pt}
\caption{\textbf{Partial sampling results} 
in SemanticKITTI~\cite{behley2019semantickitti} benchmark based on batch-wise evaluation following~\cite{jiang2021guided}. TTA indicates the test time augmentation and the results "w/o" TTA are reported based on the checkpoints from the official GitHub~{\protect\tablefootnote{\scriptsize \url{https://github.com/llijiang/GuidedContrast/tree/main?tab=readme-ov-file\#semantickitti-2}}}. The best results are marked in \textcolor{darkred}{red}. }
\vspace{-5pt}
\label{tab:partial_batch}
\centering
\resizebox{.7\linewidth}{!}{
\begin{tabular}{c|c|c|c|c|c}
\boldline
\multirow{2}{*}{TTA} & \multirow{2}{*}{Method} & \multicolumn{4}{c}{SemanticKITTI (partial)} \\
\cline{3-6}
                     &                         & 5\%         & 10\%        & 20\%       & 40\%       \\
\boldline
\multirow{2}{*}{\xmark}    & GPC~\cite{jiang2021guided}                      & 42.10     & 48.30     & 57.90    & 59.32    \\
                     & IT2                     &\textcolor{darkred}{45.97} {\tiny \textcolor{gray}{(3.87$\uparrow$)}}   &\textcolor{darkred}{50.87} {\tiny \textcolor{gray}{(2.57$\uparrow$)}}     &\textcolor{darkred}{60.33} {\tiny \textcolor{gray}{(2.43$\uparrow$)}}   &\textcolor{darkred}{63.31} {\tiny \textcolor{gray}{(3.99$\uparrow$)}}   \\
\boldline
\multirow{2}{*}{\textcolor{red}{\cmark}}    & GPC~\cite{jiang2021guided}                     & 42.45     & 48.77     & 58.78    & 59.96    \\
                     & IT2                     & \textcolor{darkred}{46.44} {\tiny \textcolor{gray}{(3.99$\uparrow$)}}    &  \textcolor{darkred}{51.97} {\tiny\textcolor{gray}{(3.20$\uparrow$)}}    &  \textcolor{darkred}{61.43} {\tiny\textcolor{gray}{(2.69$\uparrow$)}}    &  \textcolor{darkred}{64.83} {\tiny \textcolor{gray}{(4.87$\uparrow$)}}   \\
\boldline
\end{tabular}
}
\vspace{-25pt}
\end{table}
\section{Different Evaluation Process}
The partial sampling approach GPC~\cite{jiang2021guided} employs a \textbf{distinct evaluation protocol}\protect\footnote{\scriptsize \url{https://github.com/llijiang/GuidedContrast/blob/add37a8ecf68a59698d6b6aa73735d94ae9c002d/util/utils.py\#L23}} comparing to the common LiDAR point semantic segmentation methods~\cite{lai2021semi, li2023less, zhu2021cylindrical}\footnote{\scriptsize\url{https://github.com/xinge008/Cylinder3D/blob/30a0abb2ca4c657a821a5e9a343934b0789b2365/utils/metric\_util.py\#L19}}. 
We emphasise that different evaluation protocols can lead to unfair competition.
In Tab.~\ref{tab:partial_batch}, we present our results based on the evaluation process following the approach in GPC~\cite{jiang2021guided}. \textit{We highlight that our reported results with all the labelled ratios are derived from consistent checkpoints in the main paper Tab. \textcolor{red}{2}}. The `TTA' indicates the test-time augmentation, where the paper results of GPC~\cite{jiang2021guided} are all based on the `TTA'. \\
Our method achieves the best performance across various labelled ratios. For instance, in the case of 5\% and 40\% labelled data with TTA post-processing, we outperform GPC~\cite{jiang2021guided} by 3.99\% and 4.87\% in mIoU, respectively. These consistent improvements underscore the robustness of peer representation in label-efficient LiDAR segmentation and also show the effectiveness of the IT2 approach.

\begin{figure}[t!]
    \centering
    \begin{subfigure}[t]{.244\textwidth}
    \includegraphics[width=\textwidth]{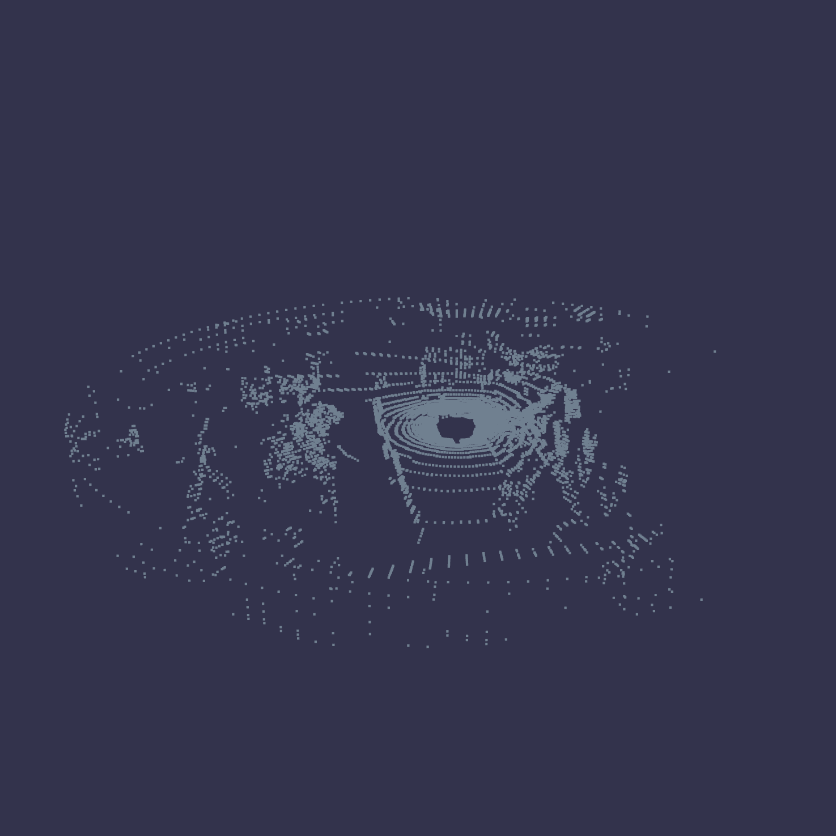}
    \includegraphics[width=\textwidth]{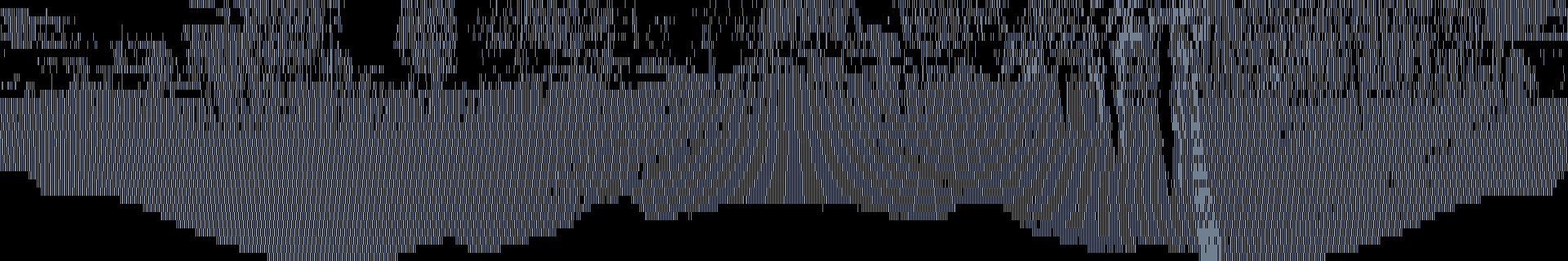}
    \caption*{\textcolor{gray}{case \textbf{(a)}}}
    \end{subfigure}
    \begin{subfigure}[t]{.244\textwidth}
    \includegraphics[width=\textwidth]{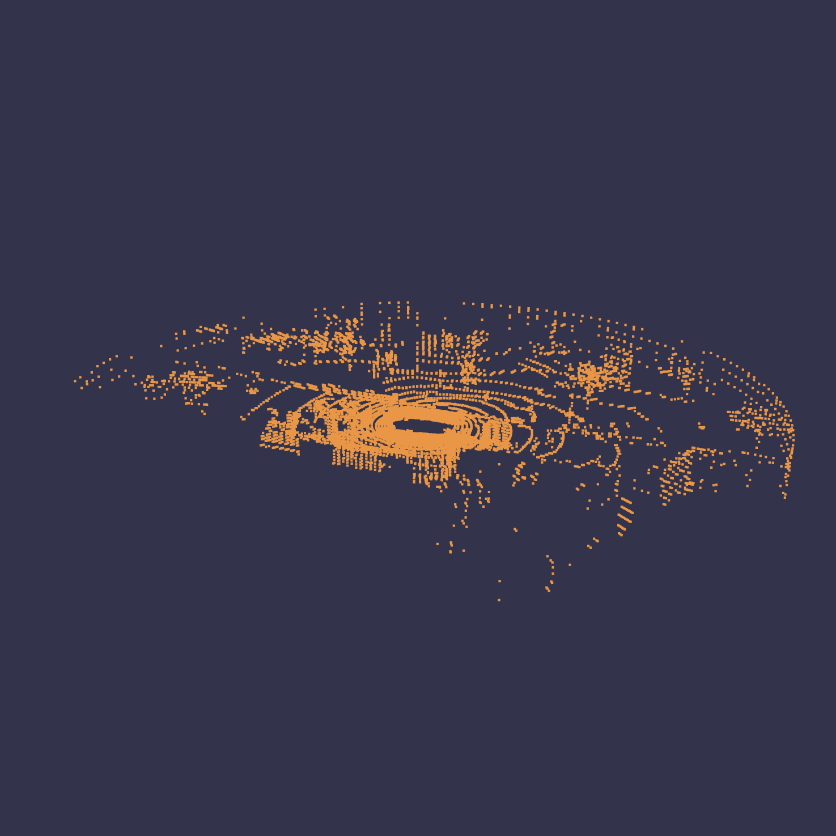}
    \includegraphics[width=\textwidth]{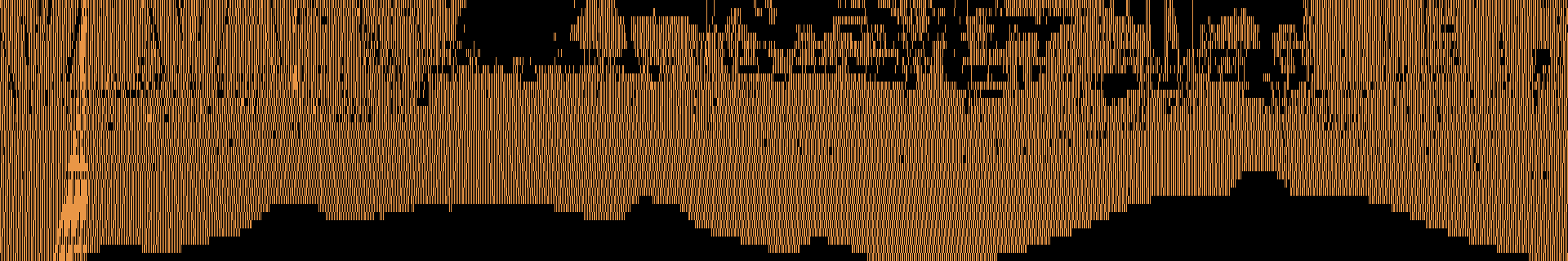}
    \caption*{\textcolor{orange}{case \textbf{(b)}}}
    \end{subfigure}
    \begin{subfigure}[t]{.244\textwidth}
    \includegraphics[width=\textwidth]{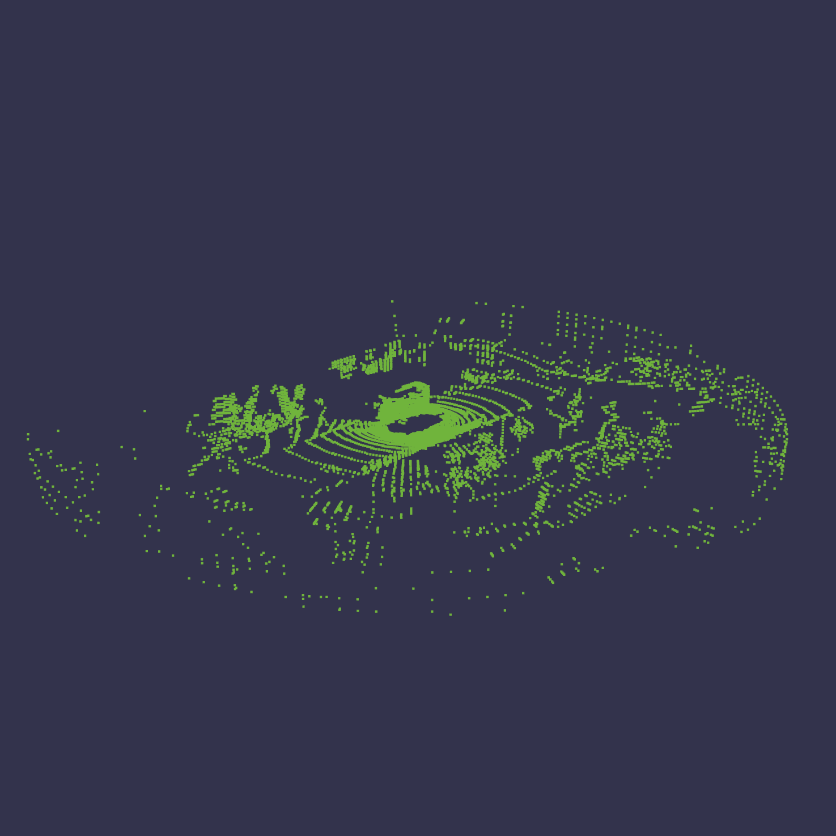}
    \includegraphics[width=\textwidth]{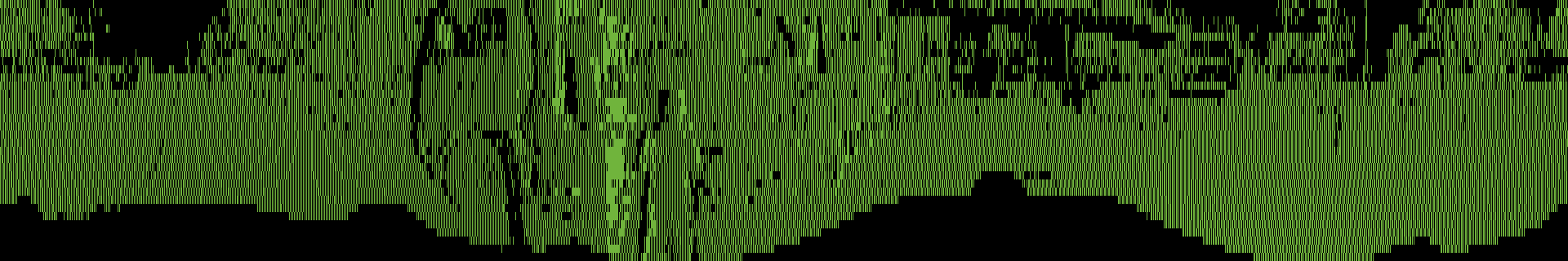}
    \caption*{\textcolor{green}{case \textbf{(c)}}}
    \end{subfigure}
    \begin{subfigure}[t]{.244\textwidth}
    \includegraphics[width=\textwidth]{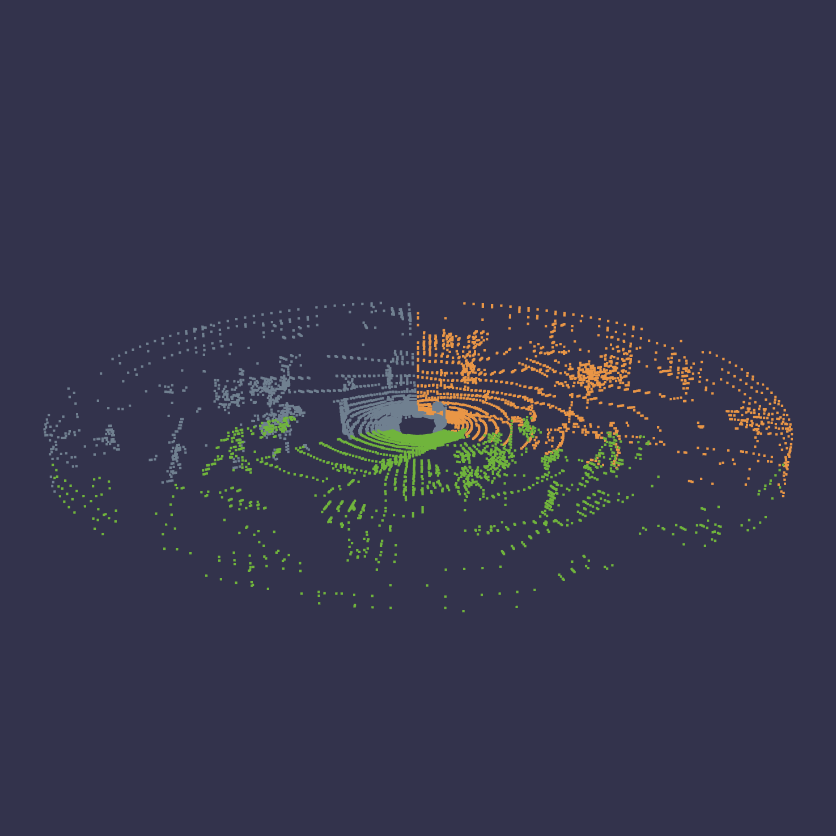}
    \includegraphics[width=\textwidth]{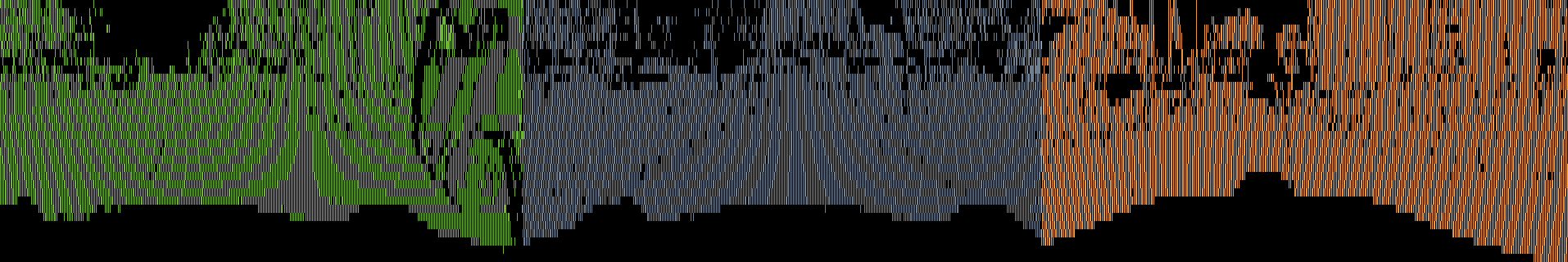}
    \caption*{\textbf{mix}}
    \end{subfigure}
    \caption{Illustration of our representation-specific data augmentation. In the first row, we apply single-inclination LaserMix~\cite{kong2023lasermix} for the voxel grids and in the second row, we apply multi-boxes CutMix~\cite{yun2019cutmix} for the range images. The mixed results for each representations are displayed in the last column, where the different colors demonstrate the mix protocol of case \textcolor{gray}{\textbf{(a)}},  \textcolor{orange}{\textbf{(b)}} and  \textcolor{green}{\textbf{(c)}}. 
    }
    \label{fig:aug}
    \vspace{-20pt}
\end{figure}
\section{Representation-specific Data Augmentation}

As shown in the Fig.~\ref{fig:aug}, we perform the augmentations for distinct representations to cases \textcolor{gray}{\textbf{(a)}},  \textcolor{orange}{\textbf{(b)}} and  \textcolor{green}{\textbf{(c)}} in a batch of inputs and the mixed results \textbf{(d)} are in the last column. We experimentally observed that utilising such different augmentations in different representations can achieve better performance.

\noindent \textbf{Voxel representation augmentation}. We implement LaserMix~\cite{kong2023lasermix} for augmentation of our voxel inputs as illustarted \ul{in the top row} of Fig.~\ref{fig:aug}. We set $\alpha=\frac{360}{\text{batch size}}$ and $\phi=1$ to achieve optimal performance, where $\alpha$ represents azimuth, and $\phi$ denotes the inclination direction. For further details, please refer to Table 4 in the LaserMix~\cite{kong2023lasermix} paper. \\
\textbf{Range representation augmentation}. As demonstrated \ul{in the bottom row} of Fig.~\ref{fig:aug}, we incorporate CutMix~\cite{yun2019cutmix} into our range images, utilising multi-boxes cropped within the same mini-batch. To prevent overlap between these boxes, we set the width of each box to $\frac{\text{image width}}{\text{batch size}}$ and they are mixed in box-by-box manner. \\

{\LinesNumberedHidden
\vspace{-30pt}
\begin{algorithm}[ht!] 
  \caption{Pseudo code of the data augmentation process}
  \label{alg:aug}
\hspace{-18pt}\vspace{3pt}
     \textcolor{gray}{\# $\mathbf{x}_\text{voxel}$, $\mathbf{x}_\text{range}$: voxel, range representations from same point scan.} \\
\hspace{-18pt}\vspace{3pt}
    \textcolor{gray}{\# $f_\text{voxel}$, $f_\text{range}$: voxel, range networks.} \\
\hspace{-18pt}\vspace{3pt}
    $\mathbf{y}_{\text{voxel}}$ = \texttt{range2voxel(model$_\text{range}(\mathbf{x}_\text{range})$)} \textcolor{gray}{\# transfer pseudo labels: range $\rightarrow$ voxel} \\
\hspace{-18pt}\vspace{3pt}
    $\mathbf{y}_{\text{range}}$ = \texttt{voxel2range(model$_\text{voxel}(\mathbf{x}_\text{voxel})$)} \textcolor{gray}{\# transfer pseudo labels: voxel $\rightarrow$ range} \\
\hspace{-18pt}\vspace{3pt}
    $\mathbf{x}_\text{voxel}$, $\mathbf{y}_{\text{voxel}}$ =  \texttt{voxel\_augment($\mathbf{x}_\text{voxel}$, $\mathbf{y}_{\text{voxel}}$)}  \textcolor{gray}{\# voxel augmentation} \cite{kong2023lasermix}  \\
\hspace{-18pt}\vspace{3pt}
    $\mathbf{x}_\text{range}$, $\mathbf{y}_{\text{range}}$ = \texttt{range\_augment($\mathbf{x}_\text{range}$, $\mathbf{y}_{\text{range}}$)} \textcolor{gray}{\# range augmentation} \cite{yun2019cutmix}  \\
\hspace{-18pt}\vspace{3pt}
    \textcolor{gray}{\# {\{$\mathbf{x}_\text{range}$, $\mathbf{y}_{\text{range}}$\} and \{$\mathbf{x}_\text{voxel}$, $\mathbf{y}_{\text{voxel}}$\} are the augmented samples and labels.}}
\hspace{-18pt} \vspace{1pt}
\end{algorithm}
\vspace{-20pt}}

Algorithm~\ref{alg:aug} shows the whole process of the representation-specific data augmentation in our approach in python coding style. Following~\cite{liu2022perturbed, yang2021st++}, these augmentations are carried out after generating the pseudo-labels (i.e., Eq. \textcolor{red}{(2)} from the main paper) for each representation. These augmentations are applied for both inputs and the labels, while their mixed results are utilised for the training of the unlabelled data.

\section{Additional Implementation Details}
In this section, we provide more implementation details of the model configuration and our proposed contrastive learning approach.

\subsection{Model Configuration}
\noindent\textbf{Cylinder3D~\cite{zhu2021cylindrical} setup.} Following Lasermix~\cite{kong2023lasermix}, we employ the Cylinder3D~\cite{zhu2021cylindrical} network for the voxel representation. For the \textit{Uniform Sampling} strategy in SemanticKITTI~\cite{behley2019semantickitti} and nuScenes~\cite{caesar2020nuscenes} benchmarks, we use an input resolution of [$240, 180, 20$] and reduce the feature map to $16$ for fair comparison. We adopt the original configuration from the paper~\cite{zhu2021cylindrical} with a resolution of [$480, 360, 32$] and a feature map size of $32$ for all other experiments.\\

\noindent \textbf{FidNet~\cite{zhao2021fidnet} setup.} We use FidNet~\cite{zhao2021fidnet} for the range images, following~\cite{kong2023lasermix}, where we employ ResNet-34~\cite{he2016deep} variants for all experiments. Similarly, in the \textit{Uniform Sampling} strategy of SemanticKITTI~\cite{behley2019semantickitti} and nuScenes~\cite{caesar2020nuscenes} benchmarks, we set the resolution to $32\times1920$ and $64\times 2048$ for a fair comparison with~\cite{kong2023lasermix}. For all other experiments, we maintain a fixed image width of $64\times 960$ for the enhanced efficiency. 
\subsection{Contrastive Learning Configuration}
Following~\cite{wang2021exploring, khosla2020supervised}, we employ a 3-layer projector to generate embedding results for intermediate features. The projector has the architecture of `\texttt{convolution layer, batch-norm layer, convolution layer}'. For the voxel representation, we use the 3D convolution layer from the SparseConv library\footnote{\url{https://github.com/traveller59/spconv}}, while 2D convolution is utilised for the range images following the common setup~\cite{liu2023residual, wang2021exploring, li2023less}. The depth of the embedding features is fixed at $64$ for all experiments. \\

\noindent \textbf{Sampling Strategies.} The current hardware cannot handle the contrastive learning in the dense tasks, and a sampling strategy is necessary for all contrastive learning based methods~\cite{liu2023residual, wang2021exploring}. We adopt the easy-hard mining for the dense embedding samples based on~\cite{wang2021exploring}. The easy embedding samples are randomly selected based on the dense results where the predictions equal with (pseudo or real) labels, while the hard embedding samples are chosen based on the results that exhibit disagreement between the prediction and the (pseudo or real) label. We follow~\cite{wang2021exploring} to split the ratio of these easy and hard samples with the ratio being $1:1$, and the total number is set to $200$ for both voxel and range representations in all experiments. 

\subsection{Training Configuration}
In the nuScenes~\cite{caesar2020nuscenes} dataset, we use a batch size of $4$ for both labelled and unlabelled data. The learning rate is set to $6e^{-3}$, and the maximum number of epochs is set to $90$. In the SemanticKITTI~\cite{behley2019semantickitti} dataset, we opt for a smaller batch size of $2$ for both labelled and unlabelled data, with a learning rate of $8e^{-3}$ and a maximum of $105$ epochs. For both datasets, we employ $4$ GPUs for distributed data parallel training, incorporating a \textit{polynomial} learning rate decay of $(1-\frac{\text{iter}}{\text{max\_iter}})^{0.9}$. We use the AdamW~\cite{loshchilov2017decoupled} optimizer for all experiments with a decay of 0.001, and beta values in the range [0.9, 0.999].

\subsection{Gaussian Mixture Model Configuration}

We employ category-wise Gaussian Mixture Model (GMM) to learn from the incoming embedding features for both range and voxel representations in Eq. \textcolor{red}{(5)} of the main paper, where each category in the dataset have $5$ Gaussian Curves. 
The GMMs' parameters are updated in each iteration based on the exponential moving average (EMA) to keep track of their historical distribution via $\Gamma^{(t+1)} = \alpha \times \Gamma^{(t)} + (1-\alpha) \times \Gamma$, where $\alpha=0.996$ for all  experiments, where $\Gamma$ is from Eq. \textcolor{red}{(6)} main paper. In the sampling strategy of the virtual prototype, we firstly identify the related GMMs based on the categories of each of the incoming feature samples. Then we choose the Gaussian  within the GMMs based on the probability distribution of $\mathbf{q}_{m}^y$ from Eq. \textcolor{red}{(7)} in the main paper for each of the samples. After that, we assign the random variable $\xi \in [0, 1]$ and generate the prototypes $\mathbf{z}^{p}$ for each feature samples, where $\mathbf{z}^{p} = \mathbf{\mu}_{m}^y + \xi \times \mathbf{\Sigma}_{m}^y$ with $\mathbf{\mu}_{m}^y$, $\mathbf{\Sigma}_{m}^y$ from Eq. \textcolor{red}{(7)}.

\section{Fusion Results}
\begin{table}[t!]
\centering
\caption{The fusion results in the nuScenes\cite{caesar2020nuscenes} dataset based on the prediction from \textcolor{rangered}{range} and \textcolor{voxelblue}{voxel} representations. The best results are in \textcolor{darkred}{red} and the second best results are in \textit{italic}.}
\label{tab:fusion}
\begin{tabular}{c|c|c|c|c}
\boldline
\multirow{2}{*}{Repr.} & \multicolumn{4}{c}{nuScenes~\cite{caesar2020nuscenes}}  \\
\cline{2-5}
                       & 1\%   & 10\%  & 20\%  & 50\%  \\
\boldline
\rowcolor{lightred}range                  & 56.76 & 71.33 & 73.27 & 74.04 \\
\rowcolor{lightblue}voxel                  & \textit{56.96} & \textit{72.12} & \textit{73.55} & \textit{74.14} \\
\hline
ensemble               & 58.14 \textcolor{gray}{(1.18$\uparrow$)} & 73.77 \textcolor{gray}{(1.65$\uparrow$)} & 75.29 \textcolor{gray}{(1.74$\uparrow$)} & 76.54 \textcolor{gray}{(2.40$\uparrow$)} \\
\boldline
\end{tabular}
\vspace{-10pt}
\end{table}

In Tab.~\ref{tab:fusion}, we present the fusion results obtained from the models' outputs of both voxel and range representations in the nuScenes\cite{caesar2020nuscenes} dataset. Notably, we observe consistent improvements in the fusion results compared to the individual representation results. For example, it demonstrates 1.18\% and 2.40\% improvements in 1\% and 50\% labelled partition protocol, respectively. Given that the fusion prediction yields better generalisation, how to ensemble the pseudo label of the unlabelled data during the training can be an interesting topic in the future research.

\section{Detailed Results}
\begin{table}[t!]
\caption{The class-wise IoU results in nuScenes~\cite{caesar2020nuscenes} dataset among different partition protocol. The mIoU results are highlighted in \textcolor{darkred}{red}.}
\label{tab:class_nusc}
\resizebox{\linewidth}{!}{
\renewcommand{\arraystretch}{1.3}
\begin{tabular}{?c|c|c?cccccccccccccccc?}
\boldline
Repr. & ratio & \textbf{mean} & barr & bicy & bus  & car  & const & moto & ped  & cone & trail & truck & driv & othe & walk & terr & manm & veg  \\
\boldline
 \cellcolor{lightred} & 1\%     & \textcolor{darkred}{\textbf{56.5}} & 63.6 & 1.9  & 63.0 & 84.8 & 9.0   & 57.7 & 62.8 & 51.1 & 17.2  & 44.1  & 94.5 & 53.2 & 64.1 & 69.9 & 83.4 & 83.5 \\
 \cellcolor{lightred} & 10\%    & \textcolor{darkred}{\textbf{71.3}} & 75.0 & 26.4 & 81.1 & 90.1 & 39.9  & 77.2 & 76.0 & 61.5 & 56.5  & 73.3  & 96.2 & 66.7 & 72.5 & 74.4 & 87.8 & 86.6 \\
 \cellcolor{lightred} & 20\%    & \textcolor{darkred}{\textbf{73.4}} & 75.9 & 36.9 & 84.4 & 90.7 & 43.9  & 79.9 & 77.8 & 64.8 & 58.1  & 75.6  & 96.3 & 68.6 & 73.2 & 74.3 & 88.0 & 87.1 \\
 \cellcolor{lightred}{\multirow{-3}{*}[2.4ex]{\rotatebox[origin=c]{90}{\centering \textit{Range}}}} & 50\%    & \textcolor{darkred}{\textbf{74.0}} & 76.1 & 38.2 & 85.0 & 89.4 & 45.5  & 76.6 & 77.9 & 64.9 & 66.7  & 76.3  & 96.3 & 69.5 & 73.4 & 74.2 & 88.2 & 87.0 \\
\boldline
 \cellcolor{lightblue} & 1\%     & \textcolor{darkred}{\textbf{57.5}} & 58.2 & 3.8  & 67.6 & 83.3 & 16.9  & 63.0 & 62.9 & 47.2 & 18.5  & 47.1  & 94.0 & 53.5 & 64.8 & 70.3 & 83.8 & 84.8 \\
 \cellcolor{lightblue} & 10\%    & \textcolor{darkred}{\textbf{72.1}} & 72.4 & 26.7 & 89.7 & 91.0 & 46.8  & 75.7 & 73.2 & 58.2 & 57.1  & 80.0  & 95.8 & 68.4 & 72.4 & 73.2 & 87.5 & 85.9 \\
 \cellcolor{lightblue} & 20\%    & \textcolor{darkred}{\textbf{73.6}} & 74.0 & 34.0 & 90.7 & 91.3 & 47.9  & 76.8 & 76.4 & 60.9 & 57.8  & 79.8  & 95.8 & 67.8 & 73.6 & 73.3 & 88.7 & 86.5 \\
 \cellcolor{lightblue}{\multirow{-3}{*}[2.4ex]{\rotatebox[origin=c]{90}{\centering \textit{Voxel}}}} & 50\%    & \textcolor{darkred}{\textbf{74.1}} & 73.3 & 33.5 & 91.7 & 91.1  & 46.9  & 77.8 & 75.2 & 59.8 & 65.3  & 80.9  & 95.8 & 72.4 & 73.3 & 74.1 & 88.1 & 86.4 \\
\boldline
\end{tabular}}
\end{table}

\begin{table}[t!]
\caption{The class-wise IoU results in SemanticKITTI~\cite{behley2019semantickitti} dataset among different partition protocol. The mIoU results are highlighted in \textcolor{darkred}{red}.}
\label{tab:class_kitti}
\resizebox{\linewidth}{!}{
\renewcommand{\arraystretch}{1.3}
\begin{tabular}{?c|c|c?ccccccccccccccccccc?}
\boldline
Repr. & ratio & \textbf{mean} & car  & bicy & moto & truck & bus  & ped  & b.cyc & m.cyc & road & park & walk & o.gro & build & fence & veg  & trunk & terr & pole & sign \\
\boldline
\cellcolor{lightred} & 5\%     & \textcolor{darkred}{\textbf{57.0}} & 93.1 & 44.8 & 51.5 & 67.3  & 42.5 & 44.8 & 54.0  & 0.0   & 93.9 & 42.4 & 80.4 & 0.0   & 87.7  & 54.0  & 87.3 & 62.5  & 77.4 & 58.6 & 40.1 \\
\cellcolor{lightred} & 10\%    & \textcolor{darkred}{\textbf{62.4}} & 95.2 & 46.3 & 56.5 & 69.7  & 47.9 & 73.2 & 80.7  & 0.0   & 95.3 & 46.9 & 83.1 & 1.9   & 87.9  & 57.4  & 88.6 & 67.5  & 77.6 & 65.1 & 44.8 \\
\cellcolor{lightred} & 20\%    & \textcolor{darkred}{\textbf{62.7}} & 95.4 & 46.3 & 59.1 & 90.6  & 51.1 & 71.1 & 81.8  & 0.0    & 95.5 & 39.1 & 82.7 & 1.6   & 86.7  & 51.1  & 87.4 & 66.5  & 76.1 & 66.5 & 43.1 \\
\cellcolor{lightred}{\multirow{-3}{*}[2.4ex]{\rotatebox[origin=c]{90}{\centering \textit{Range}}}} & 40\%    & \textcolor{darkred}{\textbf{63.9}} & 94.8 & 54.8 & 62.1 & 70.9  & 45.0 & 73.6 & 82.3  & 0.0   & 95.6 & 52.1 & 83.9 & 10.7  & 88.9  & 59.4  & 87.1 & 67.2  & 74.8 & 65.8 & 45.3 \\
\boldline
\cellcolor{lightblue} & 5\%     & \textcolor{darkred}{\textbf{60.3}} & 92.9 & 49.5 & 43.9 & 85.7  & 40.6 & 65.0 & 83.0  & 0.0   & 91.5 & 35.8 & 77.5 & 0.3   & 89.4  & 54.2  & 86.5 & 66.5  & 73.0 & 64.6 & 45.7 \\
\cellcolor{lightblue} & 10\%    & \textcolor{darkred}{\textbf{63.3}} & 94.9 & 50.9 & 70.9 & 78.8  & 48.2 & 74.5 & 84.2  & 0.0   & 94.1 & 42.9 & 79.8 & 2.9   & 87.4  & 51.3  & 87.7 & 66.3  & 74.3 & 63.0 & 50.8 \\
\cellcolor{lightblue} & 20\%    & \textcolor{darkred}{\textbf{64.0}} & 96.7 & 54.1 & 73.6 & 74.4  & 59.5 & 75.4 & 86.8  & 0.3   & 93.5 & 41.8 & 79.5 & 1.2   & 88.3  & 50.6  & 86.1 & 67.6  & 69.5 & 64.8 & 52.1 \\
\cellcolor{lightblue}{\multirow{-3}{*}[2.4ex]{\rotatebox[origin=c]{90}{\centering \textit{Voxel}}}} & 40\%    & \textcolor{darkred}{\textbf{64.8}} & 95.7 & 50.3 & 75.4 & 79.4  & 52.5 & 75.4 & 90.8  & 1.6   & 94.7 & 46.9 & 81.6 & 1.0    & 87.3  & 52.4  & 87.5 & 69.2  & 75.1 & 64.6 & 49.6 \\
\boldline
\end{tabular}}

\end{table}

\begin{table}[t!]
\caption{The class-wise IoU results in ScribbleKITTI~\cite{unal2022scribble} dataset among different partition protocol. The mIoU results are highlighted in \textcolor{darkred}{red}.}
\label{tab:class_scribble}
\resizebox{\linewidth}{!}{
\renewcommand{\arraystretch}{1.3}
\begin{tabular}{?c|c|c?ccccccccccccccccccc?}
\boldline
Repr. & ratio & \textbf{mean} & car  & bicy & moto & truck & bus  & ped  & b.cyc & m.cyc & road & park & walk & o.gro & build & fence & veg  & trunk & terr & pole & sign \\
\boldline
\cellcolor{lightred} & 1\%     & \textcolor{darkred}{\textbf{46.6}} & 79.4 & 31.8 & 28.3  & 35.9  & 11.4 & 39.2 & 60.0   & 0.0    & 73.2 & 16.7 & 65.8 & 0.2   & 86.5  & 46.7  & 82.1 & 62.2  & 66.  & 60.8 & 40.2 \\
\cellcolor{lightred} & 10\%    & \textcolor{darkred}{\textbf{57.1}} & 92.4 & 47.5 & 45.4 & 67.4  & 30.3 & 55.2 & 62.8  & 0.0    & 92.9 & 40.9 & 79.4 & 1.6   & 87.7  & 52.6  & 84.7 & 66.3  & 72.5 & 60.2 & 45.9 \\
\cellcolor{lightred} & 20\%    & \textcolor{darkred}{\textbf{57.3}} & 87.9 & 33.0 & 43.9 & 66.8  & 43.3 & 59.6 & 63.0  & 0.0    & 91.7 & 40.6 & 79.6 & 6.1   & 88.2  & 53.3  & 82.6 & 67.5  & 74.3 & 61.3 & 46.7 \\
\cellcolor{lightred}{\multirow{-3}{*}[2.4ex]{\rotatebox[origin=c]{90}{\centering \textit{Range}}}} & 50\%    & \textcolor{darkred}{\textbf{58.6}} & 86.7& 35.1& 51.7& 77.4& 49.3& 66.1& 74.5& 0.1& 87.5& 31.1& 76.3& 8.7& 88.3& 45.3& 85.0& 66.8& 71.1& 64.0& 47.7
\\
\boldline
\cellcolor{lightblue} & 1\%     & \textcolor{darkred}{\textbf{47.9}} & 85.4 & 29.2 & 37.7 & 20.4  & 24.2 & 45.9 & 53.1  & 0.0    & 77.0 & 20.3 & 67.7 & 0.5   & 83.2  & 49.4  & 77.6 & 64.5  & 64.9 & 61.7 & 48.1 \\
\cellcolor{lightblue} & 10\%    & \textcolor{darkred}{\textbf{56.7}} & 93.1 & 43.1 & 47.2 & 63.8  & 33.2 & 60.6 & 70.0  & 0.0   & 89.5 & 34.1 & 74.6 & 1.4   & 87.4  & 51.9  & 84.7 & 61.1  & 72.5 & 60.9 & 48.8 \\
\cellcolor{lightblue} & 20\%    & \textcolor{darkred}{\textbf{57.5}} & 92.0 & 49.5 & 45.1 & 68.4  & 30.6 & 56.1 & 60.8  & 0.0    & 93.2 & 41.2 & 80.1 & 2.9   & 88.5  & 52.1  & 85.4 & 66.9  & 73.4 & 60.8 & 46.2 \\
\cellcolor{lightblue}{\multirow{-3}{*}[2.4ex]{\rotatebox[origin=c]{90}{\centering \textit{Voxel}}}} & 50\%    & \textcolor{darkred}{\textbf{58.3}} & 89.2& 48.7& 46.7& 73.2& 42.0& 62.2& 74.0& 0.0& 83.3& 40.3& 79.1& 3.2& 85.3& 54.2& 80.6& 67.1& 64.4& 64.2& 50.1\\

\boldline
\end{tabular}}
\end{table}
We have provided the class-wise Intersection-over-Union (IoU) validation results in the Tab.~\ref{tab:class_nusc}, Tab.~\ref{tab:class_kitti} and Tab.~\ref{tab:class_scribble} for the nuScenes~\cite{caesar2020nuscenes}, SemanticKITTI~\cite{behley2019semantickitti} and ScribbleKITTI~\cite{unal2022scribble} datasets, respectively. The mIou results follow the table results in the main paper and they are highlighted in \textcolor{darkred}{red}.

\section{Error Maps Visualisation}
\begin{figure*}[t!]
    \centering
\begin{subfigure}[b]{.32\linewidth}
    \includegraphics[width=\textwidth]{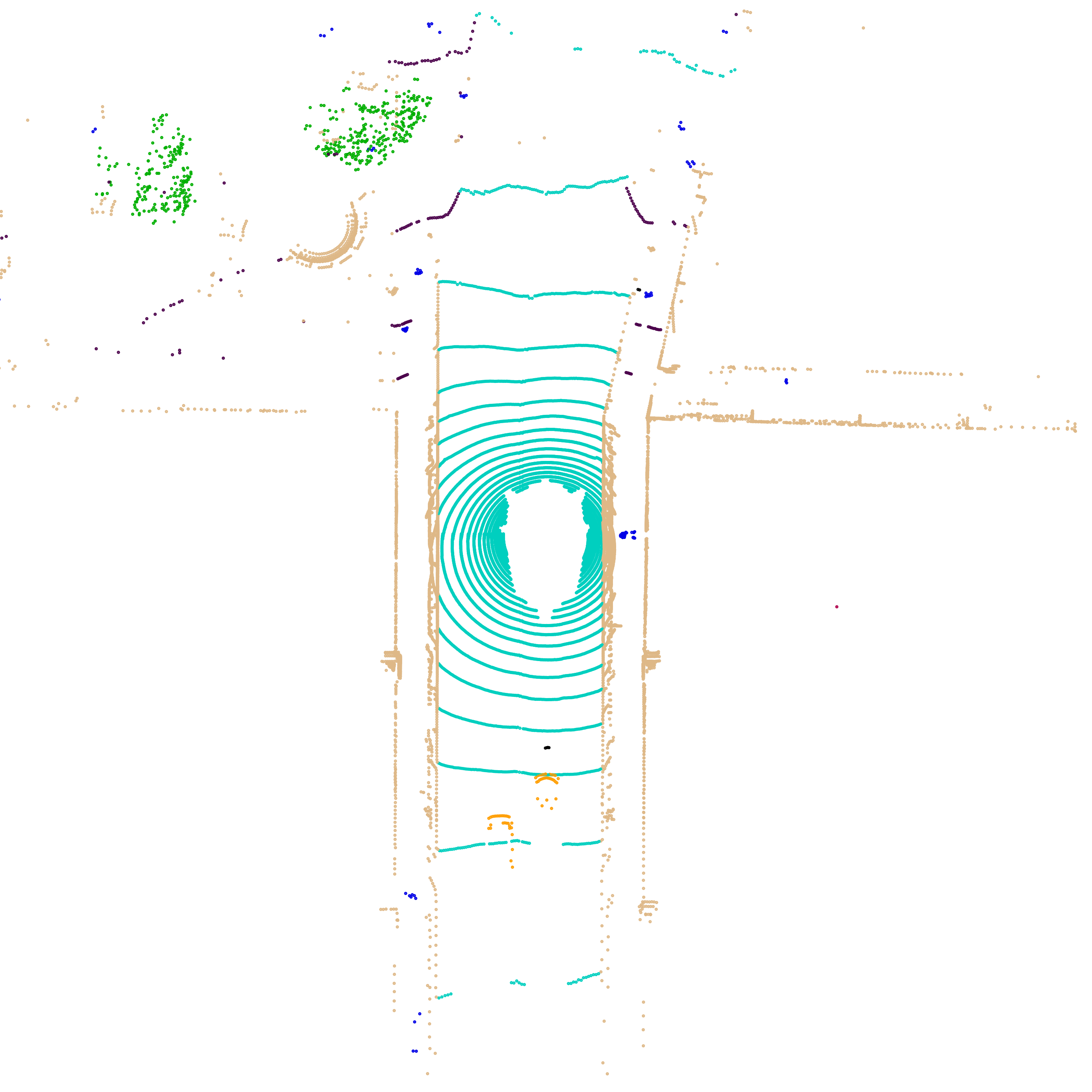}
    \includegraphics[width=\textwidth]{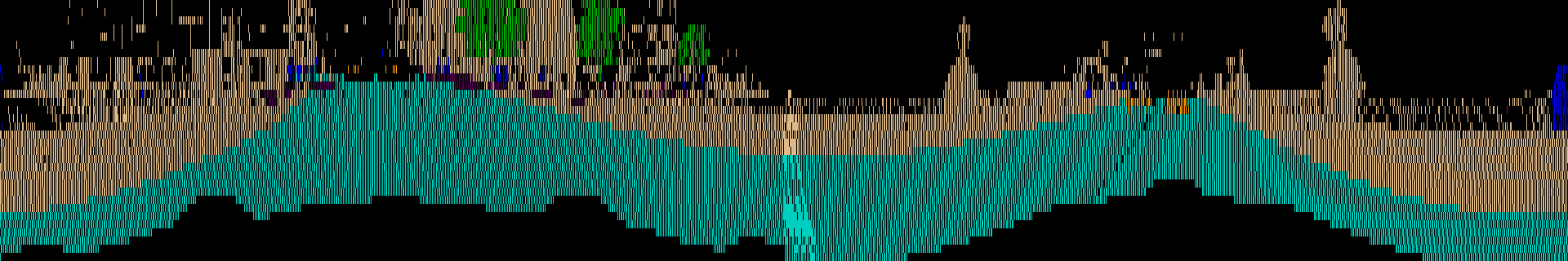}
    
    \includegraphics[width=\textwidth]{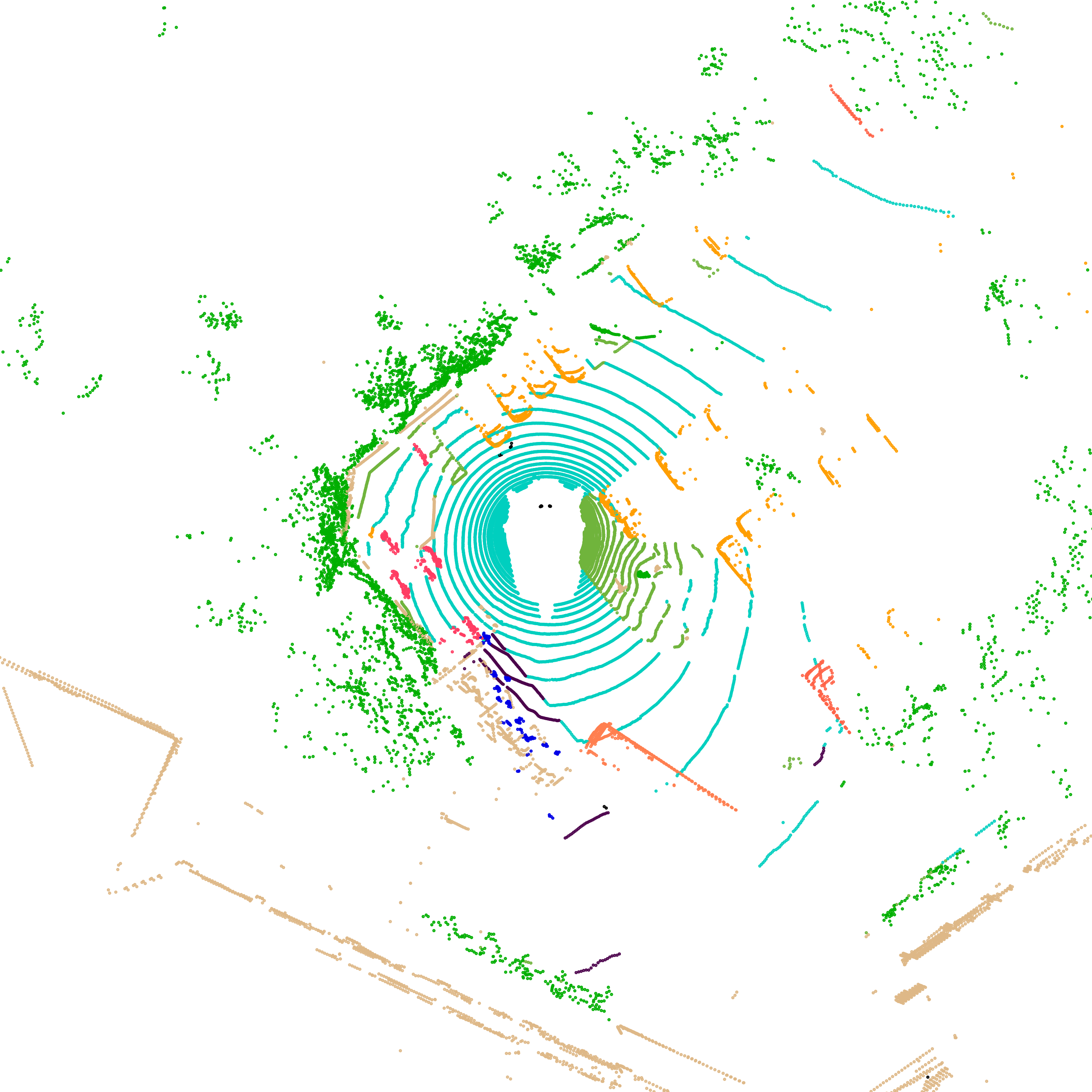}
    \includegraphics[width=\textwidth]{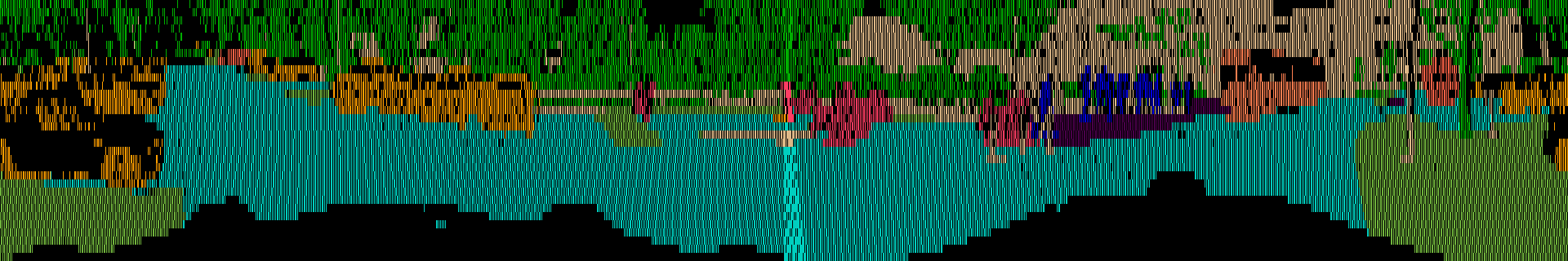}
    
    \includegraphics[width=\textwidth]{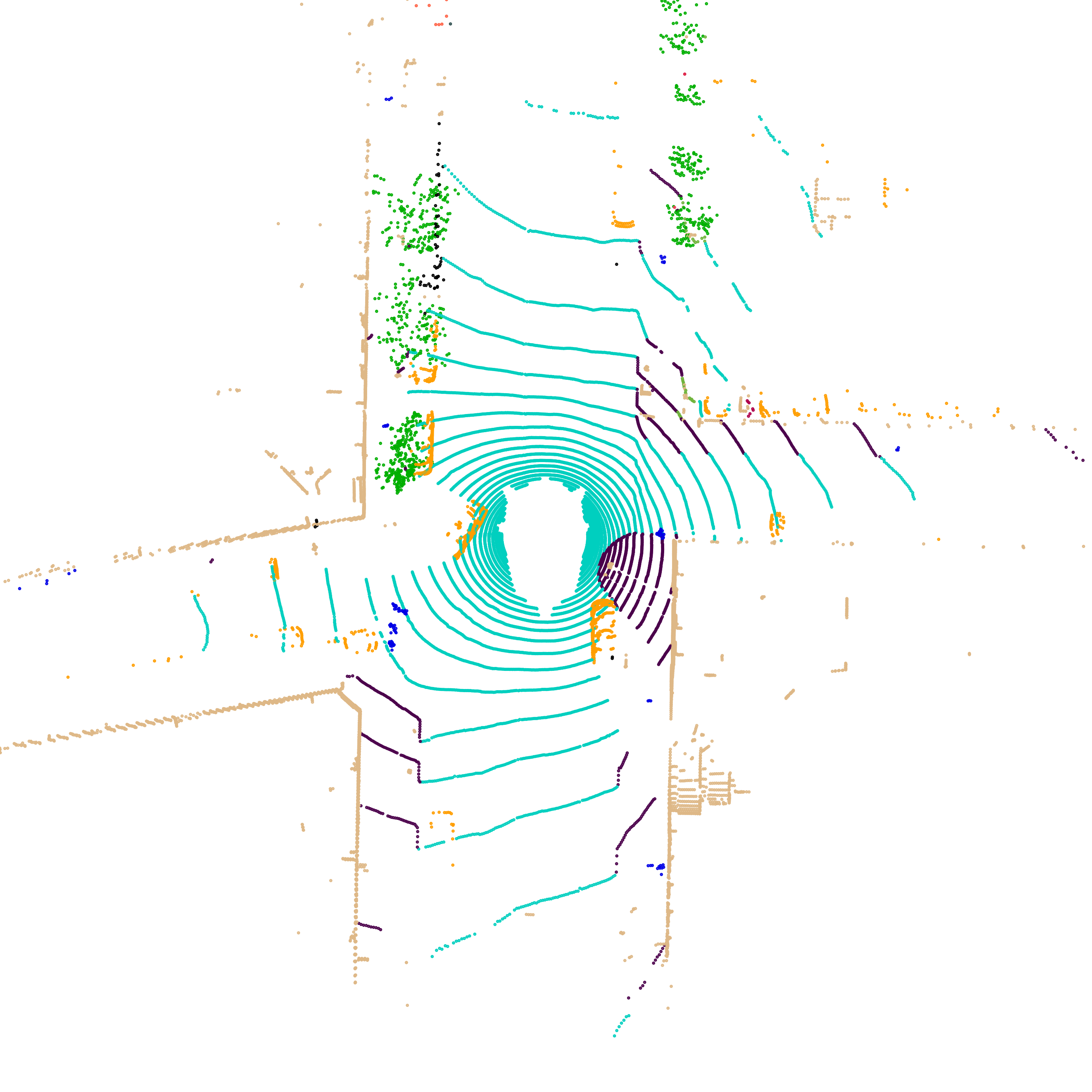}
    \includegraphics[width=\textwidth]{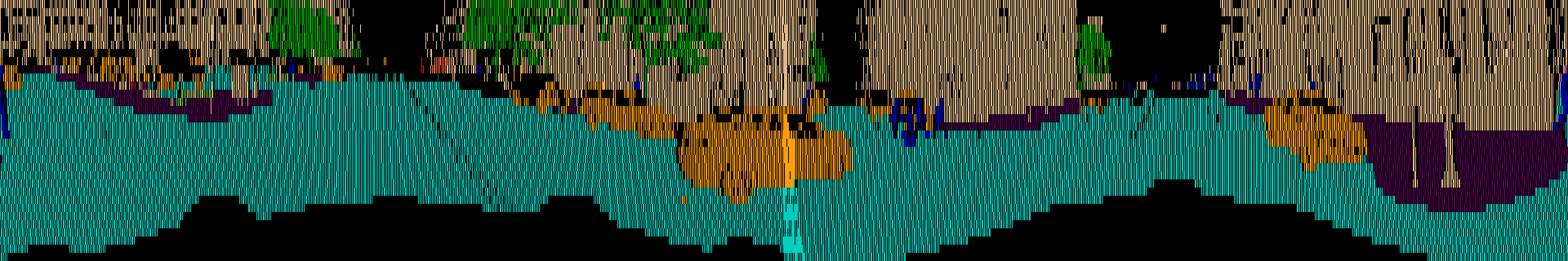}
    
    \includegraphics[width=\textwidth]{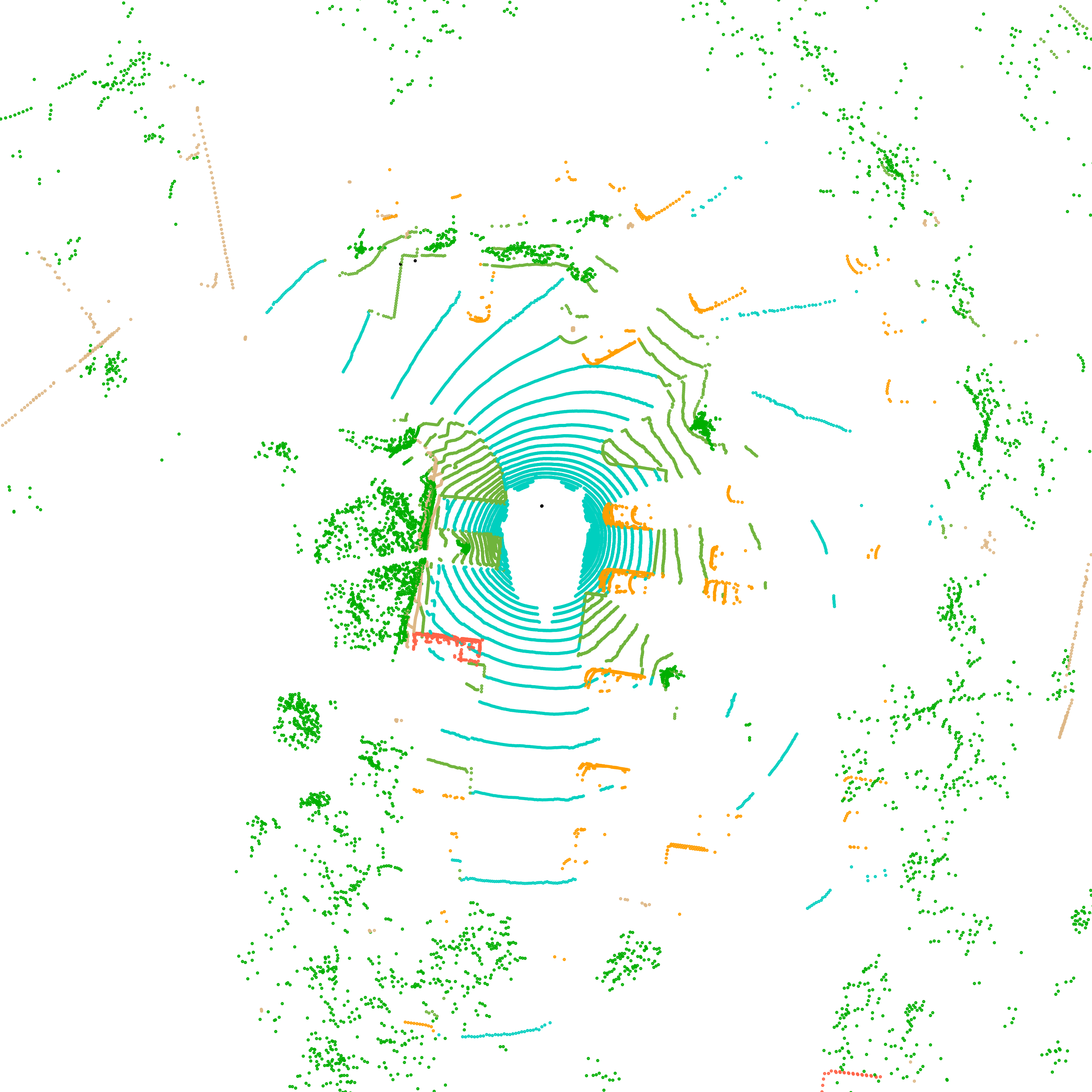}
    \includegraphics[width=\textwidth]{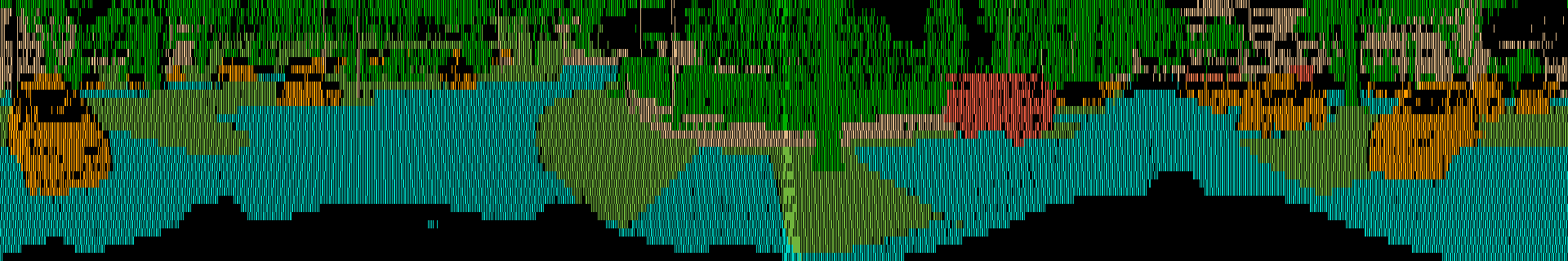}
    \caption{Ground truth}
\end{subfigure}
\begin{subfigure}[b]{.32\linewidth}
    \includegraphics[width=\textwidth]{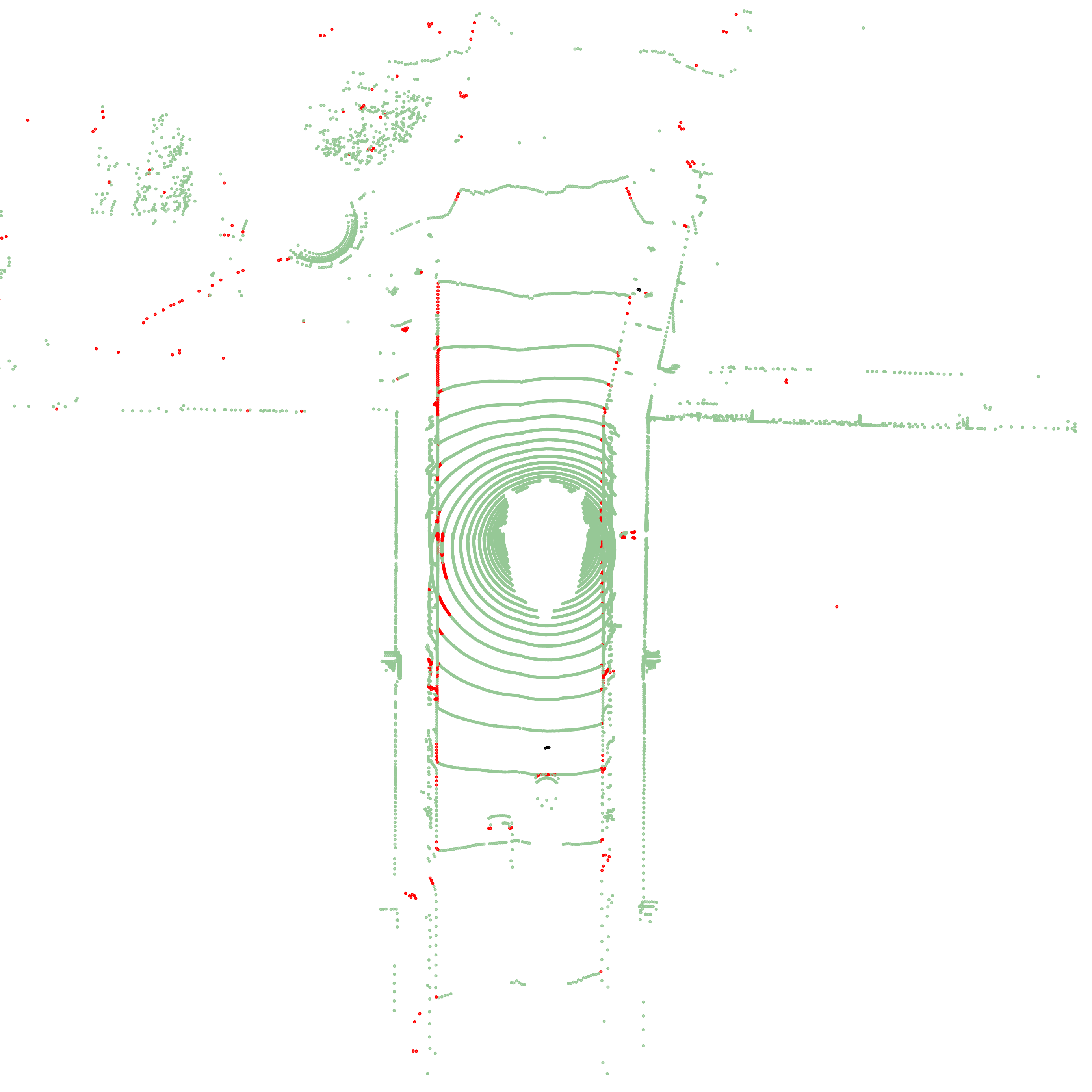}
    \includegraphics[width=\textwidth]{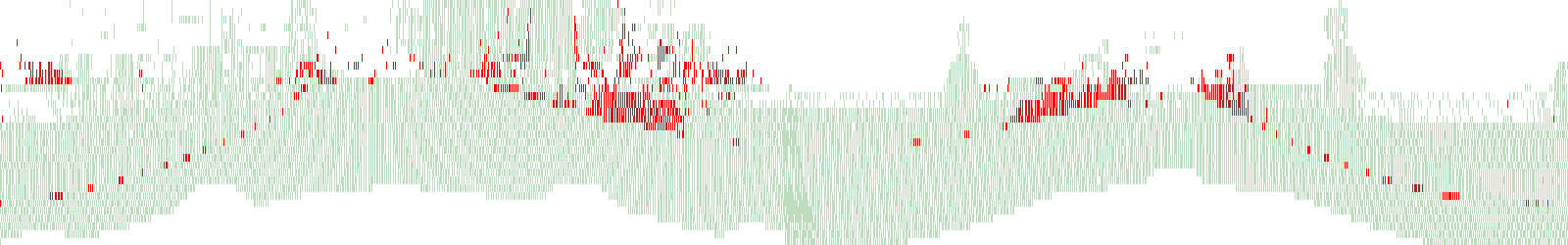}

    \includegraphics[width=\textwidth]{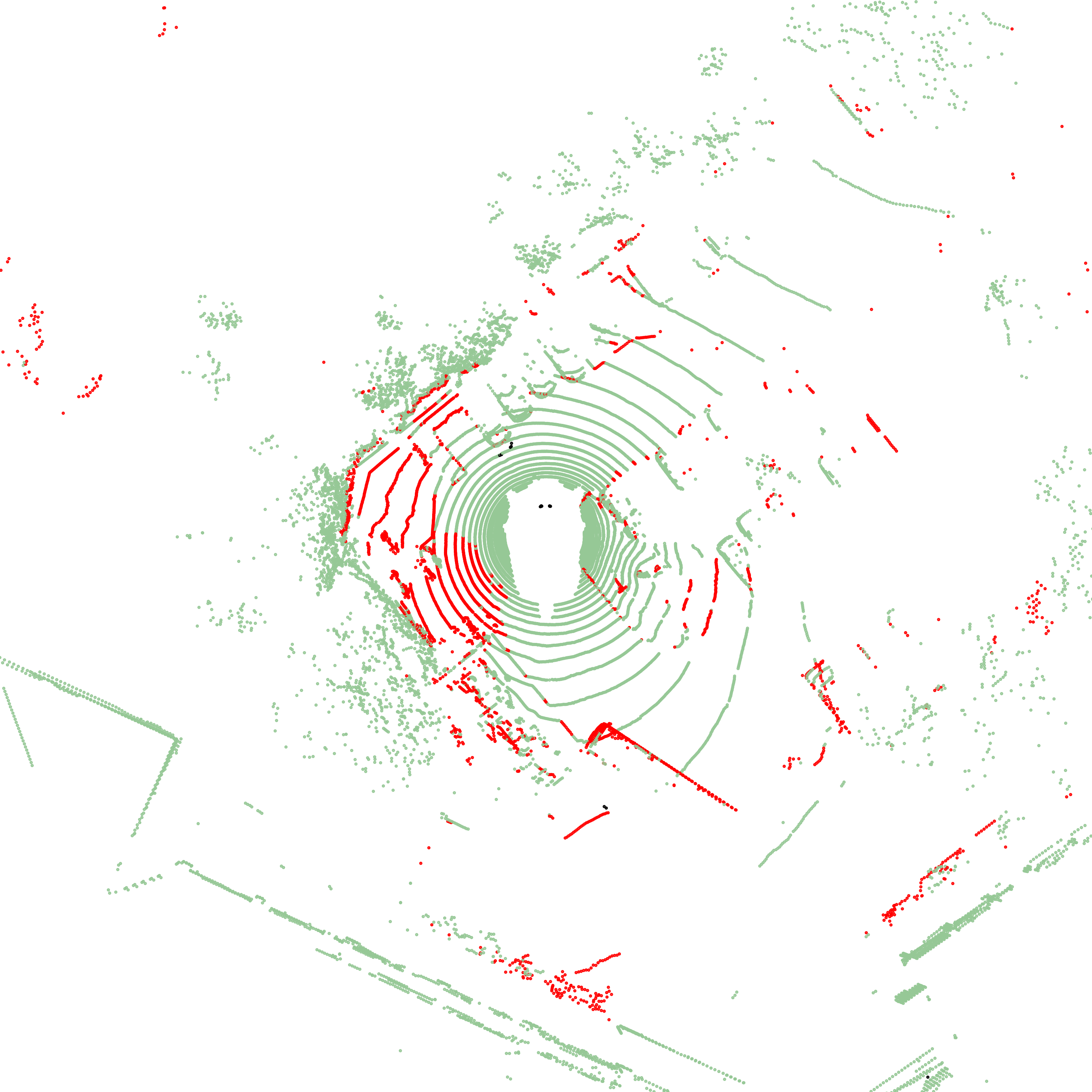}
    \includegraphics[width=\textwidth]{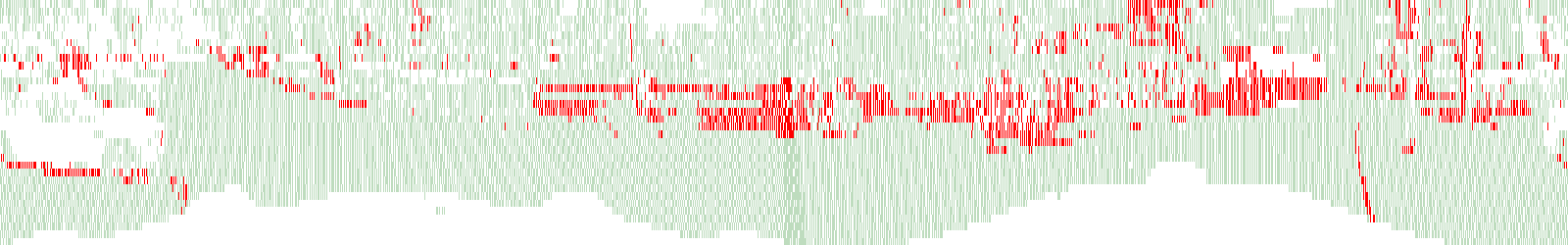}

    \includegraphics[width=\textwidth]{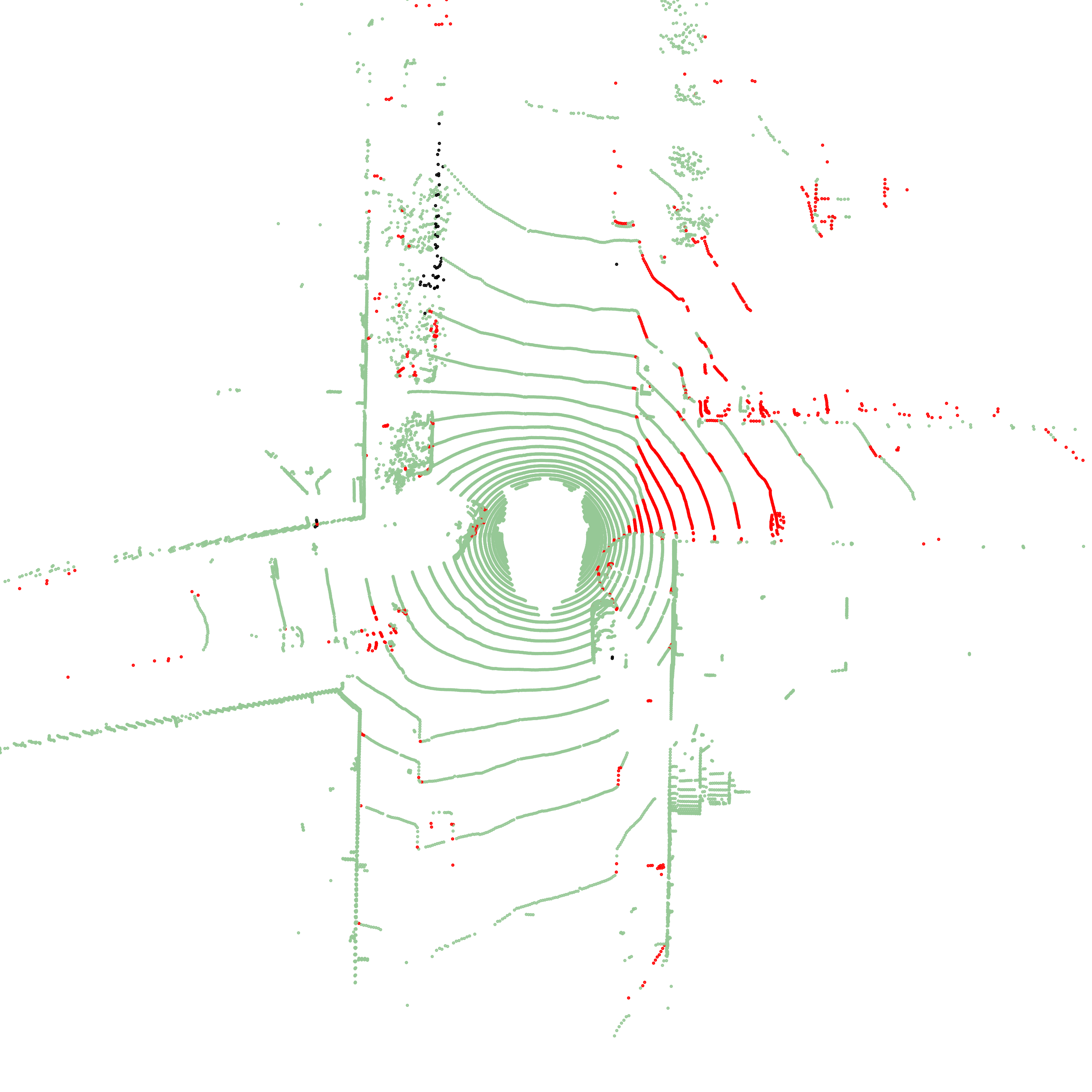}
    \includegraphics[width=\textwidth]{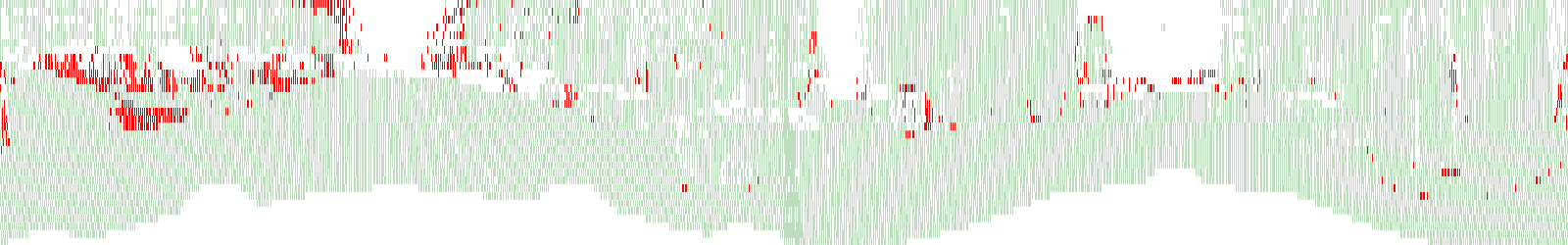}
    
    \includegraphics[width=\textwidth]{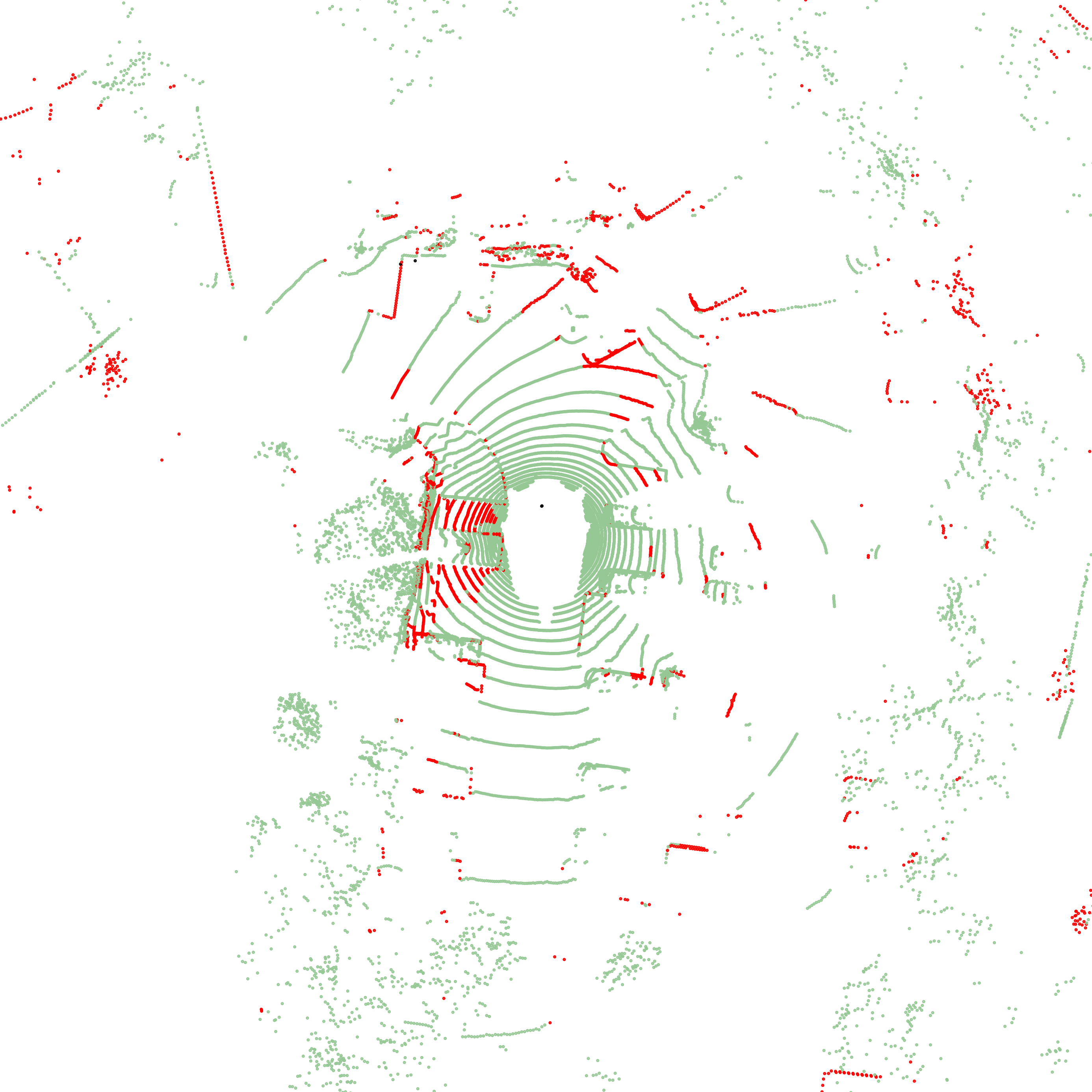}
    \includegraphics[width=\textwidth]{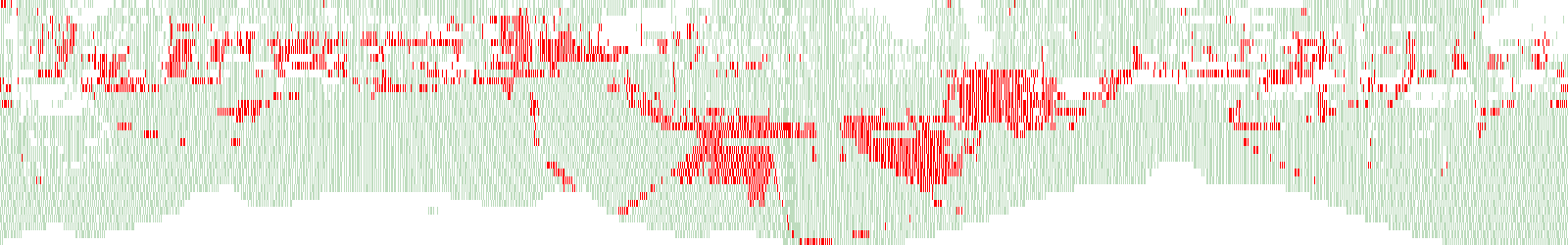}
    \caption{LaserMix~\cite{kong2023lasermix}}
\end{subfigure}
\begin{subfigure}[b]{.32\linewidth}
    \includegraphics[width=\textwidth]{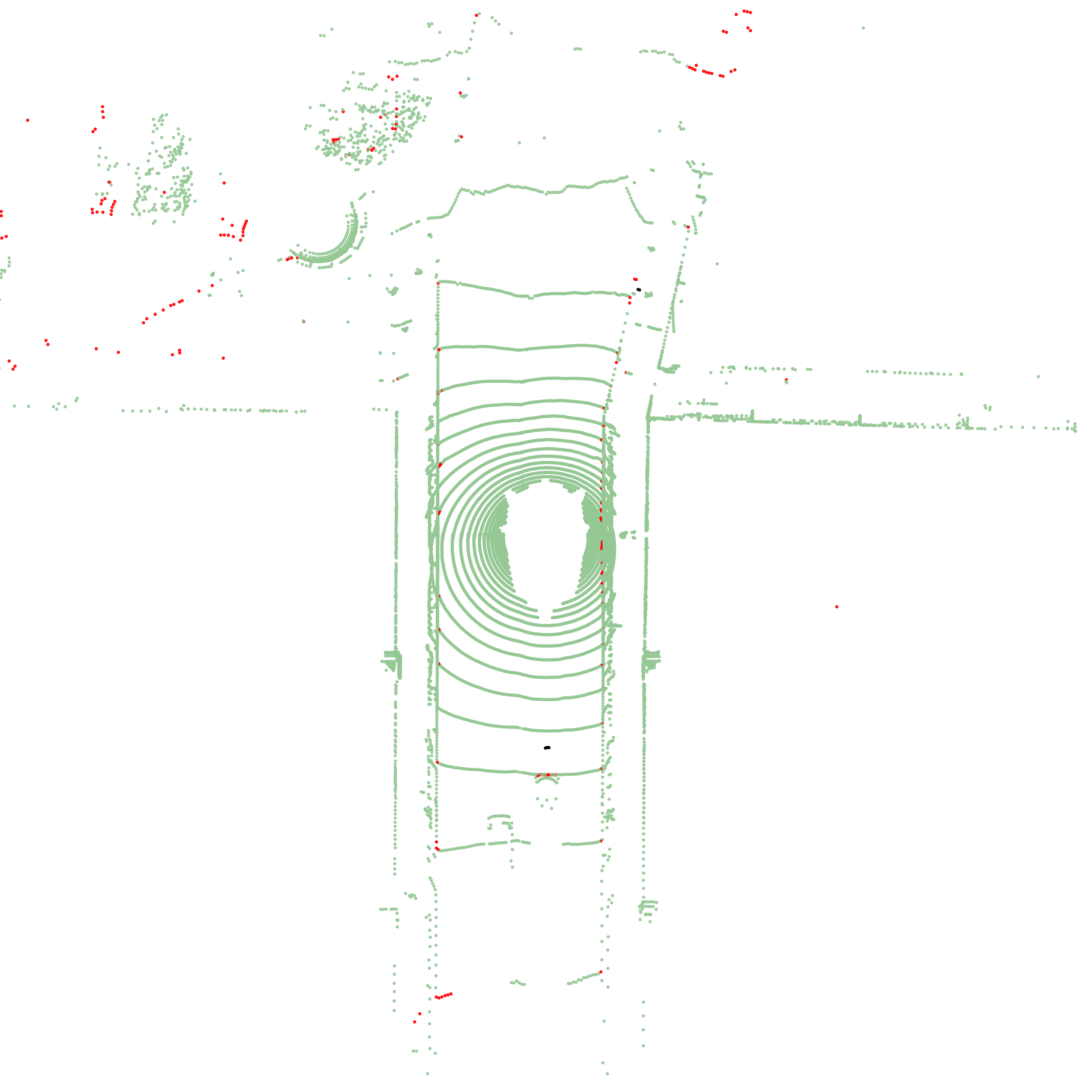}
    \includegraphics[width=\textwidth]{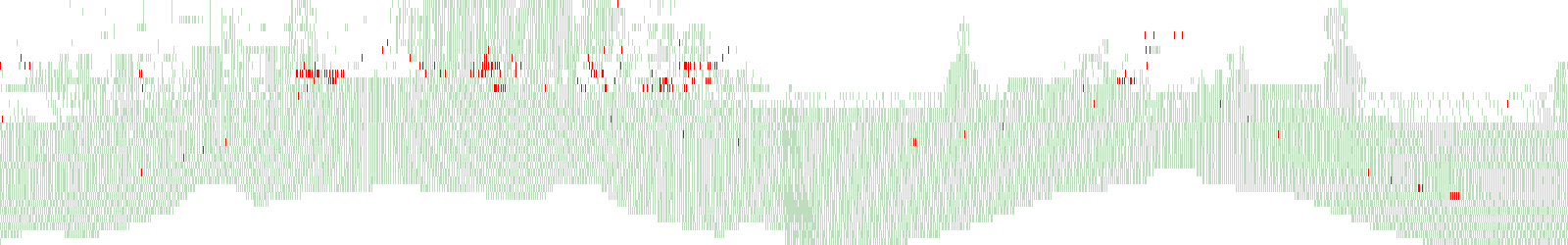}
    
    \includegraphics[width=\textwidth]{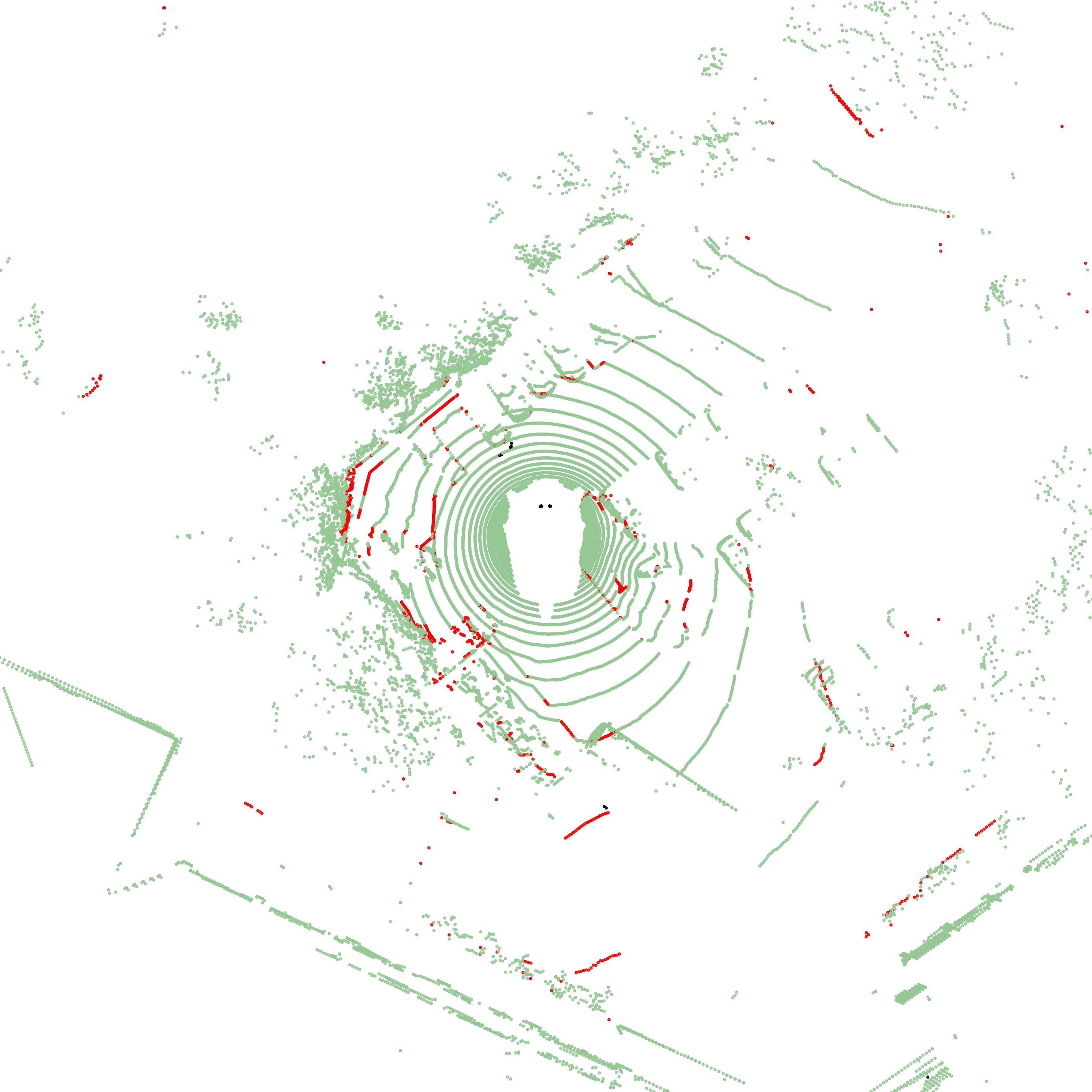}
    \includegraphics[width=\textwidth]{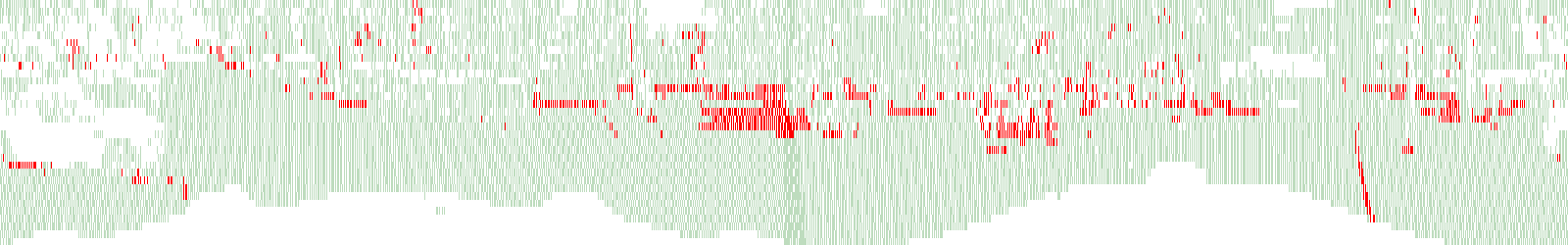}
    
    \includegraphics[width=\textwidth]{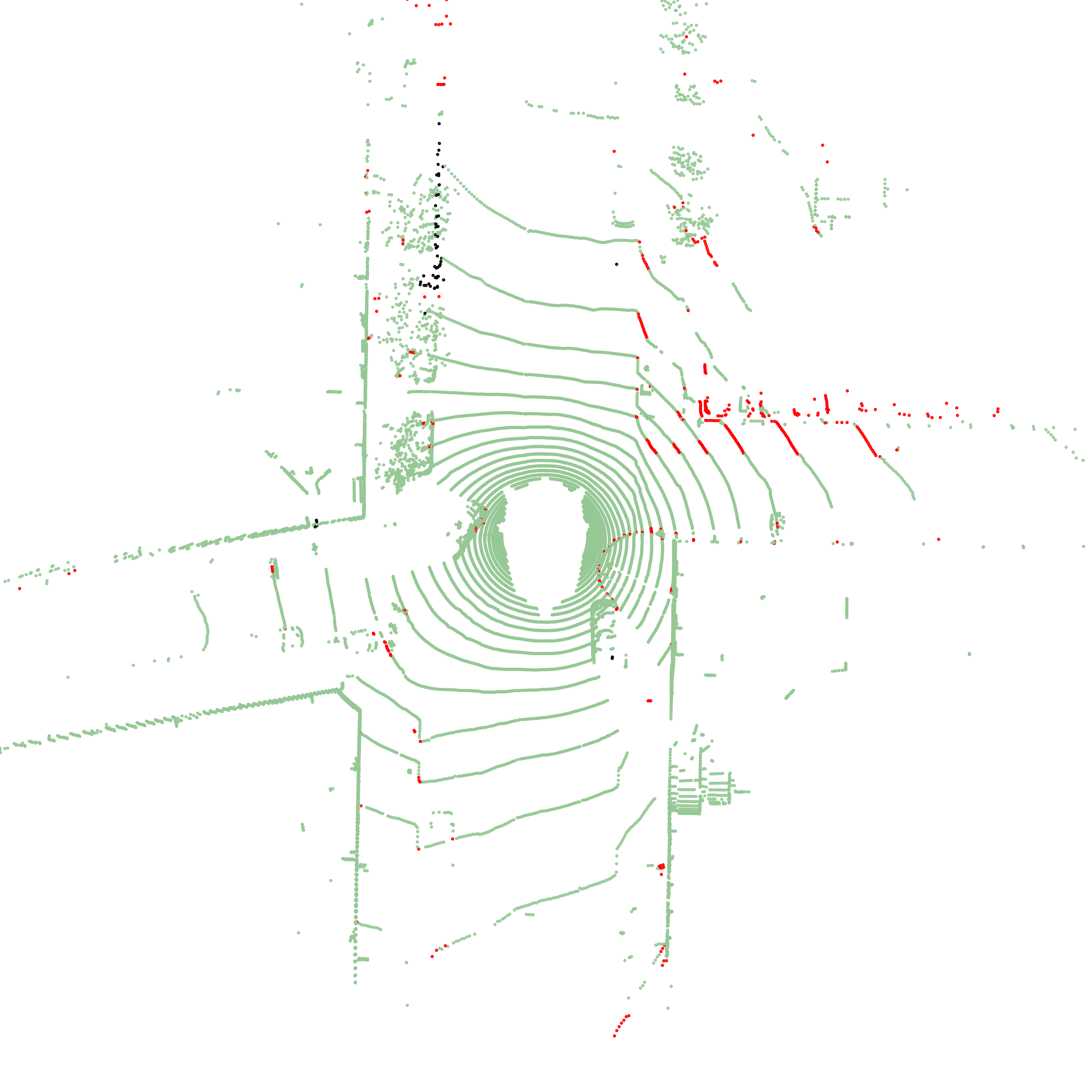}
    \includegraphics[width=\textwidth]{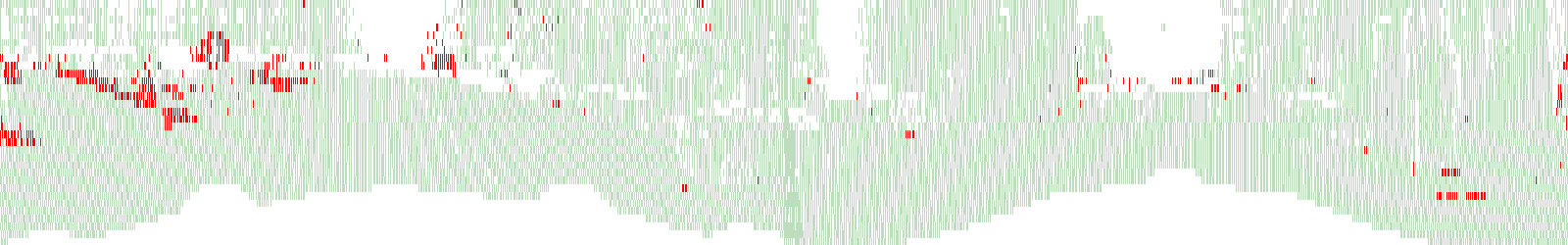}
    
    \includegraphics[width=\textwidth]{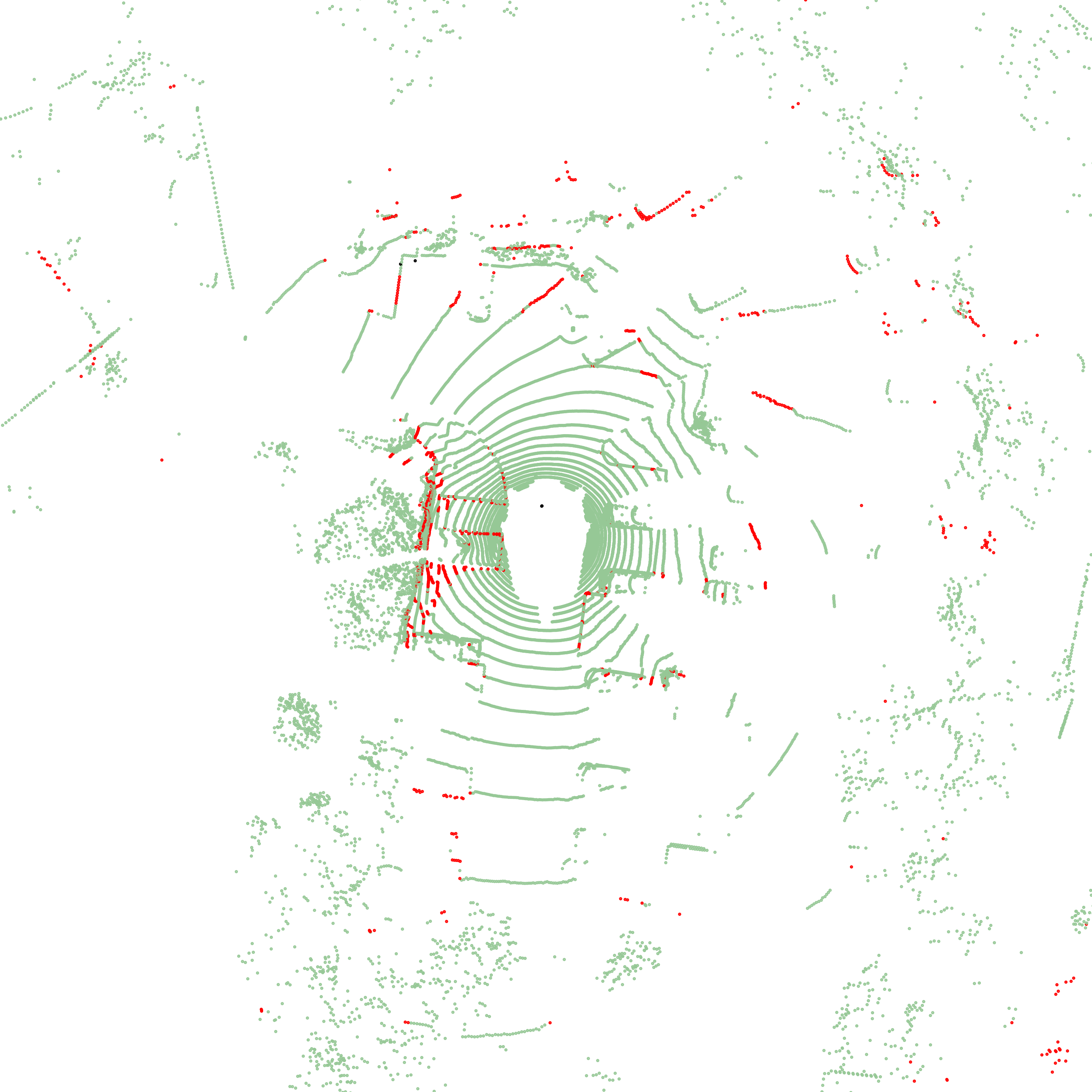}
    \includegraphics[width=\textwidth]{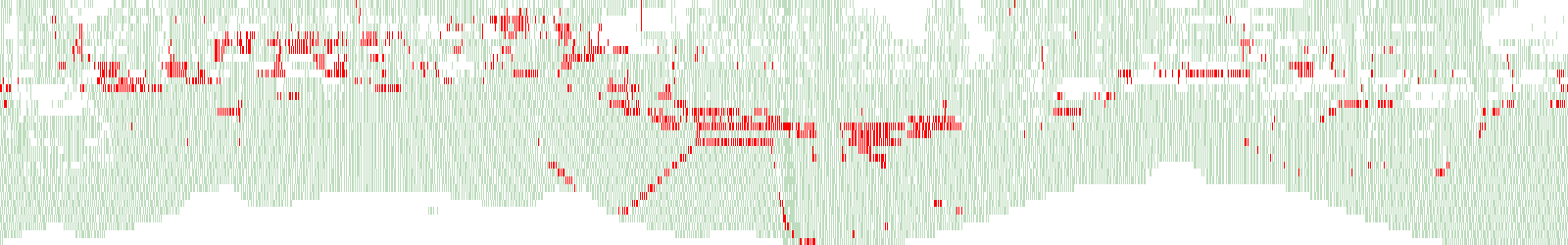}
    \caption{Ours}
\end{subfigure}
    \caption{\textbf{Additional error maps} visualised from LiDAR \textit{bird's eye view} (top) and \textit{range view} (bottom) in the  nuScenes~\cite{nunes2022segcontrast} dataset under 10\% labelled partition protocol. The correct predictions are in \textcolor{green}{green} and the mistakes are highlighted in \textcolor{red}{red}.}
    \label{fig:sup_error_map_nusc}
\end{figure*}

In Fig.~\ref{fig:sup_error_map_nusc}, we present additional visualizations of our IT2 framework on the nuScenes~\cite{caesar2020nuscenes} dataset, under 10\% labelled partition protocol, comparing it with LaserMix~\cite{kong2023lasermix}. Each case has two rows, with the top row showing the \textit{bird's eye view} results and the bottom row presenting the range images. Correct predictions are highlighted in \textcolor{green}{green}, and mistakes are indicated in \textcolor{red}{red}. Our approach consistently yields visually superior results across all cases.
\clearpage

\section{Additional Ablation studies}
\noindent\textbf{Importance of the number of GMM components.} We study the number of GMM components in Tab.~\ref{tab:rebuttal_component}, using the nuScenes dataset~\cite{caesar2020nuscenes} under the 10\% labelled data protocol. When $M=5$, we notice an improvement over the single multi-variate component ($M=1$), with an increase of 0.9 mIoU in range and 0.7 mIoU in voxel representations. However, increasing the number of components to $M=7$ results in a slight decrease in performance.\\ 
\begin{table}[ht!]
\caption{\textbf{Ablation study of Components (M)} in Gaussian Mixure Model (GMM) on the nuScenes dataset under the 10\% labelled data protocol. The best results are highlighted in \textcolor{darkred}{red}.}
\centering
\label{tab:rebuttal_component}
\resizebox{.65\textwidth}{!}{\begin{tabular}{c|cccc}
\specialrule{1pt}{0pt}{0pt}
\multicolumn{5}{c}{nuScenes (with 10\% labelled data)} \\
\hline
Component (M)     & M=1     & M=3       & M=5    & M=7    \\
range Repr.       & 70.4  & 70.8       &    \textcolor{darkred}{71.3}    & 71.2         \\
voxel Repr.       & 71.4   & 71.7       &    \textcolor{darkred}{72.1}   & 71.9         \\
\specialrule{1pt}{0pt}{0pt}
\end{tabular}}
\end{table}

\noindent \textbf{LaserMix~\cite{kong2023lasermix} with contrastive learning.} As shown in Tab.~\ref{tab:rebuttal_lasermix}, we applied our cross-distribution (GMM-based) contrastive learning (ctrs.) to LaserMix~\cite{kong2023lasermix} using the voxel representation, as their range model's code is not public. 
Although LaserMix+ctrs. improves over the original LaserMix by 0.9\% and 1.2\% under the 1\% and 10\% labelled protocols on the nuScenes dataset~\cite{caesar2020nuscenes}, respectively, our IT2 provides further improvements over LaserMix+ctrs of 1.3\% and 1.1\%, respectively.\\
\begin{table}[ht!]
\caption{Comparison between our IT2 and LaserMix~\cite{kong2023lasermix} with our contrastive learning (i.e., ctrs.) on the nuScenes dataset based on voxel representation. Our results are highlighted in \textcolor{darkred}{red}.}
\label{tab:rebuttal_lasermix}
\centering
\resizebox{.6\textwidth}{!}{
\begin{tabular}{c|cc}
\specialrule{1pt}{0pt}{0pt}
\multirow{2}{*}{Method} & \multicolumn{2}{c}{nuScenes} \\
\cline{2-3}
                        & 1\%     & 10\%       \\
\hline
LaserMix~\cite{kong2023lasermix}          &   55.3      &   69.9         \\
\hline
LaserMix~\cite{kong2023lasermix} +ctrs.          &    56.2     &   71.0          \\
IT2                     &   \textcolor{darkred}{57.5} \textcolor{gray}{(1.3$\uparrow$)}      &   \textcolor{darkred}{72.1} \textcolor{gray}{(1.1$\uparrow$)}          \\
\specialrule{1pt}{0pt}{0pt}
\end{tabular}}
\end{table}

\noindent \textbf{Sensitiveness of the parameter Temperature (Temp.) in Contrastive learning}. As demonstrated in Tab.~\ref{tab:rebuttal_temp}, the Temp. parameter significantly influences both the ContrasSeg~\cite{wang2021exploring} method and our approach. Minor adjustments from $t=0.05$ to $t=0.10$ and from $t=0.10$ to $t=0.15$ result in nearly 1\% performance differences for both methods. 
Such large differences have also been noted in  
the `Supervised Contrastive Learning' paper~\cite{khosla2020supervised} (Page 8, Fig.4). 
A potential reason for this is the relatively high weight of 1 to the InfoNCE loss for all experiments. \\
\begin{table}[ht!]
\caption{Ablation study of different \textbf{temperature (Temp.)} values comparing our methods with ContrasSeg~\cite{wang2021exploring} on the nuScenes dataset, under the 10\% labelled data protocol. The best results for both methods are highlighted in \textcolor{darkred}{red}.}
\label{tab:rebuttal_temp}
\centering
\resizebox{.65\textwidth}{!}{\begin{tabular}{c|ccc|ccc}
\specialrule{1pt}{0pt}{0pt}
Repr.     & \multicolumn{3}{c|}{range} & \multicolumn{3}{c}{voxel} \\
\hline
Temp. (t) & t=0.05  &  t=0.10 & t=0.15 & t=0.05  & t=0.10 & t=0.15 \\
\hline
ContrasSeg~\cite{wang2021exploring}   &     69.4            &      \textcolor{darkred}{70.3}  &  69.3 &  70.1            &   \textcolor{darkred}{71.2}    &  70.5 \\
Ours      &  70.3           &   \textcolor{darkred}{71.3}   &  70.4 &   71.0          &  \textcolor{darkred}{72.1}  &     71.5 \\
\specialrule{1pt}{0pt}{0pt}
\end{tabular}}
\end{table}

\clearpage

\end{document}